\RequirePackage{amsmath}
\RequirePackage{fix-cm}

\documentclass[twocolumn]{svjour3}          

\usepackage[T1]{fontenc}
\usepackage{textcomp}

\usepackage{color}
\usepackage{xcolor}
\usepackage{epsfig}
\usepackage{graphicx}
\usepackage{amssymb}
\usepackage{amsfonts}
\usepackage{array}
\usepackage[hidelinks]{hyperref}
\usepackage{bookmark}
\usepackage{algorithm}
\usepackage{algpseudocode}
\usepackage{subcaption} 
\usepackage{booktabs}
\usepackage{natbib}

\usepackage{booktabs} 
\usepackage[misc]{ifsym} 
\smartqed  

\renewcommand{\mathbf}{\boldsymbol}
\newcommand{\nop}[1]{}

\newcommand{\norm}[1]{\left\lVert#1\right\rVert}
\newcommand{\thickhline}{%
    \noalign {\ifnum 0=`}\fi \hrule height 1pt
    \futurelet \reserved@a \@xhline
}
\newcolumntype{"}{@{\hskip\tabcolsep\vrule width 2pt\hskip\tabcolsep}}
    
\DeclareMathOperator*{\argmaxA}{arg\,max}
\DeclareMathOperator{\argmaxH}{argmax}

\journalname{International Journal on Computer Vision}

\hypersetup{
    colorlinks,
    linkcolor={black!50!black},
    citecolor={black!50!black},
    urlcolor={blue!80!black}
}

\begin{document}

\title{CAPTAIN: Comprehensive Composition Assistance for Photo Taking
}

\author{ Farshid Farhat \and
         Mohammad Mahdi Kamani \and
         James Z. Wang 
}
\institute{F. Farhat (\Letter)
\at           School of Electrical Engineering and Computer Science \\
              The Pennsylvania State University, University Park, PA, USA \\
              \email{fuf111@psu.edu}           
			\and
              M.M. Kamani \and J.Z. Wang 
              \at College of Information Sciences and Technology\\
              The Pennsylvania State University, University Park, PA, USA \\
              \email{mqk5591@psu.edu}           
            \and
              J.Z. Wang\\
              \email{jwang@ist.psu.edu}           
}

\date{Received: date / Accepted: date}

\maketitle

\begin{abstract} 
Many people are interested in taking astonishing photos and  sharing with others. 
Emerging high-tech hardware and software facilitate ubiquitousness and functionality of digital photography.
Because composition matters in photography, researchers have leveraged some common composition techniques 
to assess the aesthetic quality of photos computationally. 
However, composition techniques developed by professionals are far more diverse than well-documented techniques can cover. 
We leverage the vast underexplored innovations in photography for computational composition assistance.
We propose a comprehensive framework, named CAPTAIN (\underline{C}omposition \underline{A}ssistance for \underline{P}hoto \underline{Ta}k\underline{in}g), containing integrated deep-learned semantic detectors, sub-genre categorization, artistic pose clustering, personalized aesthetics-based image retrieval, and style set matching. 
The framework is backed by a large dataset crawled from a photo-sharing Website with mostly photography enthusiasts and professionals.
The work proposes a sequence of steps that have not been explored in the past by researchers. 
The work addresses personal preferences for composition through presenting 
a ranked-list of photographs to the user based on user-specified weights in the similarity measure.
The matching algorithm recognizes the best shot among a sequence of shots with respect to the user's preferred style set. We have conducted a number of experiments on the newly proposed components and reported findings. A user study demonstrates that the work is useful to those taking photos.
\end{abstract}

\section{Introduction}
\label{sec:intro}

Digital photography is of great interest to many people, regardless of whether they are professionals or amateurs. For example, people on social networks often share their photos with their friends and family. It has been estimated that over a billion photos are taken every year, and many people take photos with smartphones primarily. Smartphones' increasing computing power and ability to connect to more powerful computing platforms via the network make them potentially useful as a composition assistant to amateur photographers~\citep{yao2012oscar}.

Besides, emerging technologies, including artificial intelligence (AI)-chips and AI-aware mobile applications, provide more opportunities for composition assistance. 
Taking stunning photos often needs expertise and experience at a level that professional photographers have. Like in other visual arts, a lack of common alphabet similar to music notes or mathematical equations makes transferring of knowledge in photography difficult. To many amateurs, as a result, photography is mysterious and gaining skills is not easy and cannot be done quickly. Nonetheless, many people are fascinated about professional-quality photos and desire to have the ability to create similar-quality photos themselves. 

\begin{figure*}[!tb]
\centering
\resizebox{1.04\textwidth}{!}{\begin{tabular}{m{0.0in}m{7in}}
\rotatebox{90}{full-body} &
\includegraphics[height=0.8in]{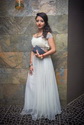}
\includegraphics[height=0.8in]{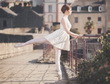}
\includegraphics[height=0.8in]{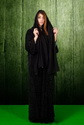}
\includegraphics[height=0.8in]{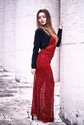}
\includegraphics[height=0.8in]{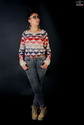}
\includegraphics[height=0.8in]{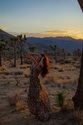}
\includegraphics[height=0.8in]{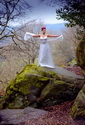}
\includegraphics[height=0.8in]{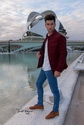}
\includegraphics[height=0.8in]{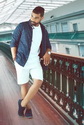}
\includegraphics[height=0.8in]{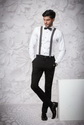}
\includegraphics[height=0.8in]{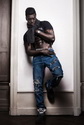}
\vspace{0.5mm} \\
\rotatebox{90}{upper-body} &
\includegraphics[height=0.85in]{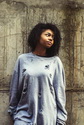}
\includegraphics[height=0.85in]{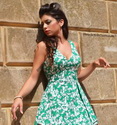}
\includegraphics[height=0.85in]{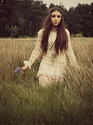}
\includegraphics[height=0.85in]{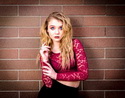}
\includegraphics[height=0.85in]{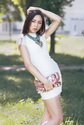}
\includegraphics[height=0.85in]{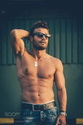}
\includegraphics[height=0.85in]{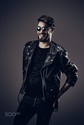}
\includegraphics[height=0.85in]{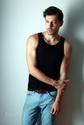}
\includegraphics[height=0.85in]{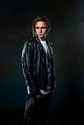}
\includegraphics[height=0.85in]{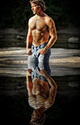}
\vspace{0.5mm} \\
\rotatebox{90}{seated female} &
\includegraphics[height=0.82in]{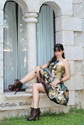}
\includegraphics[height=0.82in]{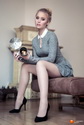}
\includegraphics[height=0.82in]{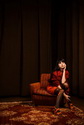}
\includegraphics[height=0.82in]{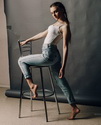}
\includegraphics[height=0.82in]{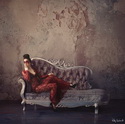}
\includegraphics[height=0.82in]{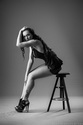}
\includegraphics[height=0.82in]{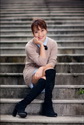}
\includegraphics[height=0.82in]{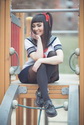}
\includegraphics[height=0.82in]{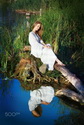}
\includegraphics[height=0.82in]{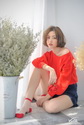}
\includegraphics[height=0.82in]{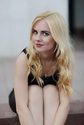}
\vspace{0.5mm} \\
\rotatebox{90}{seated male} &
\includegraphics[height=0.785in]{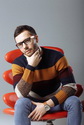}
\includegraphics[height=0.785in]{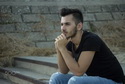}
\includegraphics[height=0.785in]{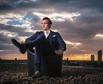}
\includegraphics[height=0.785in]{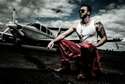}
\includegraphics[height=0.785in]{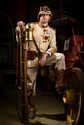}
\includegraphics[height=0.785in]{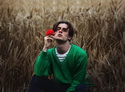}
\includegraphics[height=0.785in]{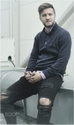}
\includegraphics[height=0.785in]{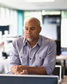}
\vspace{0.5mm} \\
\rotatebox{90}{facial} &
\includegraphics[height=0.79in]{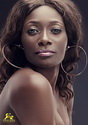}
\includegraphics[height=0.79in]{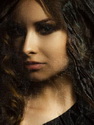}
\includegraphics[height=0.79in]{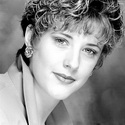}
\includegraphics[height=0.79in]{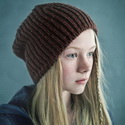}
\includegraphics[height=0.79in]{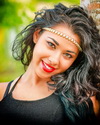}
\includegraphics[height=0.79in]{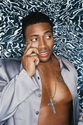}
\includegraphics[height=0.79in]{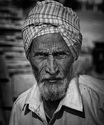}
\includegraphics[height=0.79in]{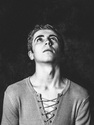}
\includegraphics[height=0.79in]{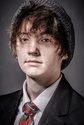}
\includegraphics[height=0.79in]{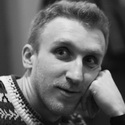}
\vspace{0.5mm} \\
\rotatebox{90}{forest} &
\includegraphics[height=0.82in]{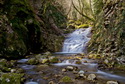}
\includegraphics[height=0.82in]{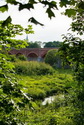}
\includegraphics[height=0.82in]{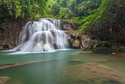}
\includegraphics[height=0.82in]{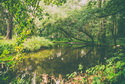}
\includegraphics[height=0.82in]{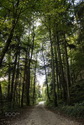}
\includegraphics[height=0.82in]{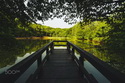}
\includegraphics[height=0.82in]{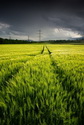}
\vspace{0.5mm} \\
\rotatebox{90}{seascape} &
\includegraphics[height=0.81in]{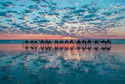}
\includegraphics[height=0.81in]{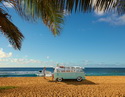}
\includegraphics[height=0.81in]{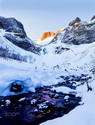}
\includegraphics[height=0.81in]{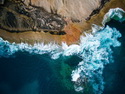}
\includegraphics[height=0.81in]{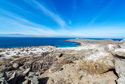}
\includegraphics[height=0.81in]{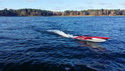}
\vspace{0.5mm} \\
\end{tabular}
}
\caption{Photos retrieved from the dataset based on the photo category and/or subject gender. Each retrieved result shows
a collection of photography ideas that can be used by an amateur to compose photos for a given situation. 
Photos of the dataset were crawled from the 500px Website.}
\label{fig:intro}
\end{figure*}

Because aesthetics in photography is strongly linked to human creativity, it is daunting for an artificial intelligence (AI) to compose photographs at a given scene or a given studio setup that can impress people in a way professional photographers do. In our work, we attempt to connect human creativity as demonstrated through their creative works with AI.

Aesthetics and composition in photography have generally been heuristically explored and known as a collection of rules or principles such as balance, geometry, symmetry, the rule of thirds, and framing~\citep{lauer2011design,valenzuela2012practice,krages2012photography}. 
It is well known that professional photographers take a lot of pictures, and through their practice they gain experience and knowledge which in turn enable them to be creative. They have written about their knowledge in photography books~\citep{smith2012posing,valenzuela2014posing,grey2014master}. Some composition rules or principles have been well articulated and many amateurs make use of these principles in their photo taking. However, we argue that the set of known rules or principles can hardly cover the creativity and experience of thousands of photographers around the world~\citep{valenzuela2012practice,matthew2010101,rice2006master}. 

In order to capture an aesthetically appealing photo, photographers often integrate different visual elements. For instance, the beauty of a full-body portrait depends on the foreground positions of the human limbs as well as the constellation of the background objects (if any). 
A good portrait is often a product of an appropriate color palette, an appealing composition of shapes, and an interesting human pose. There is no unique photography idea for a given situation, and people have different opinions on those ideas depending on their cultural background, gender, age, experience, and emotional state. As a result, if the aesthetic quality of photos is quantified by one number, one is making an unrealistic assumption that different people share the same opinions on the same photo. 

For an amateur, it would be helpful if an AI can help select {\it photography ideas}, from thousands of such ideas available through online photos taken with professional quality, for a given scene or a given studio setup. The key technical difficulties for accomplishing this goal are (1) finding a suitable mapping between a professional-quality photo of a scene and the underlying photography ideas, (2) there are virtually unlimited number of photography ideas, and (3) to provide meaningful and intuitive in-situ assistance to the photographer based on personal preference. Our work tackles these challenges using a data-driven approach based on retrieval from a large dataset of professional-quality photos. 

The multimedia and computer vision research communities have been leveraging some of the photography composition rules or principles for aesthetics and composition assessment~\citep{datta2006studying,ke2006design,luo2008photo,wong2009saliency,marchesotti2011assessing}. Other approaches manipulated (modified) the photo to comply with artistic rules~\citep{bhattacharya2010framework,bhattacharya2011holistic}, and such systems are referred to as auto-composition or re-composition. The techniques include smart cropping~\citep{suh2003automatic,santella2006gaze,stentiford2007attention,zhang2005auto,park2012modeling,samii2015data,yan2013learning}, warping~\citep{liu2010realtime,chang2015r2p}, patch re-arrangement \citep{barnes2009patchmatch,cho2008patch,pritch2009shift}, cutting and pasting~\citep{bhattacharya2011holistic,zhang2005auto}, and seam carving~\citep{guo2012improving,li2015seam}. However, they do not help an amateur photographer capture a more impressive photo to begin with. More specifically in portrait photography, there have been rule-based assessment models~\citep{khan2012evaluating,males2013aesthetic} using known photography basics to evaluate portraits, and facial assessment models~\citep{xue2013feature,lienhard2014photo,lienhard2015low,lienhard2015predict,redi2015beauty} exploiting features including smile, age, and gender from face. On-site feedback systems~\citep{yao2012oscar,li2015photo} have been developed to help amateur photographers by retrieving images with similar composition, but the system is limited to basic composition categories (e.g. horizontal, vertical, diagonal, textured, and centered). More recently, perspective-related techniques~\citep{zhou2017detecting}, the triangle technique~\citep{he2018discovering} and some portrait composition techniques~\citep{farhat2017intelligent} have also been exploited.

\begin{figure*}[!tb]
\centering
\includegraphics[width=1.0\linewidth]{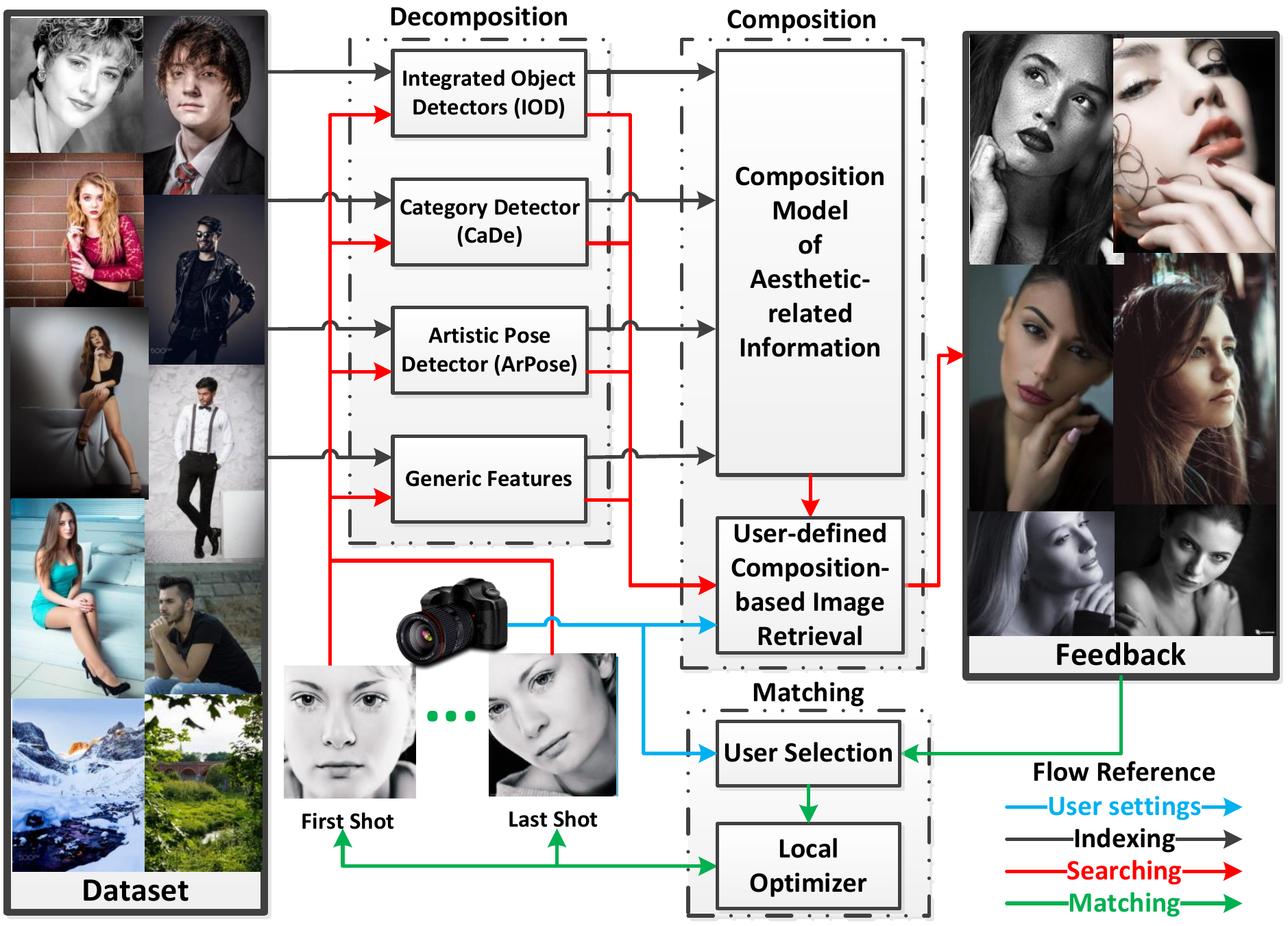}\\ 
\caption{The flowchart of our composition assistance framework: The blue, black, red, and green flows show the user settings, the indexing process, the searching process, and the matching process, respectively. The decomposition step (the box with dashed line) extracts the aesthetics-related information of the images and computes our composition model. The composition step (another box with dashed line) searches for well-composed images in our dataset based on the aesthetics-related information and user-specified preferences. The matching step (the other small box at the bottom) considers the next shots, and finds the shot that is closest to the user-preferred style set.}
\label{fig:method}
\end{figure*}

We investigate a holistic framework for helping people take a better shot with regard to their current photography location and need. The framework addresses the differences in preferences of the users through adjusting the ranking process used to retrieve exemplars. After getting a first shot from the camera, our framework provides some highly-scored related photos as pre-composed ``recipes'' (i.e. photography ideas) for the user to consider. As an example, regarding some personalized criteria (such as photo category and subject gender), Figure~\ref{fig:intro} shows sample results retrieved from the photo dataset. These photos illustrate various locations, scenes, and categories. One can argue that while photos in the same row have the same category or gender, each individual photo has a photography idea(s) that is different from those used in other photos of the same row. For example, in the 2nd and 3rd photos from the right in the 1st row, the subjects cross their legs and bend one of the knees to form a triangle in the resulting photo. As mentioned before, the triangle technique is a popular technique used by professionals. While both uses the technique, the way they use them is different, forming different photography ideas. Similarly, the first subject in the 3rd row sits beside the arch with an apropos pose, creating a triangle, but is different from the triangles formed in the earlier examples.

We address the complexity of transferring photography idea(s) to a user through providing useful exemplar-based feedbacks. Specifically, we break down the scene that the user wants to take a photo from into composition primitives, and then build them up for a better-composed shot using highly-rated similar photos from the dataset. To accommodate the user's individual preferences, we perform personalized aesthetics-related image retrieval (PAIR). Figure~\ref{fig:method} shows the flowchart of our approach for assisting photographers in taking an improved photo. Based on the first shot as a query, some highly-rated photos are retrieved from the collected dataset using the user-specified preferences (USP) and our composition model (CM). Then, the results are shown to the photographer to select some of them as a user-preferred style set. The camera then takes a sequence of shots, from which the one that is the closest match to the style set is chosen. The details of the procedure will be explained later. The {\bf main contributions} of our work are as follows:
\begin{itemize}
\item We propose a new framework that finds the mapping between a photo and its potential underlying photography idea(s) through decomposing the photo into composition ingredients. Through such a framework, it is possible to leverage the virtually unlimited number of photography ideas available on the Internet. We design the {\it decomposition} step to extract composition primitives of a query shot using various detectors including newly developed integrated object detector (IOD), category detector (CaDe), and artistic pose detector (ArPose). 
The IOD consists of a collection of performance-enhanced detectors, 
the integration of which substantially boosts the detection accuracy by hysteresis detection and makes them unified and compatible with any detector. The CaDe has top-down hierarchical clustering and multi-class categorization to leverage sub-genre information. The ArPose performs pose clustering to extract pose information using skeleton context features.

\item We address the complexity of transferring photography knowledge, caused by the existence of abundant, diverse, and correlated photography ideas for any given scene, by providing meaningful and intuitive feedback to amateur photographers. We design the {\it composition} step to perform personalized aesthetics-related image retrieval from a large managed dataset containing over 200,000 highly-rated aesthetically composed photos covering a large number of photography ideas and different categories. 

\item We accommodate the user's personal preferences for composition, through showing a ranked-list of photos to the user based on user-specified preferences (USP). The framework further helps the user to select a shot among a sequence of shots that is optimal with respect to the user's preferred style set.
\end{itemize}

Our proposed framework is not limited to any specific photography genre. It can be generalized to other genres such as architectural or closeup photography. 


In the remainder of the paper, we explain the realization of our framework in detail. After discussing related work, we describe the dataset that we have collected and indexed, which serves as the main asset for the computer program to analyze the available photography ideas on the Internet. Next, we explain our \textit{decomposition} and \textit{composition} strategies, which are the most significant parts of the design. After that, our \textit{matching} stage will be presented as the concluding stage in the photo taking process. We provide qualitative results as we describe the method, and experimental results afterward to show  how each part of our framework compares with the state of the art.

\section{Related Work}
\label{sec:related}
Existing work closely related to ours is categorized into four groups, covering aesthetic quality assessment, composition, portrait, and on-site feedback, respectively. 

\subsection{Aesthetic Quality Assessment}
Books on professional photography \citep{lauer2011design,valenzuela2012practice,krages2012photography,matthew2010101,rice2006master,smith2012posing,valenzuela2014posing,grey2014master} guide people to master skills of taking striking photos practically in various situations. However,  learning through them takes a lot of time and practice. Existing technical approaches attempt to automatize this process, but they are limited and mostly focused on offline evaluation or active manipulation of the photos after they are taken. Basic image aesthetics and composition rules in visual art~\citep{lauer2011design,valenzuela2012practice,krages2012photography}, including geometry, color palette, and the rule of thirds, have first been studied computationally by
~\citet{datta2006studying} and
~\citet{ke2006design} as visual aesthetic features. 

~\citet{luo2008photo}, and
~\citet{wong2009saliency} attempted to leverage a saliency map method, and considered the features of the salient parts, because more appealing parts of an image often reside in the prominent region.
~\citet{marchesotti2011assessing} showed that generic image descriptors was useful to assess image aesthetics, and built a generic dataset for composition assessment - the Aesthetic Visual Analysis (AVA) dataset~\citep{MurrayMP12}. Deep learning based approaches~\citep{lu2015rating,mai2016composition,talebi2018nima,liu2018deep} exploit customized architectures to train image aesthetic-quality models with annotated datasets, and the outcome is an estimation for actual (average) aesthetic rating of an image.

\subsection{Image Auto-Composition and Re-Composition} 
Auto-composition systems \citep{bhattacharya2010framework,bhattacharya2011holistic} actively manipulate and then re-compose the\break taken photo for a better view. Cropping techniques \citep{suh2003automatic,santella2006gaze,stentiford2007attention} separate the region of interest (ROI) by the help of a saliency map, an eye fixation, basic aesthetic rules \citep{zhang2005auto}, or visual aesthetics features in the salient region\citep{park2012modeling,samii2015data,yan2013learning}. Warping~\citep{liu2010realtime} is another type of re-composition that represents image as a triangular or quad mesh, to map the image into another mesh while keeping the semantics and perspective unchanged. Also, R2P~\citep{chang2015r2p} detects the foreground part in the reference image. Then, it re-targets the salient part of the image to the best-fitted position using a graph-based algorithm.

Furthermore, patch re-arrangement techniques mend two ROIs in an image together. Pure patch rearrangements~\citep{barnes2009patchmatch,cho2008patch,pritch2009shift} detect a group of pixels on the border of the patch and match this group to the other vertical or horizontal group of pixels near the patched area. Also, cut-and-paste methods~\citep{bhattacharya2011holistic,zhang2005auto} remove the salient part, and re-paint the foreground with respect to the salient part and the borders, and then paste it to the desired position in the image. Another auto-composition system, seam carving~\citep{guo2012improving,li2015seam}, replaces useless seams.

\subsection{Assessment of Portrait Aesthetics}
While there exist prior studies on image aesthetics assessment, few considered portrait photography in depth, despite the fact that the portion of portrait genre is very high in the photography domain. Even in this domain, prior works have not explored a novel method to solve the problem in photographic portraiture, rather than combining and using
well-known features or modifying trivial ones to apply in the facial domain. We categorize prior works into two main groups: rule-based evaluation models~\citep{khan2012evaluating,males2013aesthetic} exploit known photography rules to assess portraits, and facial evaluation models~\citep{xue2013feature,lienhard2014photo,lienhard2015low,lienhard2015predict,redi2015beauty} use visual features on face like smiling, age, gender, etc.

In a rule-based evaluation model,
~\citet{khan2012evaluating} show that a small set of face-centered spatial features extend the rule of thirds and perform better than a large set of aesthetics-related features. Actually, their dataset containing 500 images from Flickr scored by 40 people is limited for a general conclusion. Their aesthetic features, especially spatial features, are close to well-known photography rules which were widely investigated before. 

~\citet{males2013aesthetic} explore the aesthetic quality of head-shots by means of some famous photography rules and low-level facial features. More specifically, sharpness and depth of field, the rule of thirds as a composition rule, contrast, lightness, hue counts and face size are exploited as their fundamental features. Unfortunately, the experimental results of the paper are limited, and it is hard to conclude for general cases.
~\citet{xue2013feature} study the design inferring portrait aesthetics with appealing facial features like smiling, orientation, to name but a few. Similarly,
~\citet{harel2006graph} exploit traditional features like hue, saturation, brightness, contrast, simplicity, sharpness, and the rule of thirds. They also extract saliency map by graph-based visual saliency. Then, they calculate the standard deviation and the main subject coincidence of the saliency map. 

The other facial evaluation models~\citep{lienhard2014photo,lienhard2015low,lienhard2015predict} use well-known low-level aesthetic features such as colorfulness, sharpness, and contrast, as well as high-level face-related features such as gender, age, and smile. Their idea is based on exploiting these features for all segmented parts of the face including hair, face, eyes, and mouth.
~\citet{redi2015beauty} show that the beauty of the portrait is related to the amount of art used in it not the subject beauty, age, race, or gender. Using a dataset derived from AVA~\citep{MurrayMP12}, they exploit a high-dimensional feature vector including aesthetic rules, biometrics and demographic features, image quality features, and fuzzy properties. Based on lasso regression output, eyes sharpness and uniqueness features have the highest rank for a good portrait.

\subsection{On-site Feedback on Photographic System}
An aesthetic assessor may find a metric to evaluate aesthetic quality of an image, but the way it conveys this information to photographer is also crucial. Because, an amateur photographer probably has no idea about how to improve the image composition. That is why providing meaningful feedback to enhance the next shots and not just image aesthetic assessment is one of our main intentions.

Giving feedback on a photographic system firstly has been introduced by
~\citet{joshi2011aesthetics}, as they suggest a real-time filter to trace and aesthetically rate the camera shots, and then the photographer retakes a better shot. On-site composition and aesthetics feedback system~\citep{yao2012oscar,li2015photo} helps smartphone users improve the quality of their taken photos by retrieving similarly composed images as a qualitative composition feedback. Also, it gives color combination feedback for having colorfulness in the next photo, and outputs the overall aesthetic rating of the input photo as well. OSCAR~\citep{yao2012oscar} is assumed to fulfill future needs of an amateur photographer, but giving such a feedback may be unrelated or unrealistic to the user, and also it is restricted to a small database in terms of coverage, diversity, and copyright. 

~\citet{xu2015real} suggest using a three-camera array to enhance the quality of the taken photos by the rule of thirds. In fact, the smartphone interface using the camera array information shows some real-time guideline to the user for taking a photo from another position. More recently general aesthetic techniques including perspective-related techniques~\citep{zhou2016detecting} and triangle technique~\citep{he2018discovering} are exploited to retrieve proper images as an on-site guidance to amateur photographers, but they are limited to basic ideas in photography while many aspects such as human pose or scene content are ignored, and these methods just try to retrieve similar photos to query photo having perspective or triangles, but the retrieved results may not be necessarily useful to amateur photographer.

\section{The Dataset}
The most valuable resource used by the computer program developed in this work is the collected dataset because it contains a large number of innovative photography ideas from around the world. We have attempted many photo-sharing websites for photography purposes including Flickr, Photo.net, DPChallenge, Instagram, Pinterest, and Unsplash. However, none of them properly cover several categories such as full-body and upper-body in portrait photography as well as urban in landscape photography. The process of searching, collecting, and updating the dataset is time consuming and taxing, hence, automating this process is quite helpful.

\subsection{Portrait and Landscape Dataset}
The dataset is gradually collected by crawling the 500px website which contains photos from millions of photographers around the world expanding their social networks of colleagues while exploiting technical and aesthetic skills to make money by marketing their photographs. To get the file list and then the images sorted by rating, we have implemented a distributed multi-IP address, block-free Python script that collects the photos having tags such as portrait, pose, human, person, woman, man, studio, model, fashion, male, female, landscape, nature and so on.

\begin{figure*}[!tb]
\centering
\includegraphics[height=0.892in]{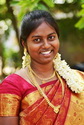}
\includegraphics[height=0.892in]{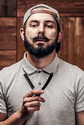}
\includegraphics[height=0.892in]{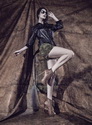}
\includegraphics[height=0.892in]{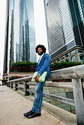}
\includegraphics[height=0.892in]{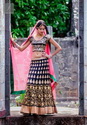}
\includegraphics[height=0.892in]{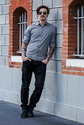}
\includegraphics[height=0.892in]{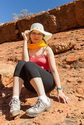}
\includegraphics[height=0.892in]{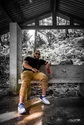}
\includegraphics[height=0.892in]{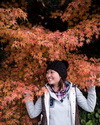}
\includegraphics[height=0.892in]{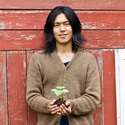}
\vspace{0.8mm} \\
\includegraphics[height=0.83in]{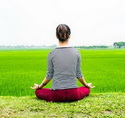}
\includegraphics[height=0.83in]{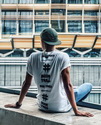}
\includegraphics[height=0.83in]{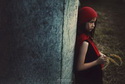}
\includegraphics[height=0.83in]{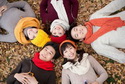}
\includegraphics[height=0.83in]{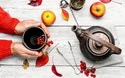}
\includegraphics[height=0.83in]{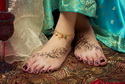}
\vspace{0.8mm} \\
\includegraphics[height=0.785in]{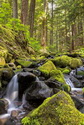}
\includegraphics[height=0.785in]{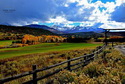}
\includegraphics[height=0.785in]{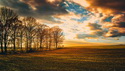}
\includegraphics[height=0.785in]{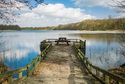}
\includegraphics[height=0.785in]{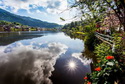}
\includegraphics[height=0.785in]{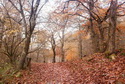}
\\
\caption{Sample images from the collected dataset.  First row: facial, full-body, seated, and upper-body. Second row: no-face, side-view, group, hand-only, and leg-only. Third row: nature, plain, water, sky, and trees.}
\label{fig:dataset_sample}
\end{figure*}

Nearly half a million images for the current dataset have been collected. The dataset has diverse photography ideas specially for the aforementioned portrait categories (full body, upper body, facial, group, couple or any two-body, side-view, hand-only, and leg-only) and landscape categories (nature, urban, etc.) from highly-rated images taken by mostly photography enthusiasts and professionals. Figure \ref{fig:dataset_sample} illustrates some sample images from the dataset including portrait categories such as facial, full-body, seated, upper-body, no-face, side-view, group, hand-only and leg-only as well as landscape categories such as nature, plain, water, sky, and trees.

Figure~\ref{fig:dataset_properties} in Section~\ref{sec:exp_dataset} shows the logarithmic distribution of the view counts, the distribution of the ratings, the logarithmic distribution of the vote counts, and the logarithmic distribution of the favorite counts of the dataset respectively. The probability of a bin represents the frequency of the images which reside in the interval starting from the current bin threshold to the next bin threshold divided by the total number of images. As a result, more than 90\% of the images were viewed more than 100, and nearly half of the images in the dataset had a rating between 40 to 50, which is a high rating.

\subsection{Automating Dataset Annotation}
The number of images in the dataset (about half a million by the end of 2017) is large. While we have manually annotated around 10\% of the dataset for training, verification, and testing purposes, to annotate the rest, we leverage multiple highly-accurate detectors to automate and accelerate the process. However, the accuracy of the annotation is not perfect, but it is high enough for getting feedback by our aesthetics-based image retrieval from the dataset. Also, the redundancy across our designed detectors makes the annotation process more accurate that we will discuss later in Section \ref{sec:decomp}.

To automate image categorization, we formulate the problem as a multi-class model for support vector machines (SVM). Therefore, we train an SVM model using radial basis function (RBF or Gaussian) kernel to predict image category (e.g. facial, upper-body, full-body, urban, nature, etc.) from multiple classes. Pose features (later explained in detail) are extracted to train our SVM model on a random subset of images from the dataset including 20K diverse images which are manually labeled. Figure~\ref{fig:catdist} depicts the distribution of the categories with respect to the number of corresponding images in each category divided by the total number of images. Consequently, the number of images for some categories like full-body, upper-body, facial, group, two, and side-view is higher than the others to cover the diversity of our image retrieval.

Instead of directly labeling photography ideas, we detect all detectable semantic classes in the scene using the object detectors and the scene parsers. In fact, we believe the scene snapshot captured by camera consists of various static and dynamic objects that constructs the constellation of the scene. The object detector partitions each shot into several boundaries (not necessarily segments) with a detection probability for each, while these boundaries can also have overlapping region. The scene parser predicts the potential objects available in each shot in pixel level, {\it i.e.} each pixel has an object label detectable by scene parser.

We enhanced the deep-learned model of object detector YOLO~\citep{redmon2016you} and scene parser PSPNet~\citep{zhao2017pyramid} to annotate the dataset. To improve the accuracy, we have trained our purpose-driven architecture of the object detector on an extended dataset including a subset of common failure cases (CFC) from the dataset with MSCOCO dataset~\citep{lin2014microsoft}. Similarly, we have trained a customized architecture of the scene parser on CFC with ADE20K dataset~\cite{zhou2017scene} as an augmented training set.

After getting all automatized annotations of the images in the dataset, we just keep those detected objects having an area greater than the 1.15\% of the image area. The probabilities of the highly-repeated semantic classes in the dataset ({\it i.e.} the frequency of the semantic class divided by the total number of images) are shown in Figure~\ref{fig:obj_dist}, while we have removed ``person'' (probability=0.9) and ``wall'' (probability=0.78) from the figure because they are dominant semantic classes in most of the images. Definitely having diverse semantic classes with high frequency in the dataset makes the proposed recommendations with respect to the query shot more helpful. After collecting the dataset, filtering unrelated images including low-quality or nudity, and auto-annotating them, we start indexing to extract the aesthetics-related information from them to accelerate the retrieval process.

\section{Photo Decomposition}
\label{sec:decomp}
To suggest a better composed photo to the user, we decompose query image from camera ({\it i.e. shot} into composition ingredients called aesthetics-related information. This information includes high-level features (such as semantic classes, photography categories, human poses, subject gender, photo tags, and photo rating) as well as low-level features (such as color, texture, and etc). To accelerate the retrieval process from the dataset based on query image, we perform the decomposition procedure on all images in the dataset as an offline process, called {\it indexing}, shown as black arrows in Figure \ref{fig:method}. We construct the composition model (CM) after indexing the whole dataset. If new images join the dataset, we index them and update our CM. In the {\it searching} step shown as red arrows in Figure \ref{fig:method}, we decompose query image, and compare with our CM. Then, we retrieve the highly-ranked photos from the dataset based on the decomposed values of the shot and user-specified preferences (USP).

Through this section, we describe our integrated object detector (IOD) to determine semantic classes in query image more comprehensively and more accurately than a single object detector. Also, our category detector (CaDe) specifies the photography genre and style. Furthermore, our artistic pose clustering (ArPose) extracts pose information specially for portrait photography. The other properties such as rating, tags, and gender in the shot are extracted from the image descriptor as a JSON file. For the low-level features, we collect all 4096 generic descriptors via public pre-trained CNN model~\citep{chatfield2014return} on ImageNet~\citep{deng2009imagenet} and the conventional features of
~\citet{mitro2016content}'s method as shown in the following equation. Note that there is no limit to collect any other aesthetics-related information from query image to extend our work depending on image style or functionality. 
\begin{eqnarray}
F_{I,vgg}=\left[f^{vgg}_{I,1} \;\; f^{vgg}_{I,2} \;\; ... \;\; f^{vgg}_{I,4096}\right]^{T},
\end{eqnarray}
where $F_{I,vgg}$ is a vector containing generic features of image $I$, and $f^{vgg}_{I,i}\;\forall i$ is $i$-th generic feature. The superscript ``$T$'' represents the transpose of the vector/matrix. Also, we extract available statistical data via the image properties including rating, view counts, and gender. Then, we similarly have them as follows: 
\begin{eqnarray}
&&\!\!\!\!F_{I,stat}=\left[f^{rating}_{I,1} \;\; f^{views}_{I,2}\right]^{T}, \\
&&\!\!\!\!F_{I,gender}=\left[f^{male}_{I,1} \;\; f^{female}_{I,2} \;\; f^{unknown}_{I,3}\right]^{T},
\end{eqnarray}
where $F_{I,stat}$ is a vector containing the statistical data of image $I$ including its rating $f^{rating}_{I,1}$ and its view counts $f^{views}_{I,2}$. Furthermore, $F_{I,gender}$ is a vector containing the gender specification of image $I$ represented by $[1\;0\;0]$ as male, $[0\;1\;0]$ as female, or $[0\;0\;1]$ as unknown.

\subsection{Integrated Object Detectors (IOD)}
Deep learning based models help computer vision researchers map from an unbounded correlated data ({\it e.g.} an image) to a bounded classified range ({\it object labels}), but there are many restrictions to exploit them for applied problems. As mentioned before, there is {\it no limit} to innovation in visual arts. Hence, it is very difficult if not impossible for available deep learning architectures to learn all of these correlated ideas and classify based on the input query with high accuracy. As the number of ideas increases, mean average precision (MAP) falls abruptly at the rate of $O\big(\frac{1}{n}\big)$. Also, manual idea labeling of a large dataset is costly in terms of computational time and available budget \citep{farhat2017carma,farhat2016stochastic,tootaghaj2017cage,farhat2016towards}.

To tackle the problem of classifying to a large number of ideas, we detect as many components as possible in the scene instead of photography ideas. In fact, we believe the scene captured in viewfinder consists of various static and dynamic objects as its high-level features. We improve the detection accuracy by training our customized object detector on an augmented dataset including a subset of common failure cases (CFC) from the 500px dataset with MSCOCO dataset~\citep{lin2014microsoft}. We train our scene parser on our CFC with ADE20K dataset~\citep{zhou2017scene} as an extended training dataset, and also our human pose estimator is trained on our CFC with MSCOCO dataset~\citep{lin2014microsoft} and MPII dataset~\citep{andriluka14cvpr}. 

We start from state-of-the-art deep-learned detectors, YOLO~\citep{redmon2016you}, PSPNet~\citep{zhao2017pyramid} and RTMPPE~\citep{cao2017realtime}, and we extend, improve and integrate them for our purpose. YOLO network partitions the query photo into several bounding boxes predicting their probabilities. Pyramid scene parsing network (PSPNet) uses global context information through a pyramid pooling module, and predicts the scene objects in the pixel level. Real-time multi-person 2D pose estimation (RTMPPE) predicts vector fields to represent the associative locations of the anatomical parts by means of two sequential prediction process exposing the part confidence maps and the vector fields. 

\begin{figure}[!tb]
\centering
\includegraphics[height=0.52in]{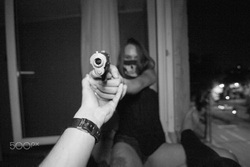}
\includegraphics[height=0.52in]{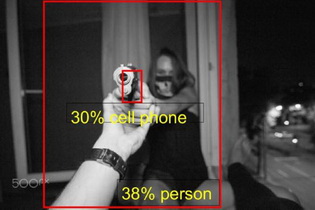}
\includegraphics[height=0.52in]{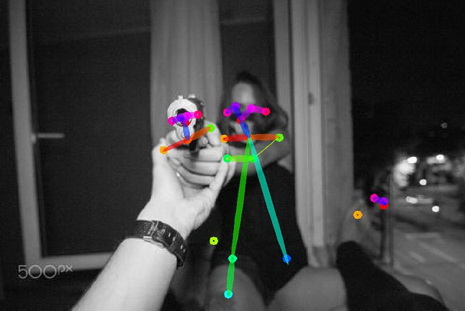}
\includegraphics[height=0.52in]{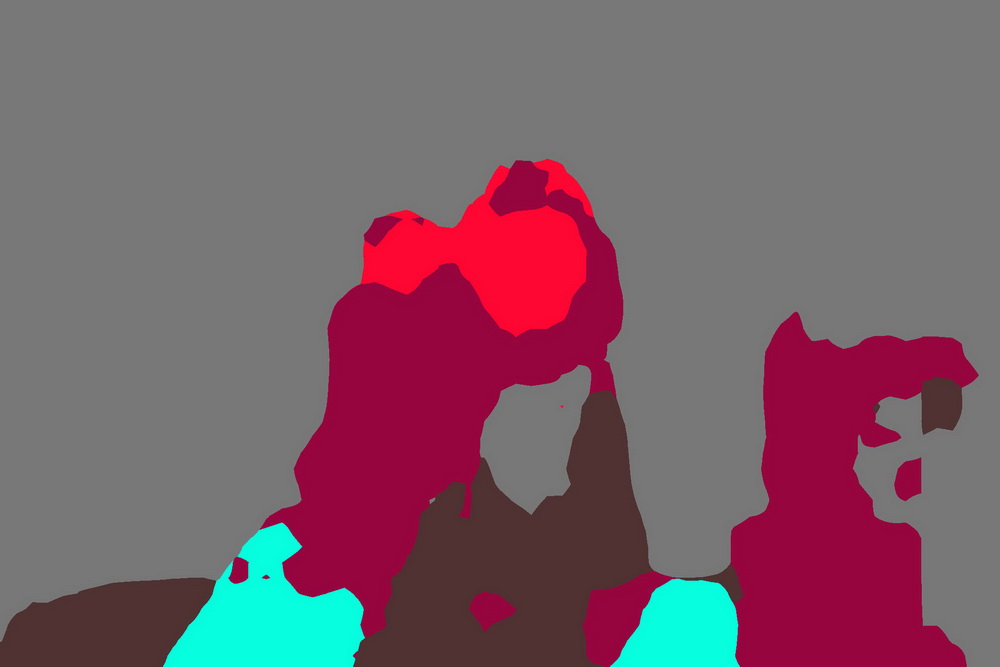}
\vspace{0.5mm} \\
\includegraphics[height=0.98in]{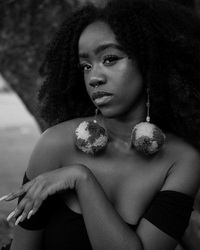}
\includegraphics[height=0.98in]{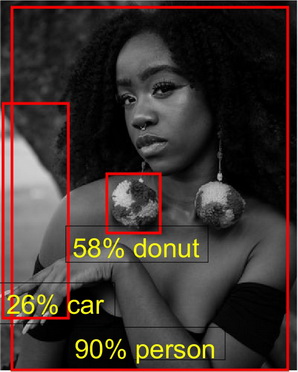}
\includegraphics[height=0.98in]{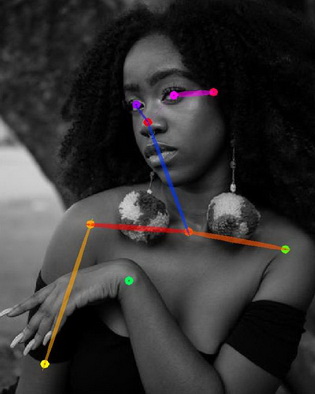}
\includegraphics[height=0.98in]{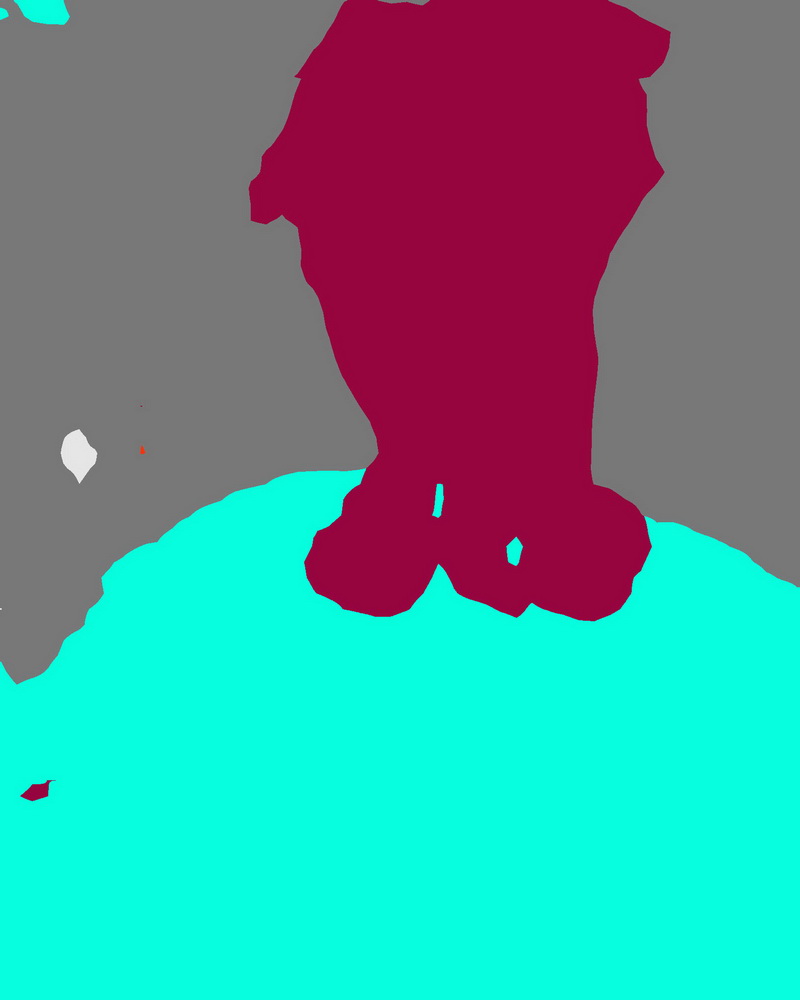}
\vspace{0.5mm} \\
\includegraphics[height=0.59in]{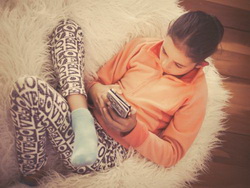}
\includegraphics[height=0.59in]{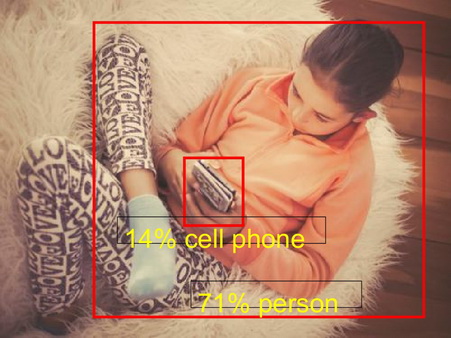}
\includegraphics[height=0.59in]{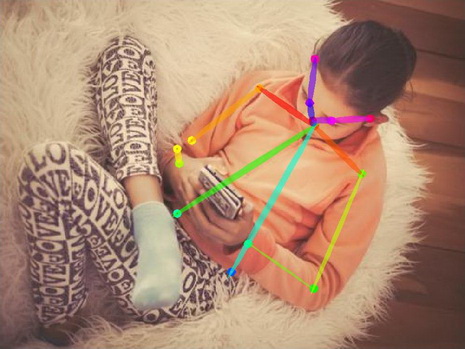}
\includegraphics[height=0.59in]{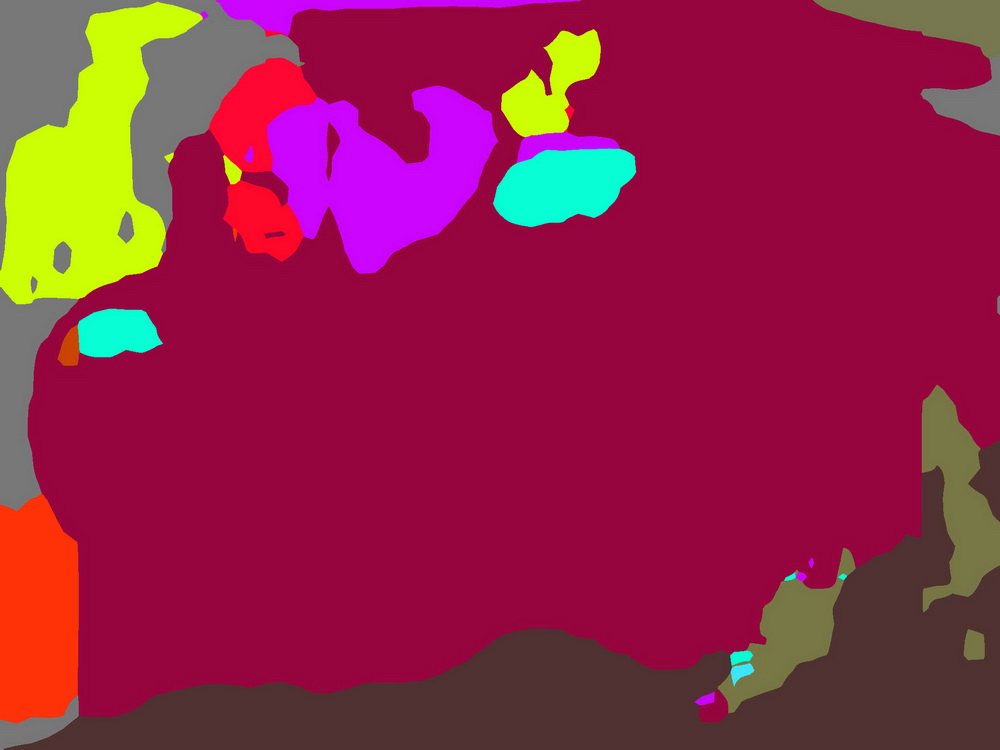}
\vspace{0.5mm} \\
\includegraphics[height=1.18in]{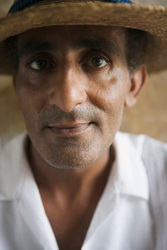}
\includegraphics[height=1.18in]{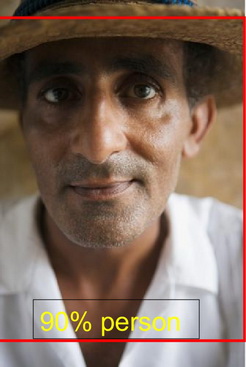}
\includegraphics[height=1.18in]{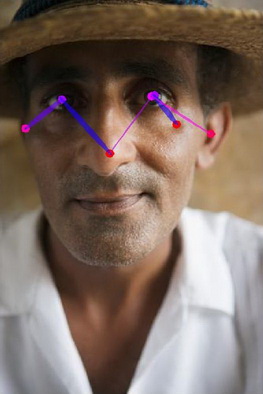}
\includegraphics[height=1.18in]{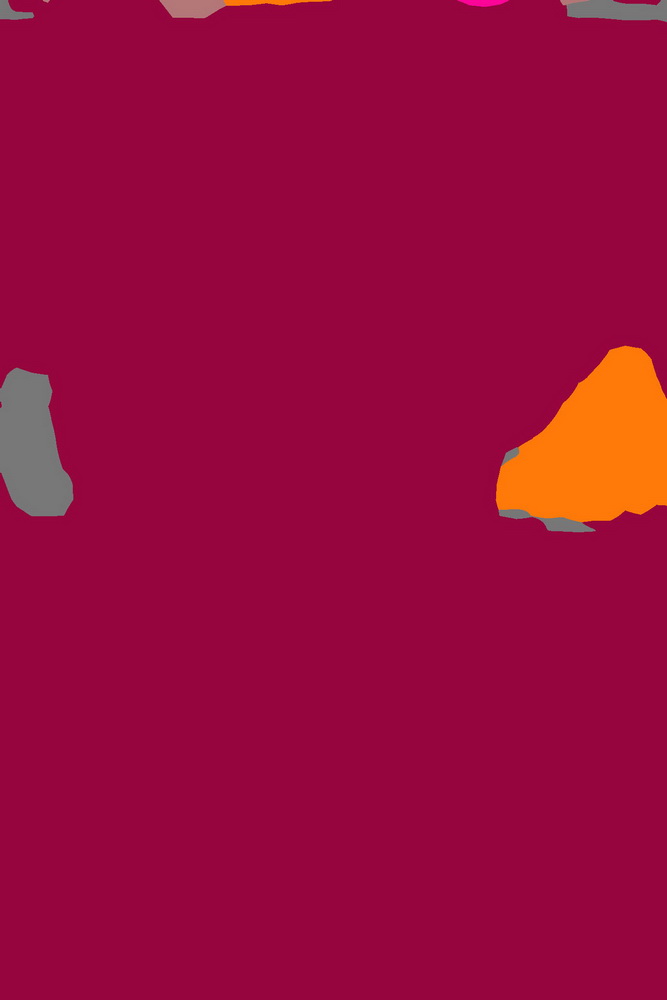}
\vspace{0.5mm} \\
\caption{Sample failed cases generated by detectors developed in the computer vision field. In each row, the images are the original, the YOLO result, the RTMPPE result, and the PSPNet result, respectively.}
\label{fig:decomp_iod_cfc}
\end{figure}

Figure~\ref{fig:decomp_iod_cfc} illustrates a small subset of common failure cases (CFC) across object detector (YOLO), human pose estimator (RTMPPE), and scene parser (PSPNet). Occasionally RTMPPE misses at facial photos to detect human parts like neck in close-up photos, and it is not very accurate at ``two'' or ``group'' categories to associate parts overlapping. The non-person detection of YOLO under 30\% probability is sometimes not reliable. PSPNet detection is partially not accurate enough at photos with many objects, as it partitions the photo into small chunks and it never considers overlapped area. Generally, to improve the accuracy of the detectors, we have changed the transformation parameters of the architecture such as maximum rotation, crop size, scale min and max. Because higher rotation and bigger portrait are more important in our work. 

For comparison with YOLO, we use the regular MAP on all intended objects. Table~\ref{tab:yolo} in Section~\ref{sec:exp_od} shows the MAP and the average accuracies of some objects by our trained model versus pre-trained YOLO model. For comparison with RTMPPE, we measure MAP of all body parts (left and right are combined) as mentioned in DeeperCut~\citep{insafutdinov2016deepercut}. Table~\ref{tab:rtmppe} in Section~\ref{sec:exp_pe} compares MAP performance between ours and RTMPPE on a subset of testing images randomly selected from the 500px dataset. For scene parsing evaluation, we measure pixel-wise accuracy (PixAcc) and mean of class-wise intersection over union (CIoU), where the performance values of our trained scene parser are 78.6\% PixAcc and 42.5\% CIoU better than PSPNet with 101-depth ResNet (74.9\% PixAcc and 40.8\% CIoU) listed in Table~\ref{tab:pspnet} of Section~\ref{sec:exp_sp}.

\begin{figure}[!tb]
\centering
\includegraphics[height=1.17in]{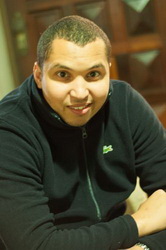}
\includegraphics[height=1.17in]{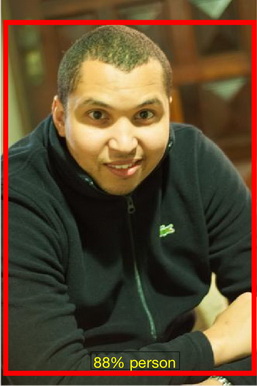}
\includegraphics[height=1.17in]{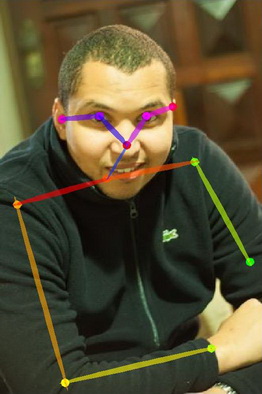}
\includegraphics[height=1.17in]{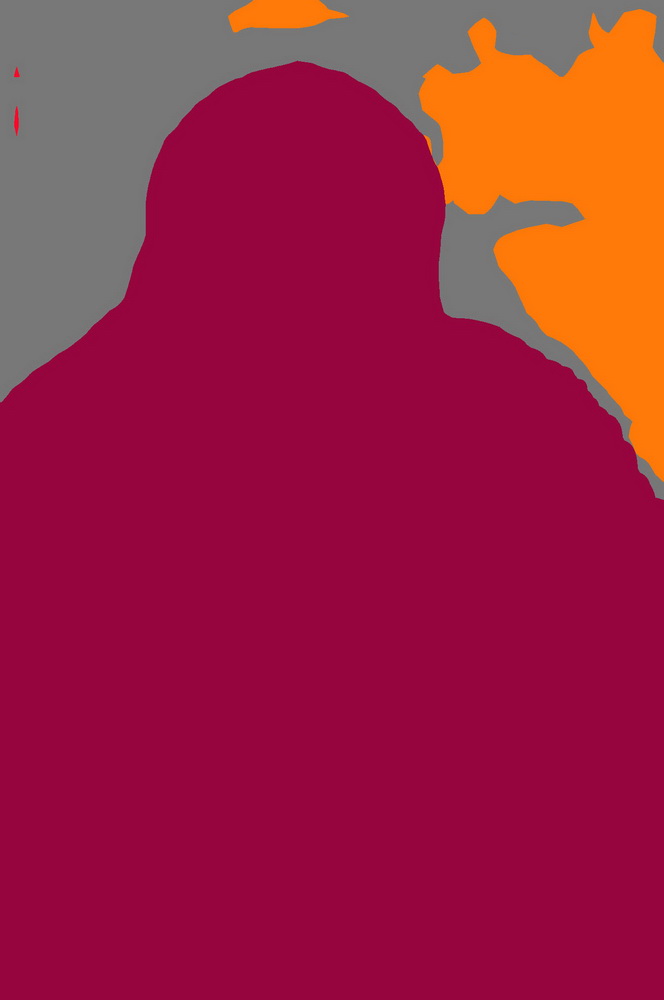}
\vspace{0.5mm} \\
\includegraphics[height=0.92in]{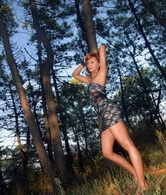}
\includegraphics[height=0.92in]{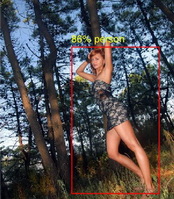}
\includegraphics[height=0.92in]{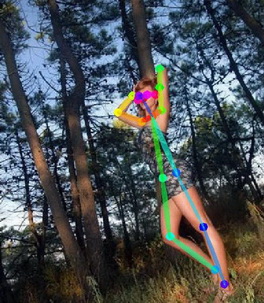}
\includegraphics[height=0.92in]{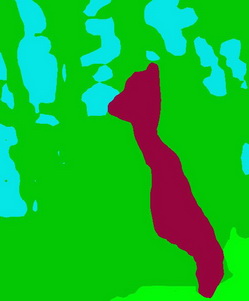}
\vspace{0.5mm} \\
\includegraphics[height=1.17in]{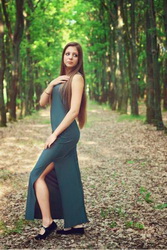}
\includegraphics[height=1.17in]{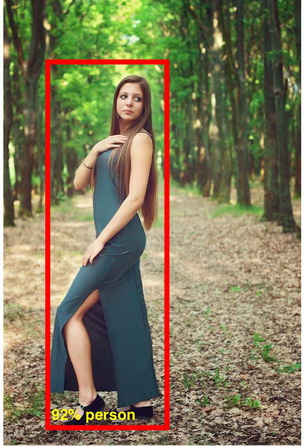}
\includegraphics[height=1.17in]{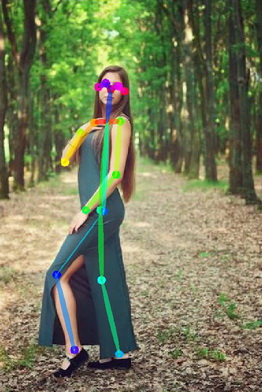}
\includegraphics[height=1.17in]{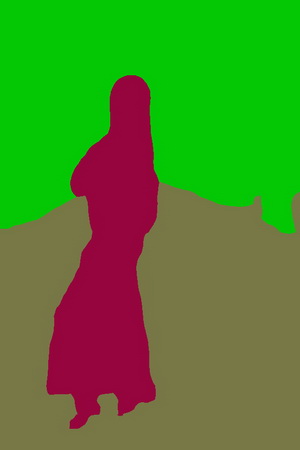}
\vspace{0.5mm} \\
\includegraphics[height=0.525in]{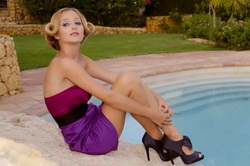}
\includegraphics[height=0.525in]{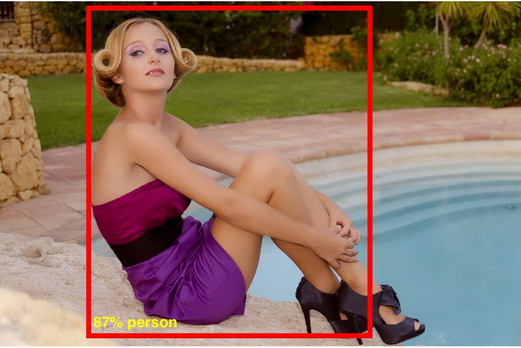}
\includegraphics[height=0.525in]{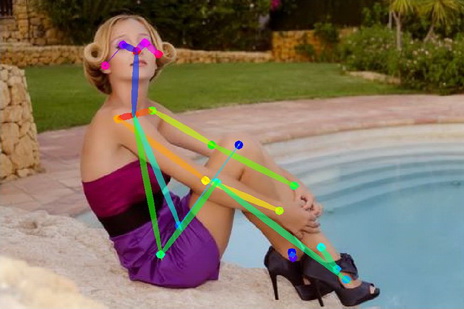}
\includegraphics[height=0.525in]{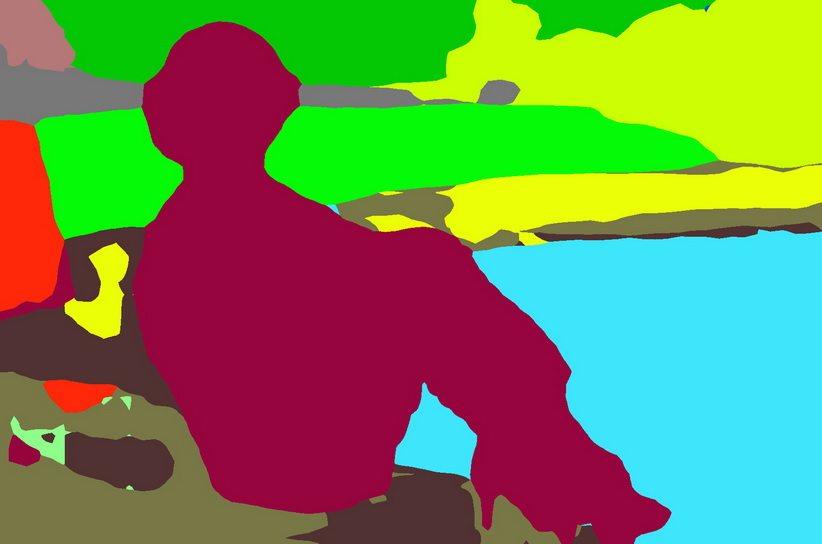}
\vspace{0.5mm} \\
\caption{Qualitative results generated by the enhanced integrated object detector (IOD) show refined samples after training. In each row, the images are the original, the object detector result, the pose estimator result, and the scene parser result, respectively.}
\label{fig:decomp_iod}
\end{figure}

Figure~\ref{fig:decomp_iod} illustrates four different qualitative results, where YOLO object names are shown in a red rectangle with a probability, RTMPPE poses are shown as a colorful connection of skeleton joints, and PSPNet scenes are colorized pixel-wisely based on the pixel codename.

\subsubsection{Value Unification}
To work more conveniently on the outputs of our customized detectors in next steps, we need to unify the outputs in terms of pixel-level tensors. Our object detector outputs MSCOCO object-IDs among 80 categories (from 1 to 80). We define their scores as the minus logarithm of their NOT probability ($-\log\left(1-p\right)$) for each pixel of the image. The object-ID and its score for each pixel is represented as a $mxnx2$ tensor. Also, our scene parser outputs ADE20K object-IDs among 150 categories (from 1 to 150), and the object-ID with its score for each pixel of the image is represented as a tensor. Similarly, our human pose estimator gives 18 anatomical part IDs with their scores as a tensor. Thus, for any image ($I_{m\times{n}}$) we have:
\begin{eqnarray}
&&T^{I,od}_{m\times{n}\times{2}}=\left[t^{I,od}_{i,j,k}\right],\\
&&t^{I,od}_{i,j,1}=C^{I,id}_{i,j}\;,t^{I,od}_{i,j,2}=-\log_2{(1-p^{I,od}_{i,j})},\nonumber
\label{eq:T_od}
\end{eqnarray}
\begin{eqnarray}
&&T^{I,sp}_{m\times{n}\times{2}}=\left[t^{I,sp}_{i,j,k}\right],\\
&&t^{I,sp}_{i,j,1}=A^{I,id}_{i,j}\;,t^{I,sp}_{i,j,2}=-\log_2{(1-p^{I,sp}_{i,j})},\nonumber
\label{eq:T_sp}
\end{eqnarray}
\begin{eqnarray}
&&T^{I,pe}_{m\times{n}\times{2}}=\left[t^{I,pe}_{i,j,k}\right],\\
&&t^{I,pe}_{i,j,1}=J^{I,id}_{i,j}\;,t^{I,pe}_{i,j,2}=-\log_2{(1-p^{I,pe}_{i,j})}\;, \nonumber
\label{eq:T_pe}
\end{eqnarray}
where $I$ is an input image, $m$ is the number of rows, $n$ is the number of columns in the image, $T^{I,od}$ is corresponding tensor of object detector ({\it e.g.} YOLO), $C^{I,id}_{i,j}\in{\{1..80\}}$ is MSCOCO ID of the pixel at $(i,j)$, $p^{I,od}_{i,j}$ is the MSCOCO ID probability of the pixel at $(i,j)$, $T^{I,sp}$ is tensor of scene parser ({\it e.g.} PSPNet), $A^{I,id}_{i,j}\in{\{1..150\}}$ is ADE20K ID of the pixel at $(i,j)$, $p^{I,sp}_{i,j}$ is the ADE20K ID probability of the pixel at $(i,j)$, $T^{I,pe}$ is tensor of pose estimator ({\it e.g.} RTMPPE), $J^{I,id}_{i,j}\in{\{1..18\}}$ is the joint ID of the pixel at $(i,j)$, and $p^{I,pe}_{i,j}$ is the joint ID probability of the pixel at $(i,j)$.

To auto-tag or auto-label the 500px dataset in indexing step, we combine these unified results in terms of the semantic classes, their coordinates, and their scores (or probabilities). The number of the detectable classes is 210 semantic objects by merging MSCOCO (80 categories) and ADE20K (150 categories) objects and de-duplicating 20 semantic objects (such as person and sky). Also we have 18 joints from RTMPPE including nose, neck, right shoulder, right elbow, right wrist, left shoulder, left elbow, left wrist, right hip, right knee, right ankle, left hip, left knee, left ankle, left eye, right eye, left ear, and right ear. YOLO detection for a small full-body person in the image is poor, but it detects big limbs of the body (as a person label) well. RTMPPE detection for occluded bodies is poor but the detection for a full-body person is acceptable. Also, PSPNet detection for objects, not a person, is relatively good compared to others. 

\subsubsection{Hysteresis Detection}
To expand our framework coverage, using our available detectors, we detect the potential objects in the image. Our detector's integration scheme has LOW (usually with the probability less than 0.09) and HIGH (usually with the probability higher than 0.44) thresholds for each binary (object,detector). These thresholds are tuned by a random set of highly-rated ground-truth images. If the average probability (score) of the pixels with object ID X in the image is higher than its HIGH threshold, there is an object ID X in the image, otherwise if the average probability (score) of the pixels with object ID X in the image is lower than the corresponding LOW threshold, there is no object ID X in the image, and we examine other objects for indexing purpose or another image for searching purpose. We call our detector's integration scheme as \textit{hysteresis} detection.

\begin{figure}[!tb]
\centering
\includegraphics[width=1\linewidth]{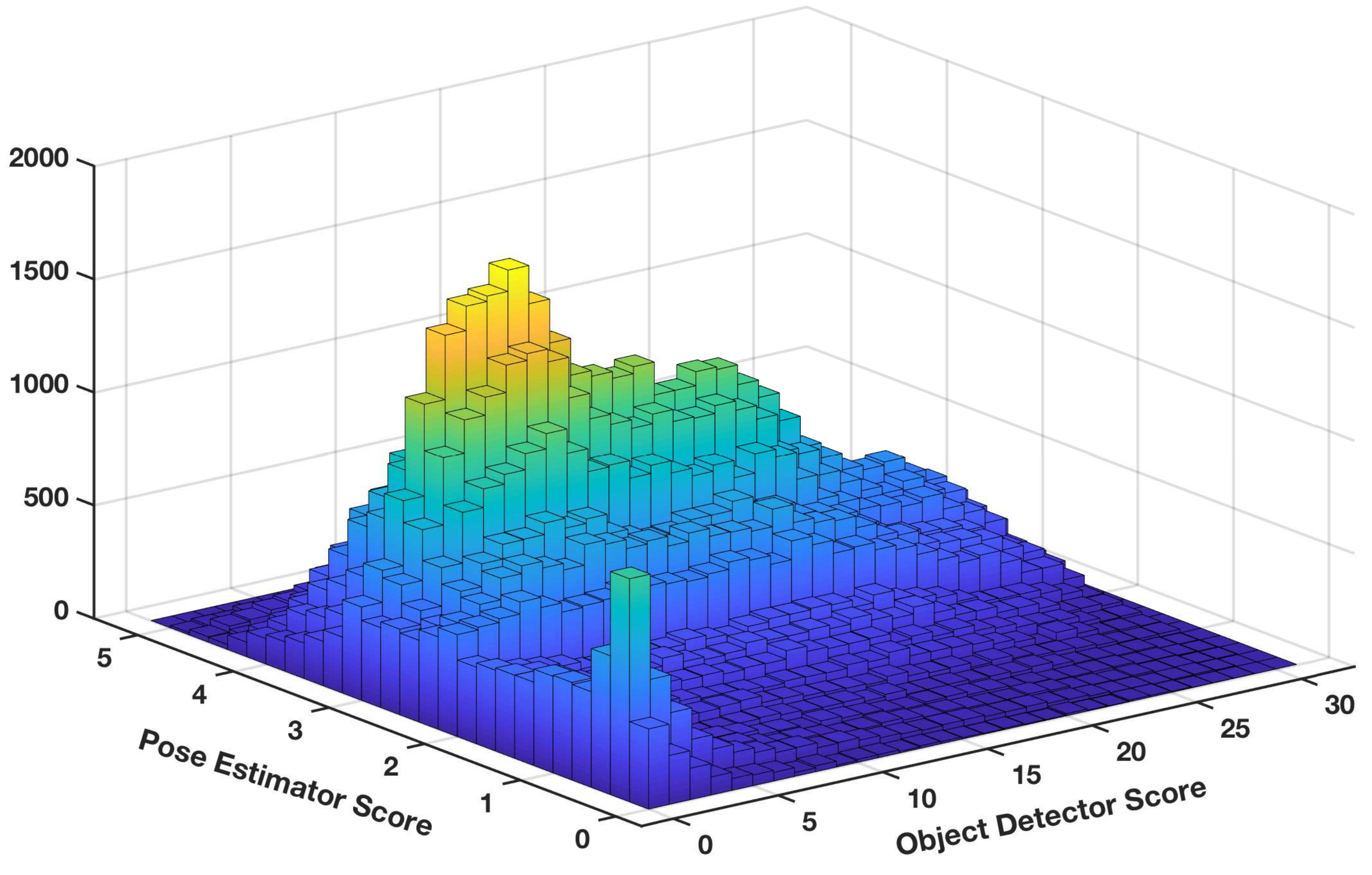}\\
\caption{The 2D histogram of the portrait images binned by the object detector and pose estimator scores.}
\label{fig:persondist}
\end{figure}

Hysteresis detection guarantees that confidence ratio of existence of an object in an image is at least higher than a tuned detection threshold for one of the detectors. A person may be detected by one detector but not by another one. Therefore, if we do not want to miss any shared detectable object, we consider the union of the detectors, \textit{i.e.}, if any detector outputs the object with HIGH probability, the object is considered inside the image. For example, if we use the hysteresis detection of the 500px dataset for ``person'' object, 90.5\% (400K+) of the images pass. The 2D histogram of the portrait dataset in Figure~\ref{fig:persondist} illustrates the frequency of the images smart-binned by the normalized object detector and pose estimator scores. In fact, it shows the effectiveness of integrating these detectors to use the coverage of the dataset more precisely, because we are unifying the detection results for a more broad range of ideas rather than intersecting them to have a more confident narrow range of ideas.

To tune the boundaries, we conduct the experiments and consider detection probability and normalized area as two features of a dominant object from the object detector, and detection score and normalized area as two features of a dominant object from the pose estimator. From the 2D ROC results in Figure~\ref{fig:person_analysis}, we infer that $probability=45\%$ for the object detector and $area=10\%$ for the pose estimator are the optimized cut-off thresholds to decide about the existence of a person in the image.

\subsubsection{Object Importance}
To prioritize the prominence of the objects in the image, we seek to use the importance map of the objects, because the subject of the image should be more important even if its detection probability is lower. To rank the order of the objects, we exploit the max score multiply by a saliency map (S) features with our centric distance (D) feature to get our weighted saliency map (W).
\begin{eqnarray}
W^I(i,j) &=& \max{\left(T^{I,od}_{*,*,2},T^{I,sp}_{*,*,2}\right)}_H{S^I(i,j){D^I(i,j)}}\;,
\label{eq:weight} \\
D^I(i,j) &=& \frac{1}{K}{e^{-\norm{[i,j] - c^I}_k}}\;,
\\
c^I &=& \frac{\sum_{i,j}{S^I(i,j).[i,j]}}{\sum_{i,j}{S^I(i,j)}}\;,
\end{eqnarray}
where $W^I(i,j)$ is our weighted saliency map pointwisely for image $I$, $max(.)_H$ operation is a hysteresis max on the 2nd plane of the tensors (score matrix), $S^I(i,j)$ is the saliency map of image $I$ inspired by~\citet{itti1998model}, and $D^I(i,j)$ is our centric distance feature of image $I$, $K$ is a tunable constant equal to $\sum_{i,j}{e^{-\norm{[i,j] - c^I}_k}}$ for image $I$, the binary value $c^I$ is the center of the mass coordinate, and $\norm{.}_k$ is the $k$-th norm operator where $k=1$ in our experiments. 

Our weighted saliency map makes the detected objects prioritized, because we sum up the scores from the semantic classes, and we end up with a total score for each semantic class. The output of this step is a weighted vector of detected semantic classes (undetected object has zero weight) in the query image. We show it as the following vector where the elements represent the importance (normalized as a probability) of the corresponding object in the image:

\begin{eqnarray}
F_{I,iod}&=&\left[f^{imp}_{1}\;\;f^{imp}_{2}\;\;...\;\;f^{imp}_{210}\right]\;, \\
f^{imp}_{k}&=&\frac{\sum_{\forall objID(i,j)=k} {W(i,j)}}{\sum_{\forall i,j} {W(i,j)}}\;, \\
\forall k &\in& \{1,2,...,210\}, \nonumber
\label{iod_feat}
\end{eqnarray}
where $f^{imp}_{k}$ is the importance ($imp$) value of $k$-th object which is the summation of the weighted saliency of the pixel $(i,j)$ with ID $k$, {\it i.e.}, $objID(i,j)=k$, and $F_{I,iod}$ is the importance vector of all objects. As we index the 500px dataset by integrated object detectors, the union of the objects by the detectors is recognized. The distribution of the highly-repeated semantic classes in the 500px dataset is shown in Figure~\ref{fig:obj_dist} in Section~\ref{sec:exp_iod}, where the ``person'' and ``wall'' were removed from the figure as mentioned before. 

\subsection{Category Detector (CaDe)}
\label{sec:cade}
The photo categories in portrait include two (couple or two people), group (more than two people), full-body, upper-body, facial, side-view, faceless, headless, hand-only, and leg-only, which are ten classes. In landscape photography, there are sea, mountain, forest, cloud, and urban, which are five classes. While we focus on portrait and landscape photography genres, we believe that this work can be extended to other genres as well.

Knowing the photo genres and categories help our framework guide the photographer more adequately because it retrieves better-related results based on the photographer preferences. The downside can be the low coverage or a limited number of contents on the leaves of this hierarchical tree of the photo styles, but the comprehensive dataset addresses this potential issue.

\subsubsection{Top-down Hierarchical Clustering}
To distinguish a portrait from a landscape photo, the number of people in the image is estimated by the max (union) number of person-IDs higher than their corresponding HIGH thresholds across the detectors in integrated object detector (IOD). If the score for detecting a person is lower than a LOW threshold for all detectors in IOD, there is no person in the image. Then, if there is a water-like, mountain-like, plant-like, cloud-like, or building-like object in the image with total area higher than 26.5\% (optimized for landscape), the landscape category will be recognized as well. Otherwise, if there is a person in the image, detecting the right category by a decision tree is one heuristic approach, {\it i.e.}, if the image contains a nose, two eyes, a hand and a leg OR a nose, an eye, two hands and two legs, it will be categorized as full body. Such combinations are determined after trying tens of random images as ground truth, because pose estimator model is not perfect and, in some cases, the limbs are occluded. After examining full-body, if the image contains a nose and two eyes and one hand, it will be categorized as upper-body. But, we do not follow this approach, because this hierarchical approach for portrait images is not very accurate, as some leaves of its decision tree have some correlations like full-body and group categories. Also, the coverage is not fair, since upper levels like full-body attract most of the photos, and the rest will remain for the lower levels.

\subsubsection{Portrait Multi-class Categorization}
Our more efficient and accurate approach to automate portrait categorization formulates the problem as a multi-class model for support vector machines (MCMSVM). The inputs are our feature vectors and the corresponding class labels, and the trained MCMSVM is a fully trained multi-class error-correcting output codes (ECOC) model, while we are using 10 portrait categories or unique class labels, it needs 45 ($=10(10-1)/2$) binary SVM models with radial basis function (RBF or Gaussian) as its kernel and a one-vs-one coding design. We have annotated 5\% (about 25K+) of portrait photos uniformly selected at random from the dataset as the ground truth of the portrait categories. Then, we train an MCMSVM with the feature vectors and the corresponding labels of 80\% (about 20K) of our ground truth and leave the rest for testing our MCMSVM. Our feature vector for each photo includes 40 different features as follows:
\begin{itemize}
\item General MAX: (1,2) max scores for detected people from IOD, (3,4) max areas for the detected people from IOD.
\item Intersected Area: (5) intersected area between highly probable people from IOD, (6,7) scores of the highly probable people for each detector in IOD, (8,9) areas of the highly probable people for each detector in IOD.
\item Number of people: (10,11) number of people higher than HIGH threshold for each detector in IOD, (12,13) number of people with area higher than 5\% for each detector in IOD, (14) max number of people by score from IOD, (15) max number of people by area from IOD, (16) max of (14) and (15).
\item Limb Features: (from 17 to 40) the limbs respectively including nose, neck, right shoulder, right elbow, right wrist, right hand, left shoulder, left elbow, left wrist, left hand, right hip, right knee, right ankle, right leg, left hip, left knee, left ankle, left leg, right eye, left eye, eyes, right ear, left ear, ears which add up to 40 features.
\end{itemize}

The output of this step for an image query is the following unitary vector that shows its category (facial, full-body, upper-body, two, group, side-view, leg, no-face, hand, and no-head) as:
\begin{multline}
F_{I,cade} = [ f^{facial}_{1} \;\; f^{fullbody}_{2} \;\; f^{upperbody}_{3} \;\; f^{two}_{4} \;\; f^{group}_{5} \\
  f^{sideview}_{6} \;\; f^{leg}_{7} \;\; f^{noface}_{8} \;\; f^{hand}_{9} \;\; f^{nohead}_{10} ],
\label{cade_feat}
\end{multline}
where $F_{I,cade}$ shows the unitary category vector of the image $I$ by CaDe detector, and only one of the vector element is one and the rest are zero. 

The mean average accuracy of our category detection is shown in Table~\ref{tab:decomp_cade} in Section~\ref{sec:exp_cade} for the dataset images divided by various styles. The CaDe indexing of the dataset results the distribution of the portrait categories shown in Figure~\ref{fig:catdist} in Section~\ref{sec:exp_cade}. Consequently, the number of photos in the categories containing full-body, upper-body, facial, group, two, and side-view is adequate.

\subsection{Artistic Pose Clustering (ArPose)}
\label{sec:arpose}
\textit{Posing}, one of the essential ingredients of the portrait photography, could substantially differentiate between amateur and professional shots. Having little experience in portrait photography, finding correct postures or coming up with novel poses is hard for amateur photographers. Hence, it is vital for our system to have an understanding of different poses and how to categorize them. Recently, there have been numerous efforts in the computer vision field for human pose estimation of images and videos. With the rise of deep learning models, these approaches are getting more accurate and more robust. One of the state-of-the-art algorithms for pose estimation is RTMPPE~\citep{cao2017realtime}, where they use VGG features as an input and then exploit a two-stage CNN to find the probability of joints and their connections together. Their architecture predicts vector fields to represent the associative locations of the anatomical parts via two sequential prediction process exposing the part confidence maps and the vector fields on MSCOCO~\citep{lin2014microsoft} and MPII~\citep{andriluka14cvpr} datasets. Although RTMPPE extract body joints in images, these joints are merely considered as our features for pose detection. Hence, we use two sets of features on top of RTMPPE in order to define the distance between different poses. These set of features are scale invariant, thus regardless of the scale of the human body in images, we measure the similarity of two poses. These features are defined as follows:
\begin{itemize}
\item \textbf{Joint to Line Distance (J2L)}:
~\citet{li2017skeleton} apply this distance in their action recognition system from body joints skeleton. They capture the distance of each joint from any line that connects two other joints. To have the scale invariant distance, we normalize these distances with the maximum J2L distance in each body in the picture. Having the joint $j_l$ and the line crossing two joints, $j_m$ and $j_n$, Joint to Line Distance is calculated as follows:
\begin{equation}
J2L(l,m,n) = 2S_{\Delta_{lmn}}/||j_m - j_n||_2, 
\label{ap_feat}
\end{equation}
where $S_{\Delta_{lmn}}$ is the area under the triangle formed by three joints. 
Based on the total number of joints in each body, which is $18$, and the total number of different distances is $18 \times \binom{17}{2} = 2448$.
\item \textbf{Skeleton Context (SC)}: 
~\citet{kamani2016shape,kamani2017skeleton} introduce a scale invariant feature applied to a skeleton matching task. Skeleton context is a polar histogram of each point in the skeleton indicating the angular and distance distribution of other points in the skeleton around that point. We benefit from the angular distribution of each point and create an $18 \times 18$ angular matrix for each body in the image.
\end{itemize}
These features are designed to capture the relative position of each joint with respect to other points, hence, they are used as a measure of distance between different poses. Next, we use these features to cluster images based on various poses.

\begin{figure*}[!tb]
\centering
\includegraphics[height=1in]{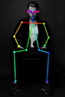}
\includegraphics[height=1in]{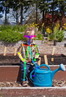}
\includegraphics[height=1in]{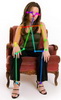}
\includegraphics[height=1in]{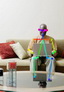}
\includegraphics[height=1in]{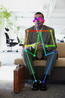}
\includegraphics[height=1in]{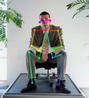}
\includegraphics[height=1in]{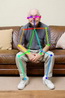}
\includegraphics[height=1in]{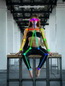}
\includegraphics[height=1in]{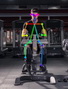}
\vspace{0.8mm} \\
\includegraphics[height=1in]{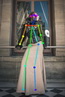}
\includegraphics[height=1in]{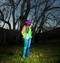}
\includegraphics[height=1in]{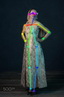}
\includegraphics[height=1in]{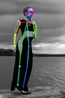}
\includegraphics[height=1in]{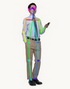}
\includegraphics[height=1in]{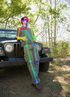}
\includegraphics[height=1in]{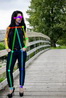}
\includegraphics[height=1in]{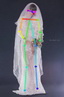}
\includegraphics[height=1in]{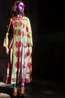}
\vspace{0.8mm} \\
\includegraphics[height=1in]{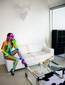}
\includegraphics[height=1in]{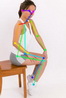}
\includegraphics[height=1in]{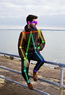}
\includegraphics[height=1in]{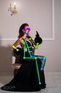}
\includegraphics[height=1in]{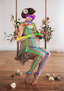}
\includegraphics[height=1in]{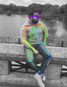}
\includegraphics[height=1in]{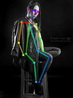}
\includegraphics[height=1in]{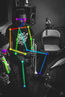}
\includegraphics[height=1in]{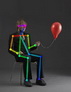}
\vspace{0.8mm} \\
\includegraphics[height=1in]{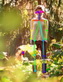}
\includegraphics[height=1in]{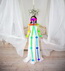}
\includegraphics[height=1in]{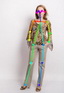}
\includegraphics[height=1in]{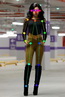}
\includegraphics[height=1in]{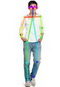}
\includegraphics[height=1in]{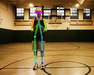}
\includegraphics[height=1in]{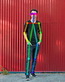}
\includegraphics[height=1in]{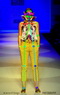}
\vspace{0.8mm} \\
\includegraphics[height=1in]{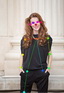}
\includegraphics[height=1in]{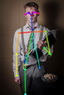}
\includegraphics[height=1in]{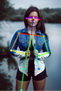}
\includegraphics[height=1in]{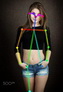}
\includegraphics[height=1in]{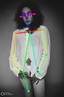}
\includegraphics[height=1in]{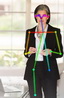}
\includegraphics[height=1in]{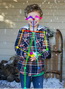}
\includegraphics[height=1in]{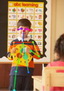}
\includegraphics[height=1in]{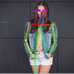}
\vspace{0.8mm} \\
\includegraphics[height=1in]{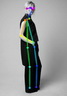}
\includegraphics[height=1in]{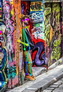}
\includegraphics[height=1in]{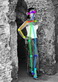}
\includegraphics[height=1in]{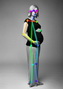}
\includegraphics[height=1in]{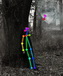}
\includegraphics[height=1in]{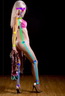}
\includegraphics[height=1in]{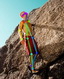}
\includegraphics[height=1in]{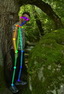}
\includegraphics[height=1in]{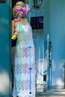}
\vspace{0.8mm} \\
\includegraphics[height=1in]{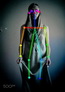}
\includegraphics[height=1in]{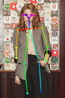}
\includegraphics[height=1in]{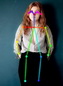}
\includegraphics[height=1in]{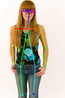}
\includegraphics[height=1in]{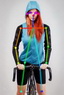}
\includegraphics[height=1in]{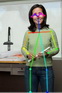}
\includegraphics[height=1in]{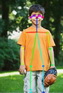}
\includegraphics[height=1in]{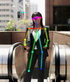}
\includegraphics[height=1in]{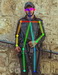}
\vspace{0.8mm} \\
\includegraphics[height=0.98in]{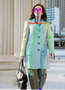}
\includegraphics[height=0.98in]{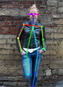}
\includegraphics[height=0.98in]{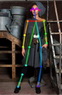}
\includegraphics[height=0.98in]{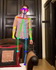}
\includegraphics[height=0.98in]{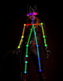}
\includegraphics[height=0.98in]{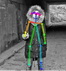}
\includegraphics[height=0.98in]{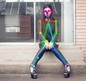}
\includegraphics[height=0.98in]{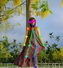}
\caption{First 8 clusters derived from our algorithm on the portrait dataset. Each row represents the top poses of each cluster fitted in a line.}
\label{fig:posekmeans12}
\end{figure*}

\begin{figure*}[!tb]
\centering
\includegraphics[height=1in]{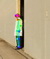}
\includegraphics[height=1in]{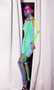}
\includegraphics[height=1in]{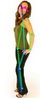}
\includegraphics[height=1in]{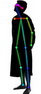}
\includegraphics[height=1in]{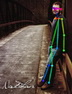}
\includegraphics[height=1in]{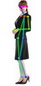}
\includegraphics[height=1in]{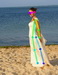}
\includegraphics[height=1in]{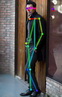}
\includegraphics[height=1in]{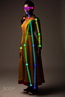}
\includegraphics[height=1in]{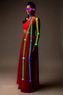}
\vspace{0.8mm} \\
\includegraphics[height=1in]{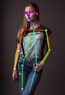}
\includegraphics[height=1in]{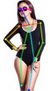}
\includegraphics[height=1in]{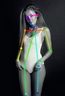}
\includegraphics[height=1in]{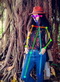}
\includegraphics[height=1in]{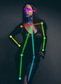}
\includegraphics[height=1in]{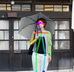}
\includegraphics[height=1in]{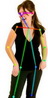}
\includegraphics[height=1in]{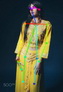}
\includegraphics[height=1in]{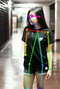}
\vspace{0.8mm} \\
\includegraphics[height=1in]{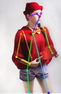}
\includegraphics[height=1in]{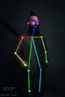}
\includegraphics[height=1in]{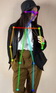}
\includegraphics[height=1in]{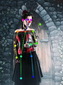}
\includegraphics[height=1in]{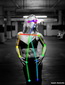}
\includegraphics[height=1in]{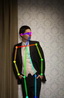}
\includegraphics[height=1in]{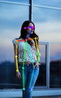}
\includegraphics[height=1in]{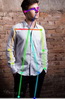}
\includegraphics[height=1in]{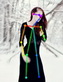}
\vspace{0.8mm} \\
\includegraphics[height=1in]{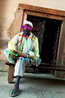}
\includegraphics[height=1in]{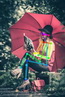}
\includegraphics[height=1in]{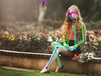}
\includegraphics[height=1in]{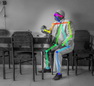}
\includegraphics[height=1in]{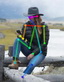}
\includegraphics[height=1in]{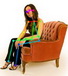}
\includegraphics[height=1in]{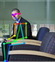}
\vspace{0.8mm} \\
\includegraphics[height=1in]{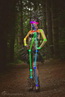}
\includegraphics[height=1in]{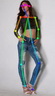}
\includegraphics[height=1in]{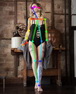}
\includegraphics[height=1in]{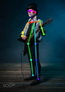}
\includegraphics[height=1in]{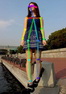}
\includegraphics[height=1in]{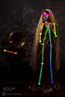}
\includegraphics[height=1in]{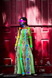}
\includegraphics[height=1in]{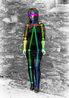}
\includegraphics[height=1in]{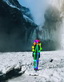}
\vspace{0.8mm} \\
\includegraphics[height=1in]{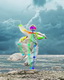}
\includegraphics[height=1in]{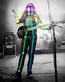}
\includegraphics[height=1in]{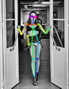}
\includegraphics[height=1in]{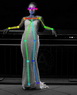}
\includegraphics[height=1in]{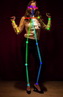}
\includegraphics[height=1in]{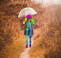}
\includegraphics[height=1in]{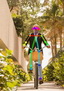}
\includegraphics[height=1in]{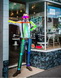}
\vspace{0.8mm} \\
\includegraphics[height=0.85in]{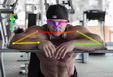}
\includegraphics[height=0.85in]{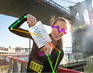}
\includegraphics[height=0.85in]{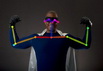}
\includegraphics[height=0.85in]{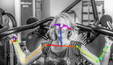}
\includegraphics[height=0.85in]{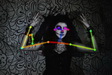}
\vspace{0.8mm} \\
\caption{The rest of 7 clusters derived from our algorithm on the portrait dataset. Each row represents the top poses of each cluster fitted in a line.}
\label{fig:posekmeans22}
\end{figure*}

\subsubsection{Pose Clustering}
To rank each body posture in images, and find the nearest professional poses to the amateur one in the query image, we use a clustering method. The clustering method should be able to distinguish between different poses and group similar ones using the features explained in Section~\ref{sec:arpose}. In order to do so, we use two clustering algorithms, Kmeans and Deep Embedding~\citep{xie2016unsupervised}. We compare the result of these two clustering on this task. In order to do the clustering, we first need to determine the number of clusters. There are several heuristic methods to estimate the optimal number of clusters for each dataset, including but not limited to {\it elbow} and {\it silhouette} methods. In this clustering task, having too many clusters would diminish the novelty and diversity of the results, in a sense that it tries to have samples as close as possible to one cluster. On the other hand, keeping the number of clusters low would affect the quality of clustering, such that irrelevant poses might appear in the same cluster. The result of our experiment using elbow method shows that the optimal number of cluster heads is around 10-15 as depicted in Figure~\ref{fig:elbow} in Section~\ref{sec:exp_arpose}.

After finding the number of clusters, we set up two clustering algorithms, namely, Kmeans and Deep Embedding Clustering (DEC). As for the Kmeans, the only parameter that we should set is the number of clusters, but in DEC we should setup the auto-encoder network in addition to the number of clusters. As suggested by~\citet{xie2016unsupervised} and tested by ourselves, the network with $4$ layers of encoder consisting of $500, 500, 2000$, and $10$ neurons in each unit performs astonishingly well on the clustering task of different supervised datasets including but not limited to MNIST~\citep{lecun1998gradient}, STL~\citep{coates2011analysis}, and REUTERS~\citep{lewis2004rcv1}. Although DEC works great on these supervised datasets, it has not been tested on an actual unsupervised dataset, simply because there is not a gold standard to evaluate the performance on those datasets. However, visual data like the unsupervised portrait dataset reveals how these algorithms perform, based on human vision evaluation of the output. Hence, we compare the results of this deep model for clustering with the base clustering algorithm, Kmeans. 

In Kmeans, to define the probability that each sample is in the cluster or the degree to which each sample belongs to a cluster, we use the same quantity in fuzzy C-means clustering~\citep{dunn1973fuzzy}:
\begin{equation}
q_{ij} = \frac{1}{\sum_{k=1}^{K}\left(\frac{||x_i - c_j||_2}{||x_i - c_k||_2}\right)^{\frac{2}{m-1}}}\;, 
\end{equation}
where $x_i$ is the sample, $c_j$ is the center of the cluster $j$, and $m$ is positive real number greater than $1$ which defines the smoothness of the function. $q_{ij}$ represents the probability or degree to which each sample belongs to a cluster. Also, DEC has a similar quantity defined by~\citet{xie2016unsupervised}, using Student's t-distribution:
\begin{equation}
q_{ij} = \frac{(1 + ||z_i - c_j||_2^2/\alpha)^{-\frac{\alpha+1}{2}}}{\sum_{j'}(1 + ||z_i - c_{j'}||_2^2/\alpha)^{-\frac{\alpha+1}{2}}}\;,
\end{equation}
in which $z_i$ is the embedded version of $x_i$, and $\alpha$ is the degree of freedom in Student's t-distribution. Using these metrics we estimate the probability that each sample belongs to a cluster.

\begin{figure*}[!tb]
\centering
\includegraphics[height=1in]{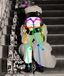}
\includegraphics[height=1in]{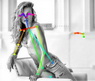}
\includegraphics[height=1in]{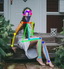}
\includegraphics[height=1in]{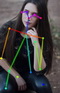}
\includegraphics[height=1in]{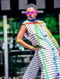}
\includegraphics[height=1in]{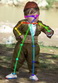}
\includegraphics[height=1in]{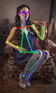}
\includegraphics[height=1in]{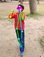}
\vspace{0.8mm} \\
\includegraphics[height=1in]{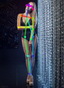}
\includegraphics[height=1in]{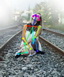}
\includegraphics[height=1in]{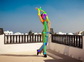}
\includegraphics[height=1in]{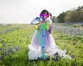}
\includegraphics[height=1in]{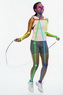}
\includegraphics[height=1in]{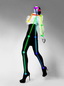}
\includegraphics[height=1in]{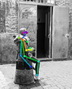}
\vspace{0.8mm} \\
\includegraphics[height=1in]{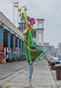}
\includegraphics[height=1in]{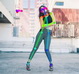}
\includegraphics[height=1in]{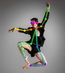}
\includegraphics[height=1in]{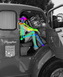}
\includegraphics[height=1in]{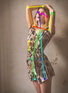}
\includegraphics[height=1in]{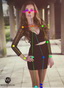}
\includegraphics[height=1in]{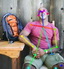}
\includegraphics[height=1in]{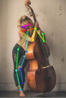}
\vspace{0.8mm} \\
\includegraphics[height=1in]{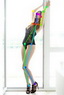}
\includegraphics[height=1in]{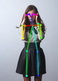}
\includegraphics[height=1in]{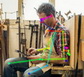}
\includegraphics[height=1in]{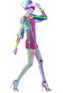}
\includegraphics[height=1in]{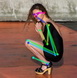}
\includegraphics[height=1in]{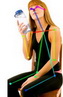}
\includegraphics[height=1in]{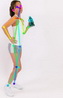}
\includegraphics[height=1in]{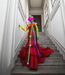}
\vspace{0.8mm} \\
\caption{Qualitative results of four clusters derived from the portrait dataset by DEC algorithm. Each row represents the top poses in each cluster. The DEC algorithm fails in clustering poses because the poses in the same cluster are not all consistent. }
\label{fig:posedec}
\end{figure*}

The qualitative results of Kmeans-based clustering algorithm are depicted in Figure~\ref{fig:posekmeans12} and Figure~\ref{fig:posekmeans22}, showing the top ranked poses in all fifteen major clusters, and those of DEC algorithm are depicted in Figure~\ref{fig:posedec} showing the best ranked poses of first 5 clusters. The top images are ranked based on their probability, calculated as above. As shown in the figures, Kmeans clusters surprisingly better represent poses in images, that is, different pose clusters distinguish between different poses and each cluster represents visually the same pose. However, DEC fails to accomplish the goal of clustering task based on poses of human subjects. Since the input features are intelligently chosen to be related to the goal, the input space is linearly separable, however, the result of the DEC shows information loss in the autoencoder. We tried the Kmeans algorithm with PCA to reduce the dimension of the input space to $10$ (as it is in the output of the autoencoder in DEC), and still the results of the Kmeans surpasses DEC's. Through that, we successfully categorize the input portrait image and retrieve similar poses close to the query or novel ideas in that pose category based on the probability of the poses.

\subsection{Construction of Composition Model}
In order to index all photos in our dataset, we decompose their values and construct our composition model (CM). If more photos are added to the dataset, we execute the decomposition for them, and update our composition model. In fact, $F_{I_{i},vgg}$, $F_{I_{i},iod}$, $F_{I_{i},cade}$, $F_{I_{i},arpose}$ and other aesthetic information vectors for all ($\forall i$) images are calculated and appended to corresponding matrices respectively including generic feature matrix $M_{vgg}$, integrated object detector matrix $M_{iod}$, category detector matrix $M_{cade}$, artistic pose detector matrix $M_{ap}$, statistics matrix $M_{stat}$ and gender matrix $M_{gnd}$. Algorithm~\ref{alg:decomp} describes different steps of our decomposition method precisely to make each row of our composition model. We have:
\begin{eqnarray}
\forall\;feat &\in& \{vgg,iod,cade,ap,stat,gnd\}\;, \nonumber\\
\forall i &\in& \{1,...,N\}\;, \nonumber\\
F_{I_i}&=&\left[ \begin{array}{c} F_{I_i,vgg} \\ F_{I_i,iod} \\ F_{I_i,cade} \\ F_{I_i,ap} \\ F_{I_i,stat} \\ F_{I_i,gnd} \end{array} \right]\;,
\end{eqnarray}
where ``$feat$'' is feature type from the set \{vgg, iod, cade, ap, stat, gnd\}, $F_{I_i}$ is the feature vector of the image $I_i$. Then, we compute the corresponding feature matrix.

\begin{eqnarray}
M_{feat}&=&\left[ \begin{array}{c} F^{T}_{I_{1},feat} \\ F^{T}_{I_{2},feat} \\ ... \\ F^{T}_{I_{N},feat} \end{array} \right]\;, \\
M&=&\left[ M_{vgg} \;\; M_{iod} \;\; M_{cade} \;\; M_{ap} \;\; M_{stat} \;\; M_{gnd} \right] \;,\\
&=& \left[ \begin{array}{c} F^{T}_{I_1} \nonumber\\ F^{T}_{I_2} \\ ... \\ F^{T}_{I_N} \end{array} \right]\;,
\label{eq:feat_matrix}
\end{eqnarray}
where matrix $M_{feat}$ is the corresponding feature matrix containing feature vector of each image in each row. The final feature matrix $M$ is the composition model matrix which is the concatenation of all feature matrices or equivalently all feature vectors.

\begin{algorithm}[!tb]
\caption{Decomposition} \label{alg:decomp}
\hspace*{\algorithmicindent} \textbf{Input:} image query $Q$. \\
\hspace*{\algorithmicindent} \textbf{Output:} the feature vector of the query $F^{T}_{Q}$.  
\begin{algorithmic}[1]
\Procedure{$Decomposition$}{$Q$}
\State Get the generic features of the query:

\noindent $F_{Q,vgg} \gets \left[f^{vgg}_{Q,1} \;\; f^{vgg}_{Q,2} \;\; ... \;\; f^{vgg}_{Q,4096}\right]^{T}$

\State Get stat and gender features:

\noindent $F_{Q,stat} \gets \left[f^{rating}_{Q,1} \;\; f^{views}_{Q,2}\right]^{T}$

\noindent $F_{Q,gnd} \gets \left[f^{male}_{Q,1} \;\; f^{female}_{Q,2} \;\; f^{unknown}_{Q,3}\right]^{T}$

\State Get the tensor of the object detector ($od$):

\noindent $T^{Q,od}_{m\times{n}\times{2}} \gets Object\_Detect(Q)$

\State Get the tensor of the scene parser ($sp$):

\noindent $T^{Q,sp}_{m\times{n}\times{2}} \gets Scene\_Parse(Q)$ 

\State Get the tensor of the pose estimator ($pe$):

\noindent $T^{Q,pe}_{m\times{n}\times{2}} \gets Pose\_Estimate(Q)$ 

\State Get the center of the mass coordinate of the query:

\noindent $c^Q \gets \frac{\sum_{i,j}{\norm{[i,j] - [0,0]}_k\times{S(i,j)}}}{\sum_{i,j}{\norm{[i,j] - [0,0]}_k}}$

\State Get the centric distance feature of the query:

\noindent $D^Q(i,j) \gets \frac{1}{K} {e^{-\norm{[i,j] - c^Q}_k}}$

\State Get the weighted saliency ID map for the query:

\noindent $C^Q(i,j) \gets S^Q(i,j) D^Q(i,j)$

\noindent $W^Q(i,j) \gets \max{\left(T^{Q,od}_{*,*,2},T^{Q,sp}_{*,*,2}\right)}  C^Q(i,j)$

\State Get the IOD feature vector as Eq.~\ref{iod_feat}:

\noindent $f^{imp}_{k} \gets \frac{\sum_{\forall objID(i,j)=k} {W(i,j)}}{\sum_{\forall i,j} {W(i,j)}},\forall k\in\{1,2,...,210\}$

\noindent $F_{Q,iod} \gets \left[f^{imp}_{1}\;f^{imp}_{2}\;...\;f^{imp}_{210}\right]$

\State Get the category feature $F_{Q,cade}$ as Eq.~\ref{cade_feat}.
\State Get the artistic pose feature $F_{Q,ap}$ as Eq.~\ref{ap_feat}.
\State Get the whole feature vector:

\noindent $F^{T}_{Q}=\left[ F^{T}_{Q,vgg} \; F^{T}_{Q,iod} \; F^{T}_{Q,cade} \; F^{T}_{Q,ap} \; F^{T}_{Q,stat} \; F^{T}_{Q,gnd} \right]$

\State If $Q$ is a dataset image, add $F^{t}_{Q}$ to the last row of composition model matrix $M$ in Equation~\ref{eq:feat_matrix}, otherwise the output is used in Algorithm~\ref{alg:comp_match} to retrieve better composed photos and match with the final shot.
\EndProcedure
\end{algorithmic}
\end{algorithm}

\section{Composition of Visual Elements}
The goal of composition step is to retrieve related photography ideas as highly-rated photos from our collected 500px dataset satisfying the proximity to the decomposed aesthetics-related information of the query image. In fact, the input to this step is the decomposed values of the image query and user-specified preferences (USP) with our composition model. The output of this stage is a collection of well-composed images from the dataset. If we focus on portraits, we desire a feedback that contains well-posed portraits with similar semantics but better aesthetic quality w.r.t USP. The ``interaction'' between the subject(s) and the objects in the image is important because system's proposed composition depends on them.

As we have collected the 500px dataset containing generally well-composed images, we should dig into the dataset and look for images with ``pretty'' similar color, pattern, category, pose, or object constellation where the term ``pretty'' is framed by USP to address the user's needs and subjectivity. The existence of this professional-quality dataset makes it possible that the retrieved photos have highly-accepted photography ideas by the people. Our image retrieval system is not supposed to find images with exactly similar colors, patterns, or poses, but it finds images with better composition having similar semantic classes. Thus, the location of the movable objects does not matter, but the detected objects are important. 

To look for a semantically composed version with respect to the image query, we exploit all of the decomposed values of the image query. Because we do not assume the query image taken by an amateur photographer to be well-composed enough to be the basic query for our personalized aesthetics-based image retrieval (PAIR). In fact, we just want to understand the location/setup around the subject, and then based on the scene ingredients, a well-composed image taken by a professional is proposed to the photographer. 

\subsection{Similarity Scores and Normalization}
Having our composition model for all images in the 500px dataset and the query image, we first calculate the similarity score between the query image and any image in the dataset. The similarity metric is different for each detector. For generic CNN descriptors is just matrix $M_{vgg}$ by the query vector $F_{vgg}$ multiplication. Similarly, category detector has a matrix by vector multiplication. For integrated object detectors, we use Gaussian function after masking unrelated objects. For statistics and gender information, it is trivial as in the following equations.
\begin{eqnarray}
&&\!\!\!\!\!S_{vgg}(I,Q)=F^{T}_{I,vgg} F_{Q,vgg}\;, \\
&&\!\!\!\!\!S_{cade}(I,Q)=F^{T}_{I,cade}  F_{Q,cade}\;, \\
&&\!\!\!\!\!S_{iod}(I,Q)=e^{- (\sum (F_{I,iod} \circ sgn(F_{Q,iod})-F_{I,iod}))^2}\;, \\
&&\!\!\!\!\!S_{stat}(I,Q)=FirstColumn(I_{stat})=I^{rating}_{stat}\;, \\
&&\!\!\!\!\!S_{gender}(I,Q)=
    \begin{cases}
      1, & \text{if}\ F_{I,gender} = F_{Q,gender} \\
      -1, & \text{otherwise}
    \end{cases}
\label{eq:sim_score1}
\end{eqnarray}
where $F^{T}$ means the transpose of $F$, $e$ is a mathematical constant about $2.72$, the $\circ$ operation is the element-wise multiplication, $sgn(.)$ is the sign function operating on each element separately. Also, $S_{vgg}(I,Q)$, $S_{cade}(I,Q)$, $S_{iod}(I,Q)$, and $S_{stat}(I,Q)$ are similarity score values between image $I$ and image $Q$ respectively for generic CNN descriptors, category detection, integrated object detectors, and statistics and gender information.

The similarity score function is easily generalized to a function between two different set of images, {\it i.e.}, $I_{m\times 1}$ and $Q_{n\times 1}$ can be a set of images not only one image, and the output will be a $m\times n$ matrix. Since we want to score the similarity between the images in the 500px dataset (say $\mathbb{I}$) and an image query ($Q$), in the above equations, vector $F^{t}_{I,det}$ will be substituted by matrix $M_{det}$, and the output will be a similarity vector, while $det$ can be any detector as follows:
\begin{eqnarray}
det\in\{vgg, iod, cade, arpose, stat, gender\} \nonumber
\end{eqnarray}

To make the scores uniform across various detectors, we normalize each detector score vector dividing by the summation of the whole output. Thus, each detector's similarity score is like a probability distribution over all images. We have:
\begin{eqnarray}
&&S^{N}_{feat}(\mathbb{I},Q) = \dfrac{S_{feat}(\mathbb{I},Q)}{\sum_{i\in \mathbb{I},q\in Q}{S_{feat}(i,q)}}\;,\\
&&feat \in \{vgg,iod,cade,arpose,stat,gender\}\;,
\label{eq:sim_score2}
\end{eqnarray}
where $S^{N}_{feat}(\mathbb{I},Q)$ is a normalized similarity score matrix between each image in $\mathbb{I}$ and each image in $Q$ for detector $feat$ which can be any of the mentioned detectors. Also, we combine the similarity scores across various detectors to create a tensor of similarity scores for each pair of images from $(\mathbb{I},Q)$. We have:
\begin{eqnarray}
&&S^{N}_{d\times m\times n}(\mathbb{I},Q)\\
&&=\left[S^{N}_{vgg}\;\; S^{N}_{iod}\;\; S^{N}_{cade}\;\; S^{N}_{arpose}\;\; S^{N}_{stat}\;\; S^{N}_{gender} \right]\;,\nonumber 
\label{eq:sim_score3}
\end{eqnarray}
where $S^{N}_{d\times m\times n}(\mathbb{I},Q)$ is a tensor of size $d\times m\times n$ where $d$ is the number of the $feat$ detectors ($||feat||$ here is 6), $m$ is the number of the images in $\mathbb{I}$, and $n$ is the number of the images in $Q$.

\subsection{User Preferences and Ranking}
\begin{figure*}[!tb]
\centering
\begin{tabular}{c|l} 
\includegraphics[height=0.8in]{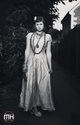}
& 
\includegraphics[height=0.8in]{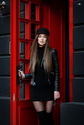}
\includegraphics[height=0.8in]{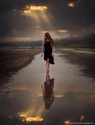}
\includegraphics[height=0.8in]{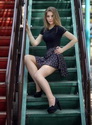}
\includegraphics[height=0.8in]{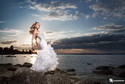}
\includegraphics[height=0.8in]{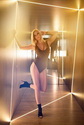}
\includegraphics[height=0.8in]{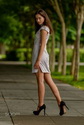}
\includegraphics[height=0.8in]{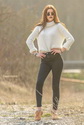}
\includegraphics[height=0.8in]{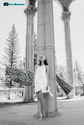}
\\
\includegraphics[height=0.8in]{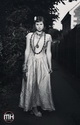}
& 
\includegraphics[height=0.8in]{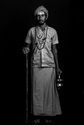}
\includegraphics[height=0.8in]{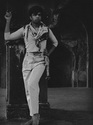}
\includegraphics[height=0.8in]{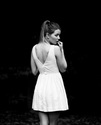}
\includegraphics[height=0.8in]{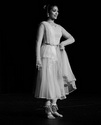}
\includegraphics[height=0.8in]{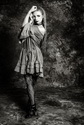}
\includegraphics[height=0.8in]{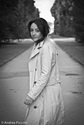}
\includegraphics[height=0.8in]{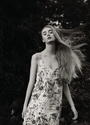}
\includegraphics[height=0.8in]{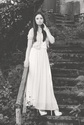}
\\
\includegraphics[height=0.8in]{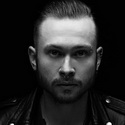}
& 
\includegraphics[height=0.8in]{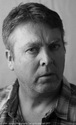}
\includegraphics[height=0.8in]{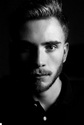}
\includegraphics[height=0.8in]{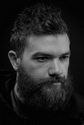}
\includegraphics[height=0.8in]{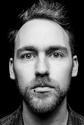}
\includegraphics[height=0.8in]{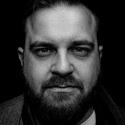}
\includegraphics[height=0.8in]{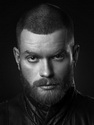}
\includegraphics[height=0.8in]{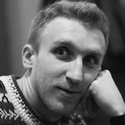}
\includegraphics[height=0.8in]{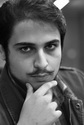}
\\
\includegraphics[height=0.8in]{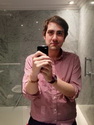}
& 
\includegraphics[height=0.8in]{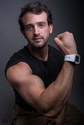}
\includegraphics[height=0.8in]{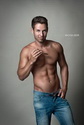}
\includegraphics[height=0.8in]{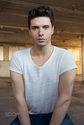}
\includegraphics[height=0.8in]{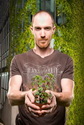}
\includegraphics[height=0.8in]{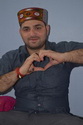}
\includegraphics[height=0.8in]{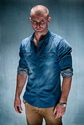}
\includegraphics[height=0.8in]{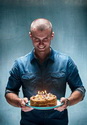}
\includegraphics[height=0.8in]{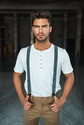}
\includegraphics[height=0.8in]{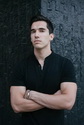}
\\
\includegraphics[height=0.6in]{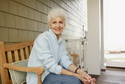}
& 
\includegraphics[height=0.8in]{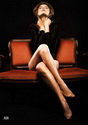}
\includegraphics[height=0.8in]{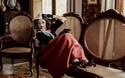}
\includegraphics[height=0.8in]{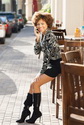}
\includegraphics[height=0.8in]{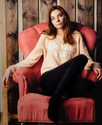}
\includegraphics[height=0.8in]{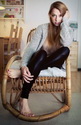}
\includegraphics[height=0.8in]{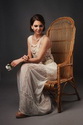}
\includegraphics[height=0.8in]{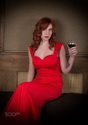}
\includegraphics[height=0.8in]{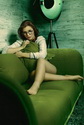}
\\
\includegraphics[height=0.8in]{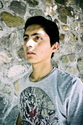}
& 
\includegraphics[height=0.8in]{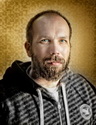}
\includegraphics[height=0.8in]{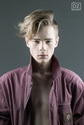}
\includegraphics[height=0.8in]{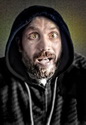}
\includegraphics[height=0.8in]{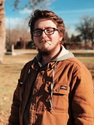}
\includegraphics[height=0.8in]{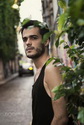}
\includegraphics[height=0.8in]{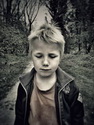}
\includegraphics[height=0.8in]{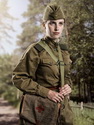}
\includegraphics[height=0.8in]{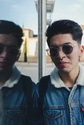}
\\
\includegraphics[height=0.6in]{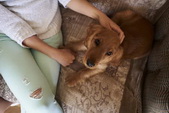}
& 
\includegraphics[height=0.8in]{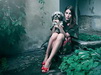}
\includegraphics[height=0.8in]{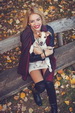}
\includegraphics[height=0.8in]{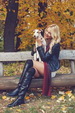}
\includegraphics[height=0.8in]{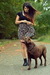}
\includegraphics[height=0.8in]{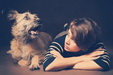}
\includegraphics[height=0.8in]{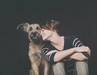}
\\
\includegraphics[height=0.6in]{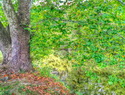}
& 
\includegraphics[height=0.6in]{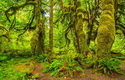}
\includegraphics[height=0.6in]{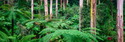}
\includegraphics[height=0.6in]{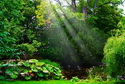}
\includegraphics[height=0.6in]{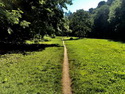}
\includegraphics[height=0.6in]{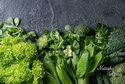}
\\
\hline
\\
\includegraphics[height=0.8in]{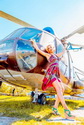}
& 
\includegraphics[height=0.8in]{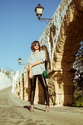}
\includegraphics[height=0.8in]{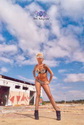}
\includegraphics[height=0.8in]{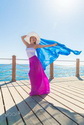}
\includegraphics[height=0.8in]{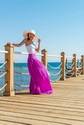}
\includegraphics[height=0.8in]{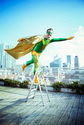}
\includegraphics[height=0.8in]{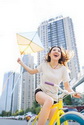}
\includegraphics[height=0.8in]{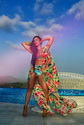}
\includegraphics[height=0.8in]{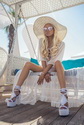}
\includegraphics[height=0.8in]{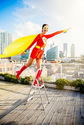}
\\
\includegraphics[height=0.6in]{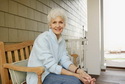}
& 
\includegraphics[height=0.6in]{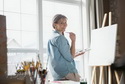}
\includegraphics[height=0.6in]{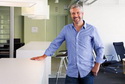}
\includegraphics[height=0.6in]{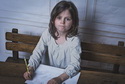}
\includegraphics[height=0.6in]{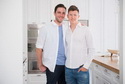}
\includegraphics[height=0.6in]{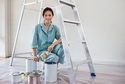}
\includegraphics[height=0.6in]{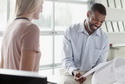}
\\
\end{tabular}
\caption{Qualitative results of the composition step. Each row shows a query image and the images retrieved w.r.t. the query and user-specified preferences (USP). Last two rows show failed cases because of an undetected object (e.g. helicopter) or incompletely entered USP. The USP for each row is: 1) $W_{cade}=W_{stat}=0.5$, 2) $W_{cade}=W_{vgg}=0.5$, 3) like (2), 4) like (2), 5) $W_{cade}=W_{iod}=0.5$, 6) $W_{gender}=W_{vgg}=0.5$, 7) $W_{gender}=W_{iod}=0.5$, 8) $W_{stat}=W_{vgg}=0.5$, 9) $W_{cade}=W_{vgg}=W_{iod}=0.33$, and 10) $W_{vgg}=W_{stat}=0.5$.}
\label{fig:compose}
\end{figure*}

The user-specified preferences are a probability vector containing the weights of the decomposed vectors of the image query. We have:
\begin{eqnarray}
&&\!\!W_{USP} \\
&&=\left[W_{vgg} \; W_{iod} \; W_{cade} \; W_{arpose} \; W_{stat} \; W_{gender} \right]^{T}, \nonumber
\end{eqnarray}
where $W_{USP}$ is a $d\times 1$ vector showing the weights of the user for each $feat$ detector, and ``t'' shows the transpose operation. Then, to retrieve the highest-ranked candidates as the results, the normalized similarity score matrix is multiplied by the USP vector. Consequently, we have:
\begin{equation}
V_{pref}(\mathbb{I},Q) = W^{T}_{USP} S^{N}(\mathbb{I},Q)\;,
\label{eq:pref}
\end{equation}
where $V_{pref}(\mathbb{X},Q)$ is the user's preferred image vector, and if we sort it in a descending manner with respect to the vector values, the indexes of the rows represent the highest-ranked candidates with the nearest feedbacks to the image query ($Q$). The whole process of the photography idea retrieval for an input image query ($Q$) is shown in Algorithm~\ref{alg:comp_match}, and our experimental results show the quality of the results. Also, sample results for some queries with USP are shown in Figure~\ref{fig:compose}.

\subsection{Offline Indexing and Real-time Searching}
\begin{figure}[!tb]
\centering
\includegraphics[width=0.75\linewidth]{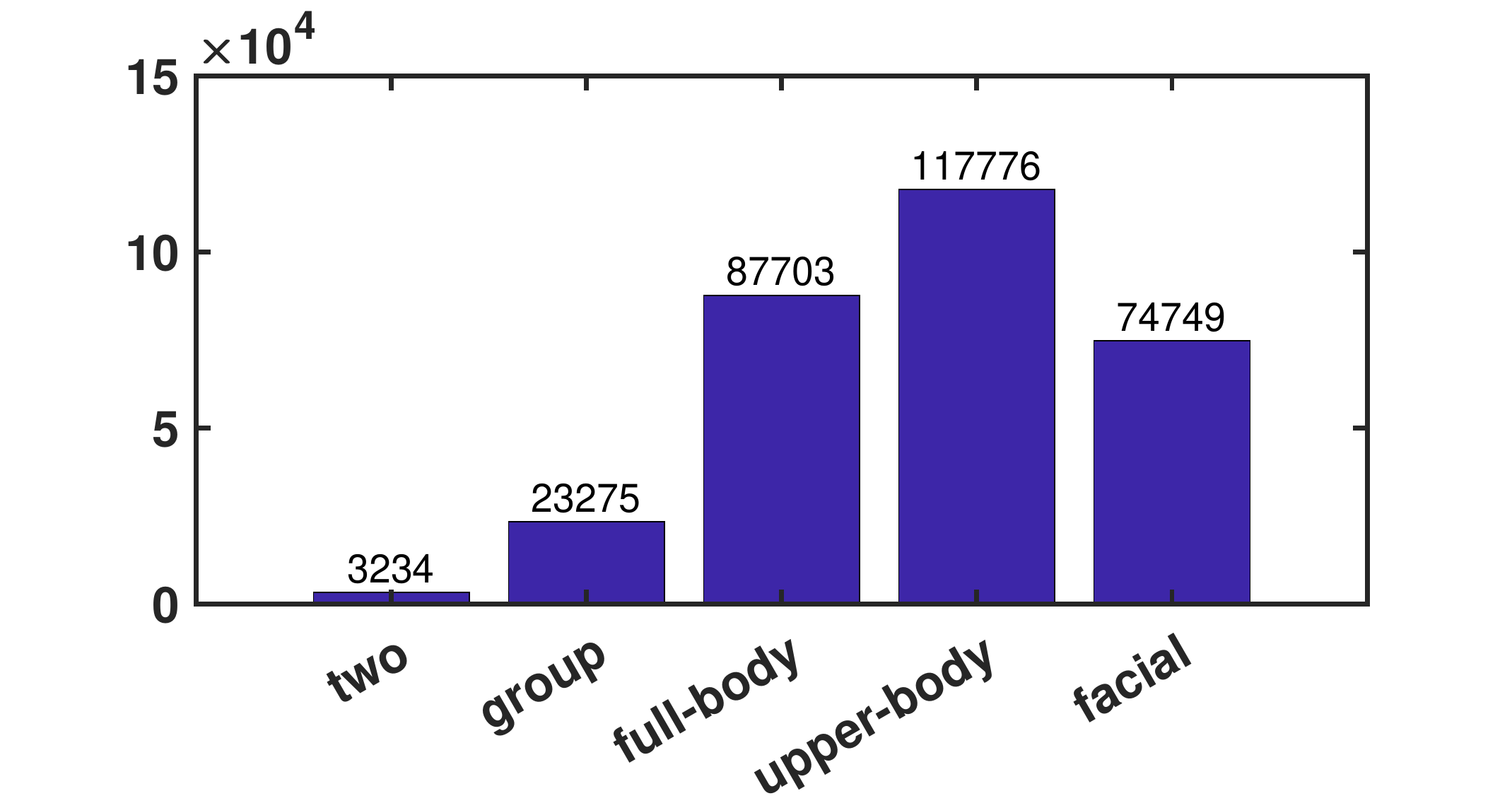}\\
\caption{The frequency of the portrait shots with respect to the highly-requested portrait categories.}
\label{fig:scenecount}
\vspace{-1\baselineskip}
\end{figure}

Our framework consists of the flows of indexing, searching, and matching shown in Figure~\ref{fig:method}. Practically, there are many challenges in image retrieval systems~\citep{smeulders2000content,lew2006content,datta2008image} as well as in our case. To improve the performance of our image retrieval system, we compute the decomposition step for all images ({\it i.e.}, indexing as an offline process). Indexing procedure is lengthy for the first time, but at the time of update, it is faster because the detections for an image is real-time using GPU. Furthermore, indexing procedure for our retrieval system organizes the decomposed values of the images into categorized matrices. Consequently, the composition step is real-time using GPU, as it just extracts the decomposed values of the query image similar to updating process, and then finds the target similar images, and finally retrieves best results from the dataset with respect to USP and normalized similarity score vector (Eq.~\ref{eq:pref}). Having various semantic classes for portrait genre in the dataset, we have indexed the 500px dataset from previous step. Figure~\ref{fig:scenecount} depicts the results as the number of portrait shots for some of the highly-frequent portrait categories.

\section{Matching}
\label{sec:match}

Professional photographers arrange the photograph from top to down (general to detail) step-by-step, while there are many to-do lists and not-to-do lists for photography in his/her mind. We want to create the same environment in a smart camera to accompany an amateur photographer gradually to his/her perfect shot. From composition step, we retrieve the proper photography ideas given a query shot from the camera. We assume that the photographer has chosen some of the retrieved images as a {\it preferred style set}, and disregarded the others as an ignored set. Now we explain how to capture the proper moment of the subject in the scene, and trigger the ``moment shot'' for the camera.

\subsection{Pose Shot}
The best-fitting genre for matching step is portrait photography that we start with, and then we extend for general genre. The variant component in our framework is the human pose. In this scenario, it is assumed that the user has no personal preference on human pose, {\it i.e.}, the user has given zero weight to the ArPose detector to see various proper poses via PAIR, and then the user has selected a preferred style set from the available choices. In this case, we need a mechanism to continue guiding the user to a desired shot by tracking his/her pose via camera viewfinder.

The relative positions of the human body components (including nose, eyes, ears, neck, shoulders, elbows, wrists, hips, knees, and ankles) with respect to the nose position as portrait origin are consisting our pose model. Preferably, we would like to start from the position of the nose ($J_{0}=\left(0,0\right)$) that is connected to neck $\left(J_{1}\right)$, right eye $(J_{2})$, and left eye $(J_{3})$ are connected to right ear $(J_{4})$ and left ear $(J_{5})$ as they are on a plane of the head. Also, shoulders $(J_{6}$ and $J_{7})$ are recognized by a length and an angle from neck, and similarly elbows ($J_{8}$ and $J_{9}$) from shoulders, wrists ($J_{10}$ and $J_{11}$) from elbows, hips ($J_{12}$ and $J_{13}$) from neck, knees ($J_{14}$ and $J_{15}$) from hips, and ankles ($J_{16}$ and $J_{17}$) from knees, {\it i.e.} these joints are connected as follows:
\begin{eqnarray}
Pre\left(J_{i}\right)&=&J_{0}\;,\; i \in\lbrace0,1,2,3\rbrace\;, \\
Pre\left(J_{j}\right)&=&J_{1}\;,\; j \in\lbrace6,7,12,13\rbrace\;, \\
Pre\left(J_{k}\right)&=&J_{k-2}\;, \\
k &\in& \lbrace4,5,8,9,10,11,14,15,16,17\rbrace\;. \nonumber
\end{eqnarray}
Thus, we always calculate the absolute position using 2D polar coordinates as follows:
\begin{equation} 
J_{i} = J_{j} + r_{i,j}.e^{\mathrm{i}\theta_{i,j}} \;,\; i\in\lbrace{0..17\rbrace} \;,
\label{eq:jointsrel}
\end{equation}
where $j=Pre(i)$ {\it i.e.} part $j$ is the previous part connected to part $i$, $r_{i,j}$ is the length from joint $J_{i}$ to joint $J_{j}$, $\theta_{i,j}$ is the angle between the line from joint $J_{i}$ to joint $J_{j}$ and the line from joint $J_{j}$ to joint $Pre(J_{j})$, and the line crossing $J_{0}$ is the image horizon. $\mathrm{i}$ is the unit imaginary number. Note that for a 2D human body $\{r_{i,j} | \forall i, j\}$ are fixed, but $\theta_{i,j};\forall i,j$ can be changed to some fixed not arbitrary extents. Similarly, having 3D pose-annotated/estimated single depth images, we can calculate the relative 3D position of the joints using spherical coordinates. Thus, we have such action boundaries for joints as follows:
\begin{eqnarray} 
\theta_{i,j}^{min} \leq \theta_{i,j} \leq
\theta_{i,j}^{max}\;,\; j=Pre(i)\;,
\label{eq:jointboundaries1} \\
\phi_{i,j}^{min} \leq \phi_{i,j} \leq
\phi_{i,j}^{max}\;,\; j=Pre(i)\;.
\label{eq:jointboundaries2}
\end{eqnarray}
As a result, a human body pose ($\mathrm{J}$) is represented by:
\begin{equation} 
\mathrm{J^{k}} = \left(J_{1}^{k},J_{2}^{k},...,J_{17}^{k}\right)\;,
\label{eq:humanbodypose} 
\end{equation}
where $\mathrm{J^{k}}$ is the pose for $k$-th person (or $k$-th image with one person), and $\forall i \in \{1..17\}:J_{i}^{k}$ is the $i$-th coordinate of the $k$-th person. Also, we need a distance metric to calculate the difference between two pose features. Thus, we define the distance metric as follows:
\begin{equation} 
D\left(\mathrm{J^{k}},\mathrm{J^{l}}\right) \doteq \sum_{i=1}^{17} {\lVert{Phase(J_{i}^{k}) - Phase(J_{i}^{l})}\rVert_{q}}\;,
\label{eq:featdist}
\end{equation}
where $D\left(.\right)$ is the distance operator, $\mathrm{J^{k}}$ is the pose feature for $k$-th person (or $k$-th image with one person), $\forall i \in \{1..17\}:J_{i}^{k}$ is the $i$-th coordinate of the $k$-th person, and ${\lVert{.}\rVert_{q}}$ (usually L1-norm or L2-norm) is the $L_q-norm$ function of two equal-length tuples. Our chosen function to track phase, $Phase(J{i})$, is $\sin(\theta_{i,i-1})$ which $\theta_{i,i-1}$ (the angle between current joint and the previous one) is from $-\Pi /2$ to $+\Pi /2$. Because the length of the limb may change by going far or near, but the angles between consecutive limbs matter for posing purposes. 

Now, the camera may take and hold several photos gradually from the scene, and finally choose the best among them to save onto the camera storage. Actually, our matching algorithm searches among the taken photos to get the nearest pose to one of the collected ideas. The problem is formulated as an integer programming problem to find the best seed among all photography ideas. Given the distance operator of two pose features explored in~\ref{eq:featdist}, we construct our optimization problem by maximizing the difference of the minimum distance of the ignored set and the minimum distance of the preferred style set of taken photos. Mathematically, we compute the following optimization problem subject to~\ref{eq:jointboundaries1} and~\ref{eq:jointboundaries2}:
\begin{eqnarray}
I_{w} =&& \argmaxA_{\forall I_{i}\in I^t} \\
&&\left( \min_{\forall Q_j^g \in Q^g}{D\left(\mathrm{J^{Q_j^g}},\mathrm{J^{I_{i}}}\right)}\right.
\left. -\min_{\forall Q_k^d \in Q^d}{D\left(\mathrm{J^{Q_k^d}},\mathrm{J^{I_{i}}}\right)} \right)\;,\nonumber
\end{eqnarray}
where $I_{w}$ is the wish-image, $I^t$ is the set of taken photos, $Q^g$ is the set of ignored ideas, $Q^d$ is the set of preferred ideas, $D\left(.\right)$ is the distance operator in~\ref{eq:featdist}, and $\mathrm{J^{x}}$ is the pose for $x$-th image with one person in~\ref{eq:humanbodypose}. The optimization problem in continuous mode (not over all taken image set) may have (a) solution(s) in feasible region, and in L1-norm case, it is equivalent to multiple linear programming problems but the complexity of the problem is 
exponential. Further, the solution does not always give the desired shot. 

\subsection{User Favorite Shot}
Given a query shot from the camera, related photography ideas have been already retrieved. Suppose that the photographer selects a {\it preferred style set}, denoted as $\mathbb{C}=\{C_{1},C_{2},...,C_{m}\}$, and we also have a set of shots from camera called next query shots, denoted as $\mathbb{Q}=\{Q_{1},Q_{2},...,Q_{n}\}$. The problem of finding the user favorite shot among query shots while satisfying the closest similarity score to the {\it preferred style set} is an integer programming. We have: 
\begin{eqnarray}
Q_{fav}=\argmaxA_{q\in \mathbb{Q}} \sum_{j\in \{1,...,m\}} W^{T}_{USP} S^{N}(C_{j},q)\;,
\label{fav_shot}
\end{eqnarray}
where $Q_{fav}$ is the favorite shot, $\mathbb{Q}$ is the set of the query shots by camera, $W^{t}_{USP}$ is the transpose of the user-specified preference vector, and $S^{N}(C_{j},q)$ is the similarity vector between query $q$ and each photo $C_{j}$ in the preferred style set of the user $\mathbb{C}$. The computational detail of the composition and matching steps has been explained in Algorithm~\ref{alg:comp_match} and inspired from \citep{diyanat2011image,farhat2011multi,farhat2012eigenvalues,farhat2014towards}.

Mathematically $\mathbb{Q}$ set is not finite, or its size $n$ is not bounded. Also, there are many constraints such as color value ranges, human pose angles, category limits, and etc. In the reality, the number of the query shots is limited, and the matching solver gradually determines and updates the user favorite shot. But the solution is not necessarily optimal, because finding the optimal shot needs the whole shot space which is impractical. The good news is that the user can follow the retrieved professional shots to optimize his/her photography adventure, and the last shot would be close enough to the optimal shot. 

Our approach to giving hints to the user includes two steps: 1) defining the query shot space with dynamic parameters in the scene like movable objects or human pose, 2) finding the max over the defined space. This second step is similar to pose shot approach, and some extra parameters such as photographic lighting may be adjustable as well. The solution of the problem can give a hint to the user to make a change in his/her lighting condition, pose, or any other dynamic parameter.

\begin{algorithm}[!tb]
\caption{Composition and Matching} \label{alg:comp_match}
\hspace*{\algorithmicindent} \textbf{Input:} query $Q$, user pref. $W_{USP}$, and the set of the images in the 500px dataset $\mathbb{I}$. \\
\hspace*{\algorithmicindent} \textbf{Output:} user favorite shot $Q_{fav}$.  
\begin{algorithmic}[1]
\Procedure{$IdeaRetrieval$}{$Q$, $W_{USP}$, $\mathbb{I}$}
\State Get $F^{t}_{Q}$ from Algorithm~\ref{alg:decomp}.
\State Get the similarity score through Eq.~\ref{eq:sim_score1}, ~\ref{eq:sim_score2}, and~\ref{eq:sim_score3}:

\noindent $S^{N}(\mathbb{I},Q)=\left[S^{N}_{vgg} \; S^{N}_{iod} \; S^{N}_{cade} \; S^{N}_{arpose} \; S^{N}_{stat} \; S^{N}_{gender} \right]$

\State Get the preferred image vector through Eq.~\ref{eq:pref}:

\noindent $V_{pref}(\mathbb{I},Q) = W^{t}_{USP} S^{N}(\mathbb{I},Q)$

\State $Retrieved\_Indexes \gets Index\_Sort(V_{pref}(\mathbb{I},Q))$
\State $Show\_Top(Retrieved\_Indexes)$
\State Now, the user selects some of the retrieved results, and the camera takes multiple shot as $\mathbb{Q}$.
\State Find the favorite shot through Eq.~\ref{fav_shot}:

\noindent $Q_{fav}=\argmaxH_{q\in \mathbb{Q}}(\sum_{j\in {1,...,m}}(W^{t}_{USP} S^{N}(\mathbb{C},q)))$

\State Take $Q_{fav}$ as user favorite shot. 
\EndProcedure
\end{algorithmic}
\end{algorithm}

\section{Experiments}
\label{sec:exp}
In the following sub-sections, we describe our experimental results which are categorized into different components of our method including (i) the dataset, (ii) the decomposition step, and (iii) composition step. Furthermore, the decomposition step includes object detector, pose estimator, scene parser has multiple parts to demonstrate the effectiveness of our method compared to an state-of-the-art available approach.
\vspace{-2\baselineskip}

\subsection{Dataset Properties}
\label{sec:exp_dataset}
\begin{figure*}[!tb]
\begin{center}
\begin{subfigure}[b]{0.30\textwidth}
    	\includegraphics[height=1.2in]{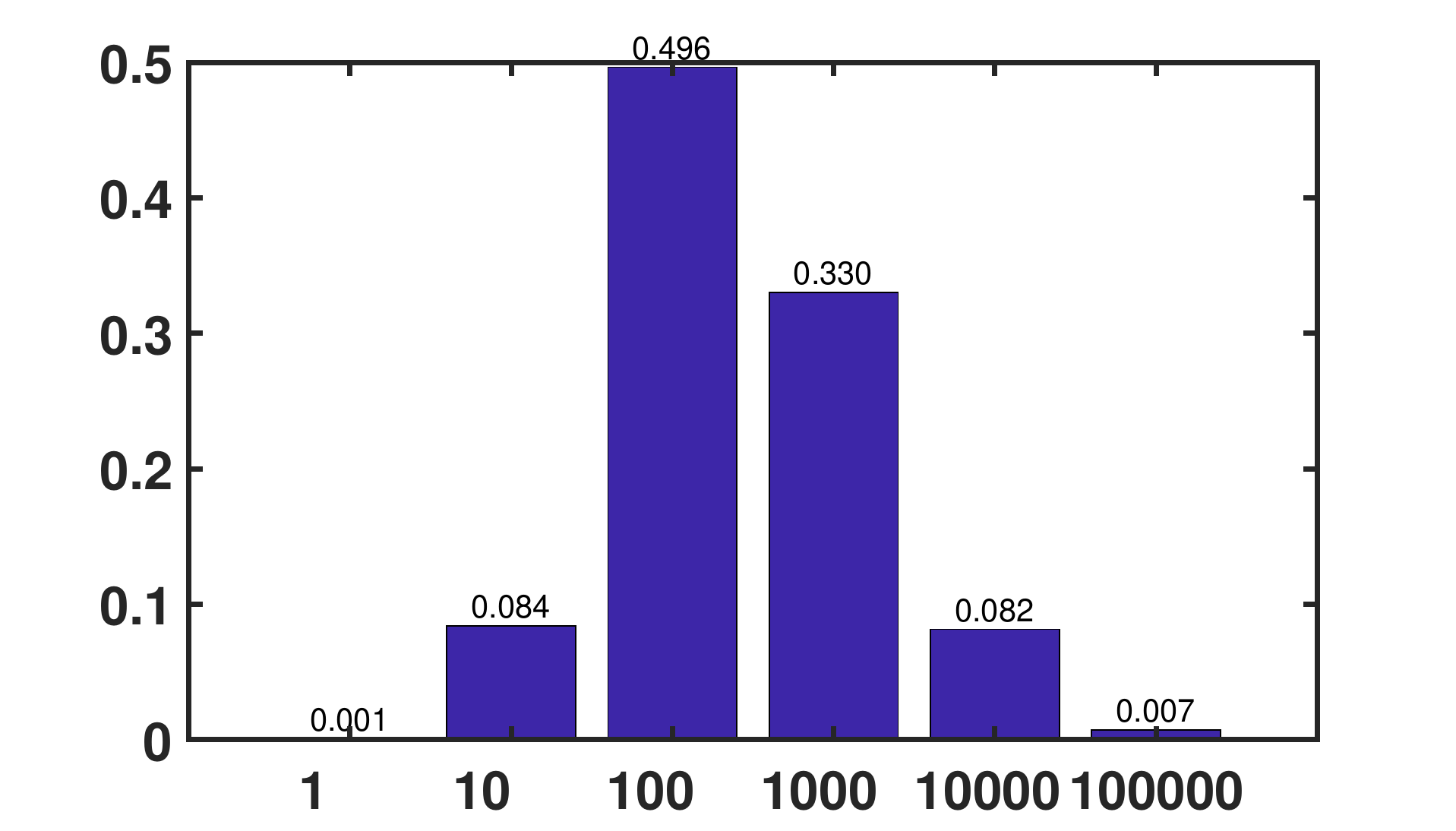}
      \caption{view count.}
      \label{fig:view_counts}
    \end{subfigure}%
    ~ 
    \begin{subfigure}[b]{0.23\textwidth}
    	\includegraphics[height=1.2in]{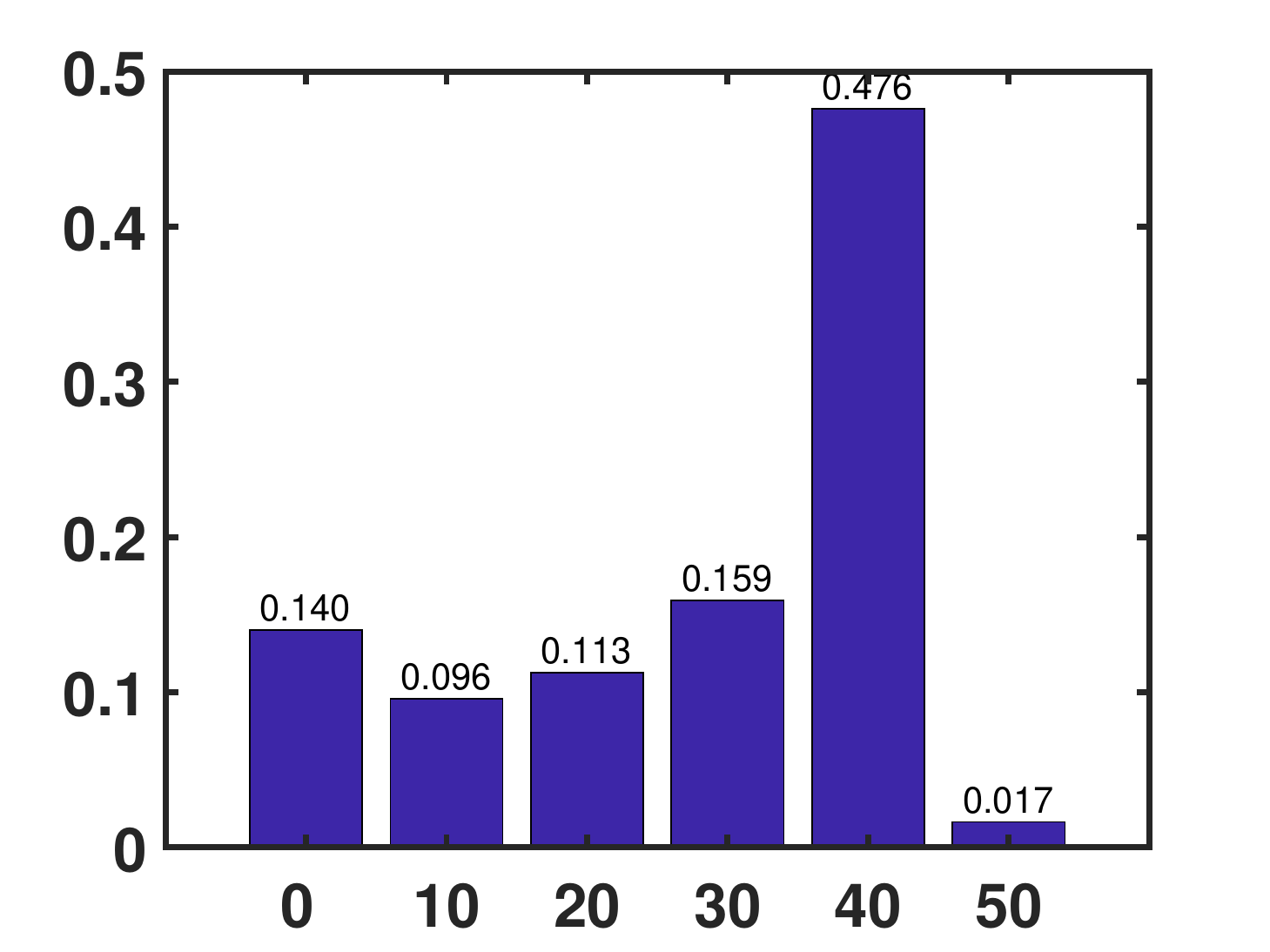}
      \caption{rating.}
      \label{fig:ratings}
    \end{subfigure}%
    ~
    \begin{subfigure}[b]{0.23\textwidth}
    	\includegraphics[height=1.2in]{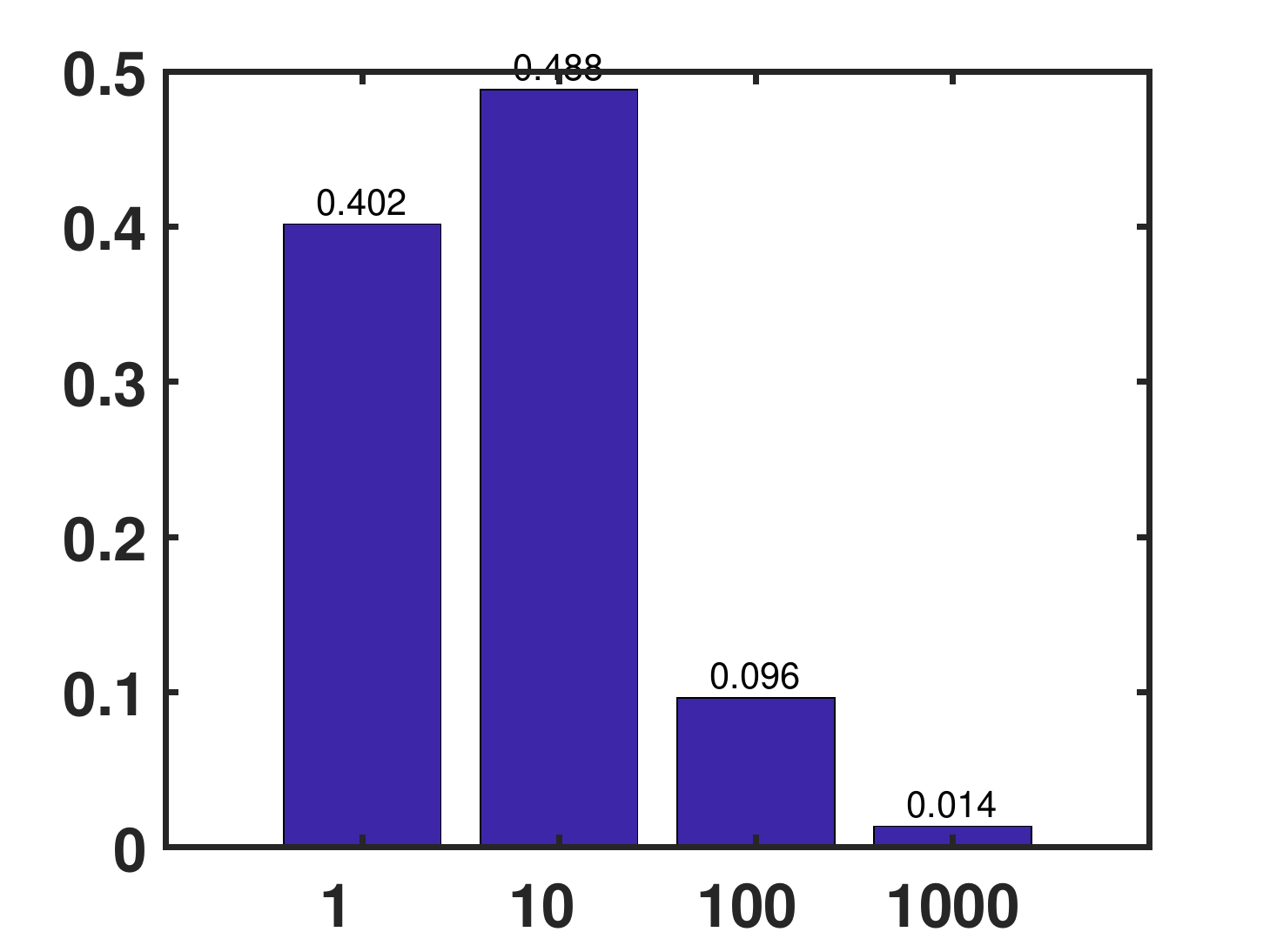}
      \caption{vote count.}
      \label{fig:vote_counts}
    \end{subfigure}%
    ~
    \begin{subfigure}[b]{0.24\textwidth}
    	\includegraphics[height=1.2in]{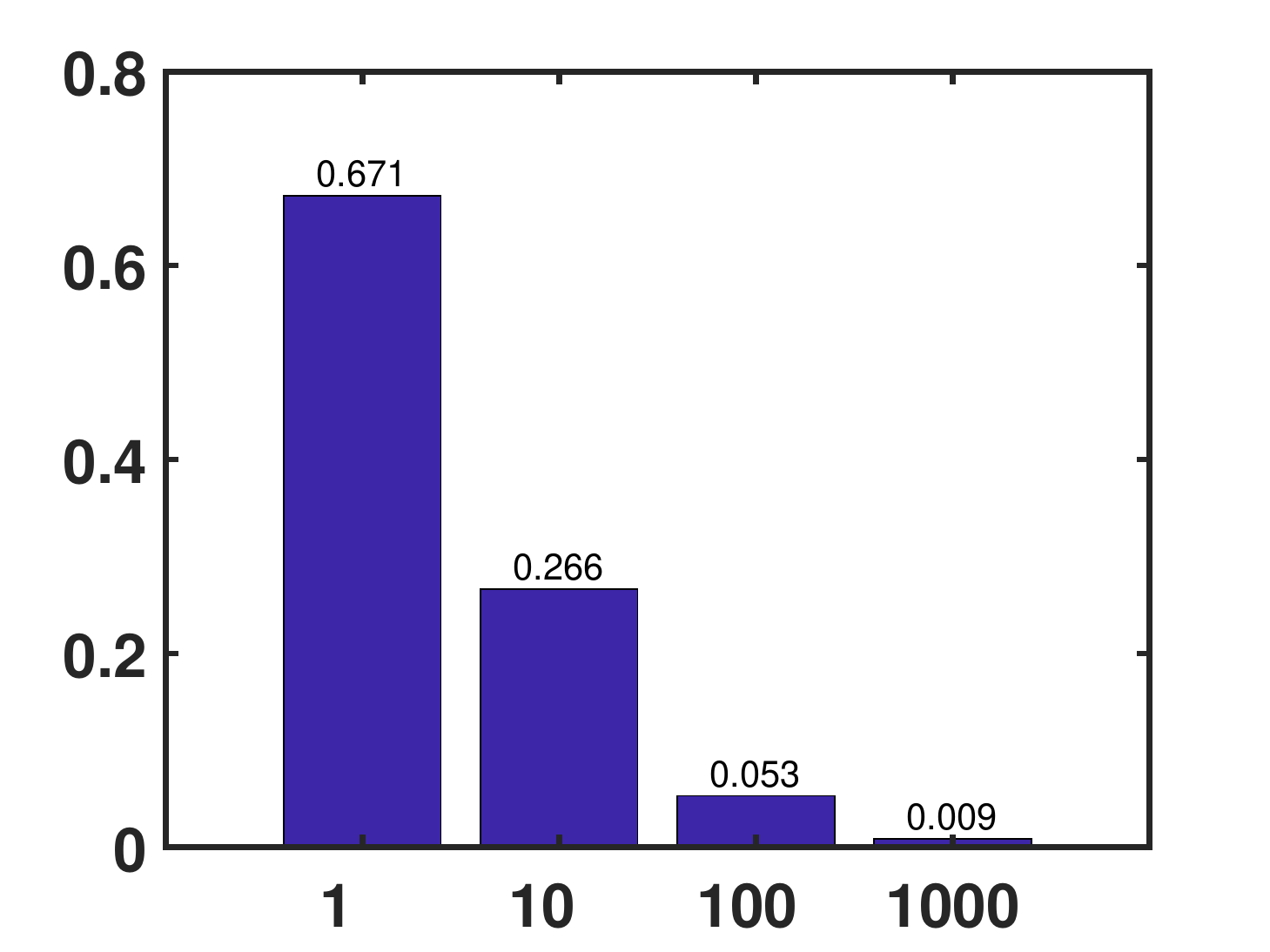}
      \caption{favorite count.}
      \label{fig:fav_counts}
    \end{subfigure}%
\caption{The properties of the managed 500px dataset respectively from left to right including (a) the logarithmic distribution of the view counts, (b) the distribution of the ratings, (c) the logarithmic distribution of the vote counts and (d) the logarithmic distribution of the favorite counts.}
\label{fig:dataset_properties}
\end{center}
\vspace{-1\baselineskip}
\end{figure*}

We have collected the images in the portrait and landscape categories from 500px Website and saved them as smaller images where their highest dimension has been resized to 500 pixels. Then, we have collected available metadata for each image including the number of views, the average ratings, the number of vote clicks, and the number of favorite clicks. Both jobs are very time consuming, and we are still crawling and managing to update the dataset for better coverage with up-to-date data.

We conduct statistical experiments to get the properties of the collected dataset. The best way to show the distribution of the number of views, the number of votes, and the number of favorites is using the logarithmic bin interval versus the frequency or probability of each bin interval. Because, these properties change dramatically in linear scale, and their trends can be captured intuitively in logarithmic x-axis. But, we show the average ratings using a normal bin interval. Figure~\ref{fig:dataset_properties} illustrates the distributions of the view counts, ratings, vote counts, and favorite counts of the dataset. Each bar represents a bin where its interval is from the corresponding number written under the bin to right before the number written under the next bin. The last bin interval is from its corresponding number to the next predictable number in the sequence of the axis.

Figure~\ref{fig:dataset_properties} shows that most of the images have been seen more than 100 times, i.e., 500px Website has a live community, while many images have at least 1-10 votes or favorite clicks. Having a rating higher than 10 is considered high by us, because the rating trend changes its slope direction from bin 0-9 to bin 10-19 negatively, and after that, the slope will positively grow until bin 40-49. Most of the images in the dataset have a rating more than 40 which is a very high rating, and it indicates that the 500px community of the photographer has many highly-rated photos.

\subsection{Decomposition Analysis}
To show the improvement and the effectiveness of our decomposition step, we conduct some experiments on various detectors used in our framework including object detector, human pose estimator, and scene parser. Also, we examine our hysteresis detection, category detector, and pose clustering.

\subsubsection{Object Detection}
\label{sec:exp_od}
Our network for object detection is inspired by YOLO as it is fast compared to the others\citep{redmon2016you}. The output of the network is some bounding boxes where detected objects are respectively with their detection probabilities. As we have tested, non-person object detection of YOLO under 30\% probability is not accurate enough on the 500px dataset, and any wrong detection affects all pixels in the bounding box based on our method. As a result, we divide the input image into bigger chunks of $5x5$ grid for a higher accuracy, and small objects are less important for detection as a secondary subject of photography. We implement our model as 24 convolutional layers with two fully connected layers. We train the whole network on ImageNet~\citep{deng2009imagenet} for about a week, and three times on a labeled subset of 768 common failure cases (CFC) from the 500px dataset.

We evaluate our model compared to YOLO on a test subset from the dataset. We use the regular MAP on all intended objects. Table~\ref{tab:yolo} shows the MAP and the average accuracies of some objects (person, seat, plant, animal, and car) for our trained model versus YOLO model. The ``seat'' average accuracy is the average for ``seat, bench, and chair'', ``plant'' average accuracy is the average for ``plant, tree, and grass'', and ``animal'' average accuracy is the average for ``bird, cat, dog, cow, and sheep''.

\begin{table}[!tb]
\begin{center}
{\renewcommand{\arraystretch}{1.3}
\begin{tabular}{@{\hskip 0.0in}c@{\hskip 0.05in}|cccccc@{\hskip 0.05in}}
  \hline
  Method & MAP & person & seat & plant & animal & car \\
  \hline
  YOLO & 52.0 & 73.5 & 40.1 & 33.2 & 71.2 & 54.8 \\ 
  Ours & 60.2 & 78.1 & 53.2 & 46.8 & 69.8 & 58.2\\
  \hline
 \end{tabular}
 }
\caption{The accuracy comparison between our object detector model versus the YOLO model on the 500px dataset to detect some of its known objects.}
\label{tab:yolo}
\end{center}
\vspace{-2.0\baselineskip}
\end{table}

\subsubsection{Pose Estimator}
\label{sec:exp_pe}
Our pose estimator architecture has two parallel lines predicting limb confidence map and encoding limb-to-limb association, which is inspired by RTMPPE architecture \citep{cao2017realtime}. We adjust the transformation parameters of the architecture including maximum rotation degree to $60$, crop size to $500$, scale min to $0.6$ and scale max to $1.0$. Since higher rotation degrees and bigger persons are used frequently in our work. Then, we train our model on MSCOCO \citep{lin2014microsoft}, MPII \citep{andriluka14cvpr} and three times on our 317 CFC from the 500px dataset.

To evaluate the performance of our pose estimator model, we leverage MAP of all limbs like DeeperCut \citep{insafutdinov2016deepercut}. The comparison results of the MAP performance between RTMPPE and our approach on a subset of 507 testing images from our dataset are shown in Table~\ref{tab:rtmppe}, where the left limb and the right limb are merged.

\begin{table}[!tb]
\begin{center}
\setlength{\tabcolsep}{0.45em}
{\renewcommand{\arraystretch}{1.3}
 \begin{tabular}{c|cccccccc}
  \hline
  Method & MAP & hea & sho & elb & wri & hip & kne & ank \\
  \hline
  RTMPPE & 73.3 & 90.4 & 83.3 & 74.6 & 65.1 & 70.9 & 66.3 & 62.6 \\ 
  Ours & 74.3 & 92.2 & 86.7 & 75.3 & 64.8 & 70.4 & 67.3 & 63.4\\
  \hline
 \end{tabular}
 }
\caption{The comparison between our pose estimator model versus the RTMPPE model on the 500px dataset to detect body parts.}
\label{tab:rtmppe}
\end{center}
\vspace{-1.0\baselineskip}
\end{table}

\subsubsection{Scene Parser}
\label{sec:exp_sp}
To parse any scene, we ignore confused categories like building and skyscraper. We place the related objects in the same object category. Also, our scene parser architecture exploits a 4-level pyramid pooling module \citep{zhao2017pyramid} with sizes of $1×1$, $2×2$, $3×3$ and $4×4$ respectively. We do not consider detecting small objects in the scene since they are mostly not the main subject of the photographer.

To train our model, we use ADE20K dataset \citep{zhou2017scene} with 576 common failure cases annotated by LabelMe \citep{russell2008labelme}. To evaluate scene parsing performance, pixel-wise accuracy (PixAcc) and mean of class-wise intersection over union (CIoU) are measured. The performance values of our scene parser model versus PSPNet \citep{zhao2017pyramid} with 101-depth ResNet is shown in Table~\ref{tab:pspnet} where indicates better PixAcc and CIoU than PSPNet has been achieved on the 500px dataset.

\begin{table}[!tb]
\begin{center}
\setlength{\tabcolsep}{1em}
{\renewcommand{\arraystretch}{1.3}
\begin{tabular}{c|cc}
  \hline
  Method & Pixel Accuracy ($\%$) & Mean IoU ($\%$) \\
  \hline
  PSPNet & 74.9 & 40.8 \\ 
  Ours & 78.6 & 42.5\\
  \hline
 \end{tabular}
 }
\caption{The accuracy comparison between our scene parser model versus the PSPNet model with 101-depth ResNet on the 500px dataset.}
\label{tab:pspnet}
\end{center}
\vspace{-1.0\baselineskip}
\end{table}

\subsubsection{Hysteresis Detection}
Hysteresis detection covers more photos by allowing the union of all images above HIGH thresholds across the detectors. We show how we configure these tunable thresholds, while we trade-off between the total coverage and the partial accuracies by the detectors. When we have more than one detector with common detectable objects, we consider multiple features from all detectors to decide about the detection of the common objects. For example, ``person'' is a common object between object detector and pose estimation. We perform our pose estimator on our ground-truth images with a person or without any person from the dataset, and calculate (a) the detection score (as mentioned in Eq.~\ref{eq:T_pe}) and (b) the normalized area ({\it i.e.} the detected object area divided by the image area) of the dominant person ({\it i.e.} the person with the highest score) detected in each image as our pose estimator features. Also, we perform our object detector on those images, and compute (c) the detection probability and (d) the normalized areas of the dominant person ({\it i.e.} the person with the highest probability) detected in each image as our object detector features.

\begin{figure*}[!tb]
    \begin{subfigure}[b]{0.23\textwidth}
    	\includegraphics[height=1.1in]{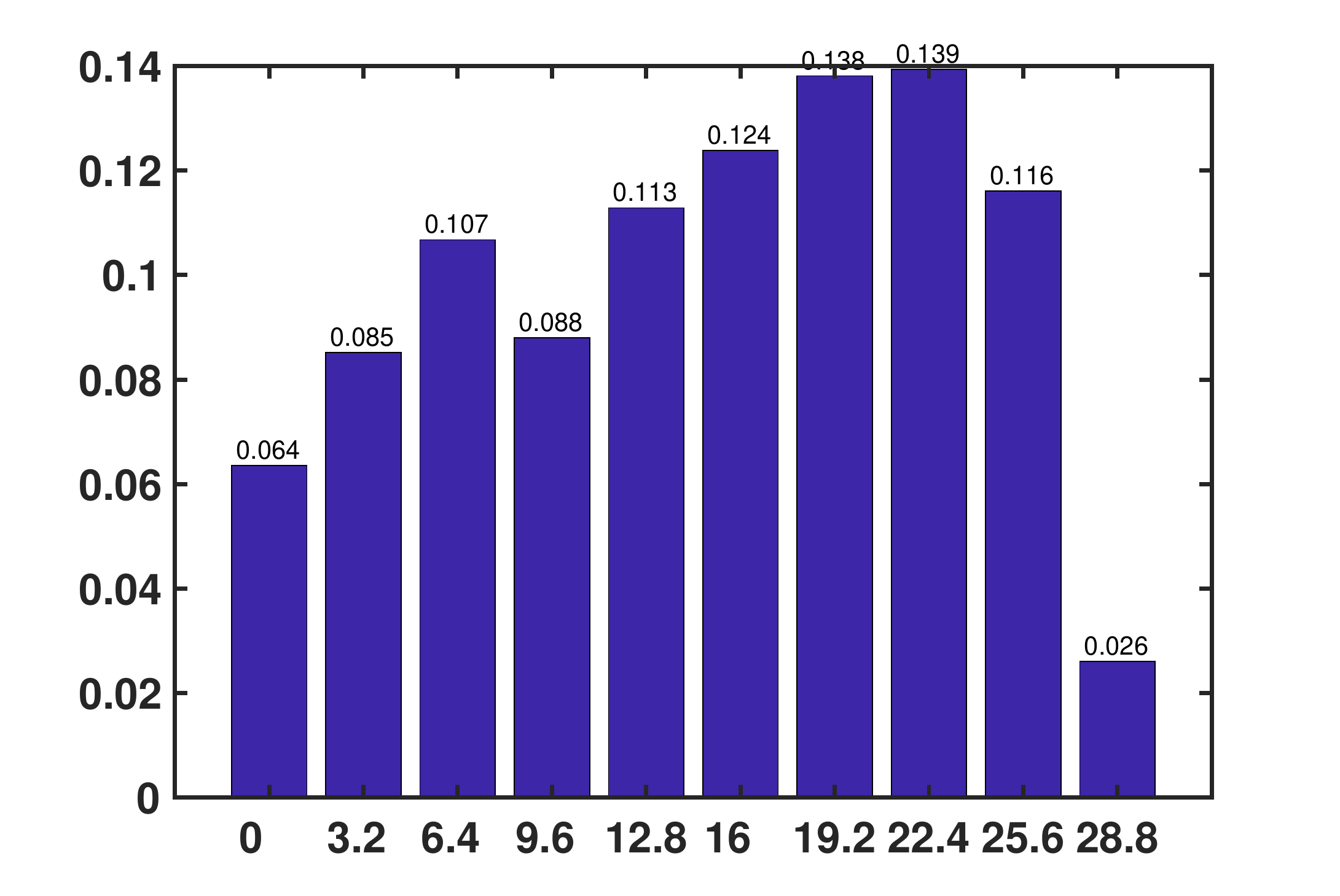}
      \caption{pose score}
      \label{fig:portrait_rtmppe_score}
    \end{subfigure}
    \begin{subfigure}[b]{0.27\textwidth}
    	\includegraphics[height=1.1in]{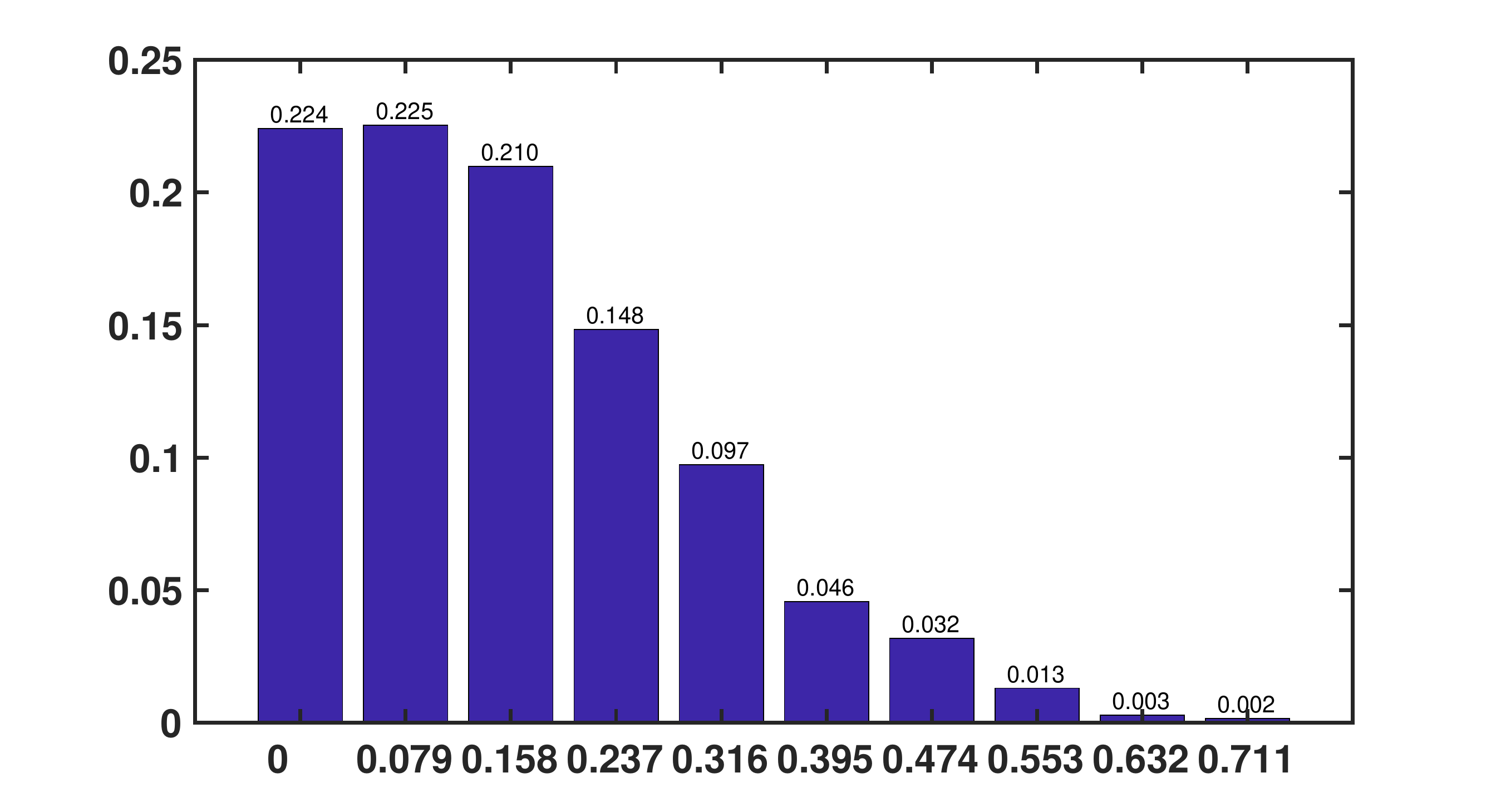}
      \caption{pose area}
      \label{fig:portrait_rtmppe_area}
    \end{subfigure} 
    \begin{subfigure}[b]{0.24\textwidth}
    	\includegraphics[height=1.1in]{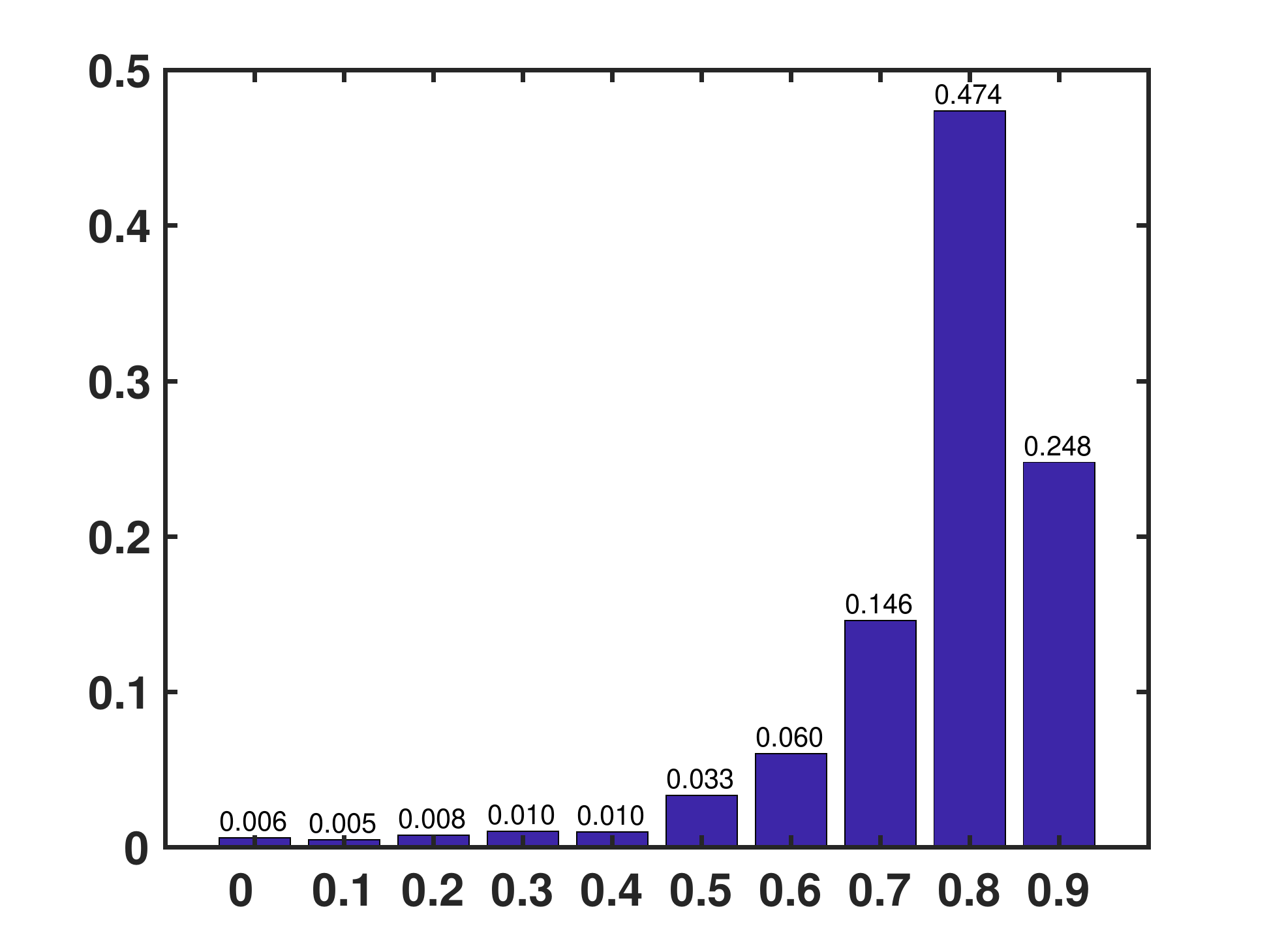}
      \caption{object probability}
      \label{fig:portrait_yolo_prob}
    \end{subfigure} 
    \begin{subfigure}[b]{0.24\textwidth}
    	\includegraphics[height=1.1in]{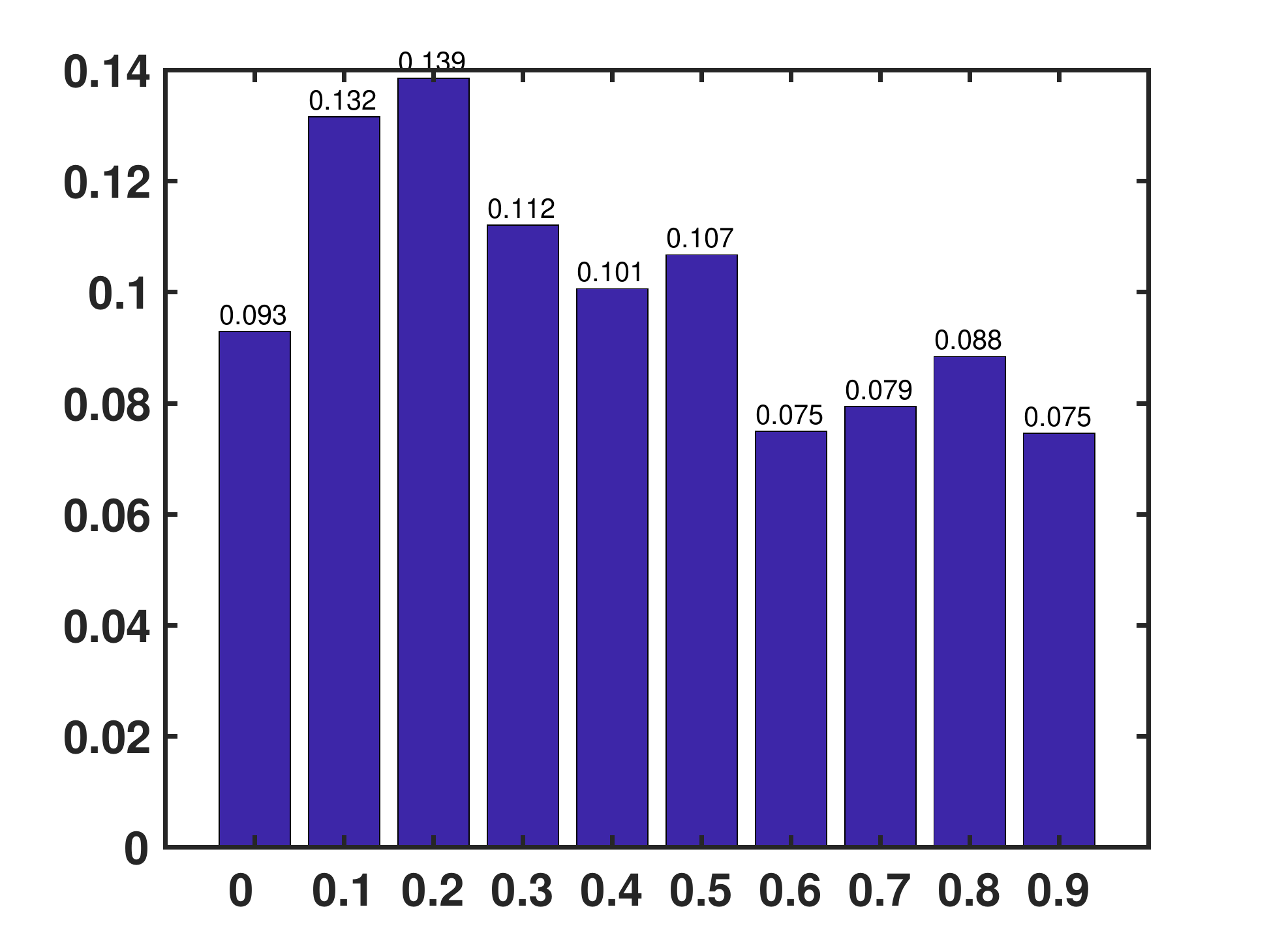}
      \caption{object area}
      \label{fig:portrait_yolo_area}
    \end{subfigure}
\caption{The distributions of (a) the score obtained from the pose estimator, (b) the normalized area obtained from the pose estimator, (c) the detection probability obtained from the object detector, and (d) the normalized area obtained from the object detector for our ground-truth images with a ``person'' as a common detectable object by the pose estimator and the object detector.}
\label{fig:person_analysis}
\end{figure*}

\begin{figure*}[!tb]
    \begin{subfigure}[b]{0.245\textwidth}
    	\includegraphics[height=1.0in]{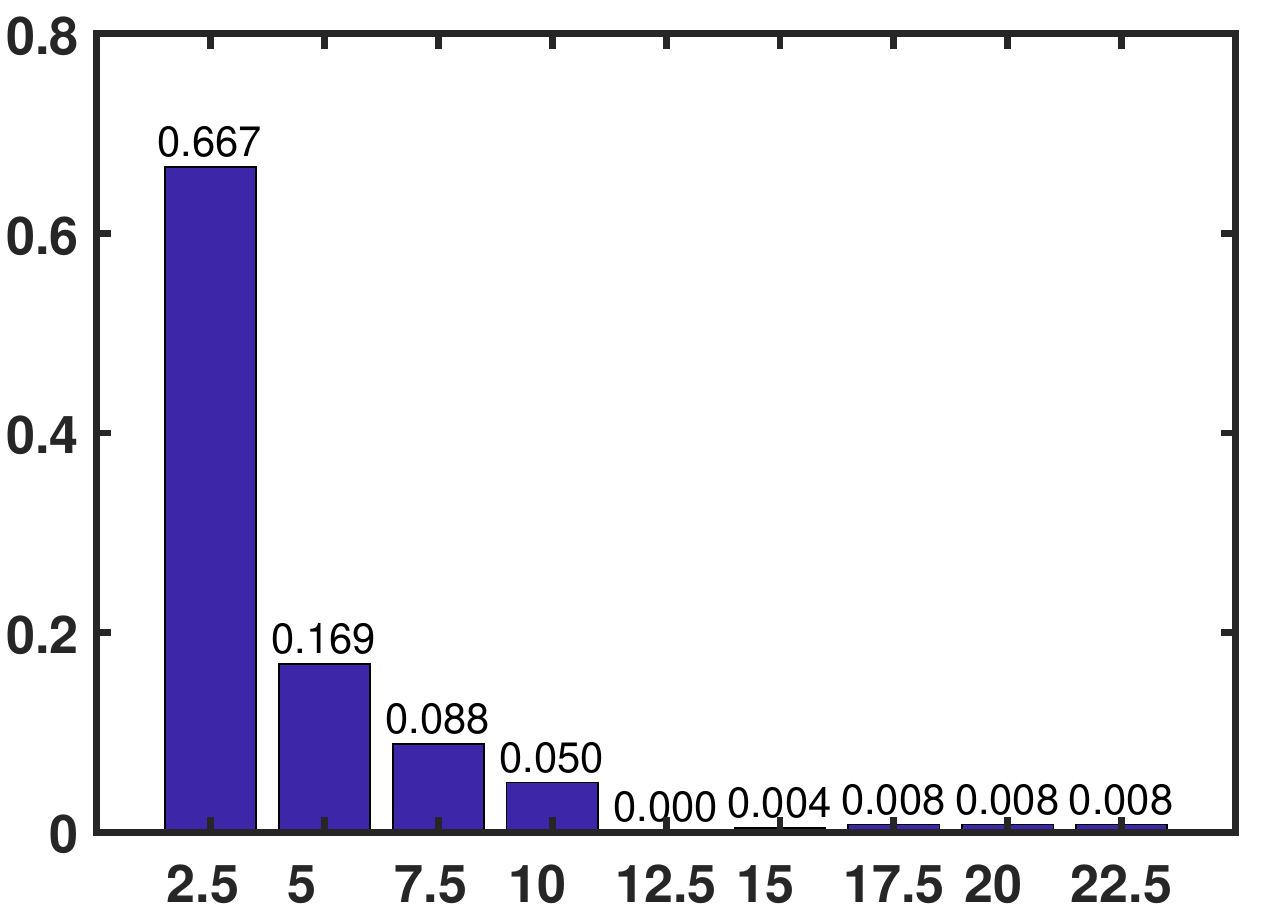}
      \caption{pose score}
      \label{fig:other_rtmppe_score}
    \end{subfigure} 
    \begin{subfigure}[b]{0.245\textwidth}
    	\includegraphics[height=1.0in]{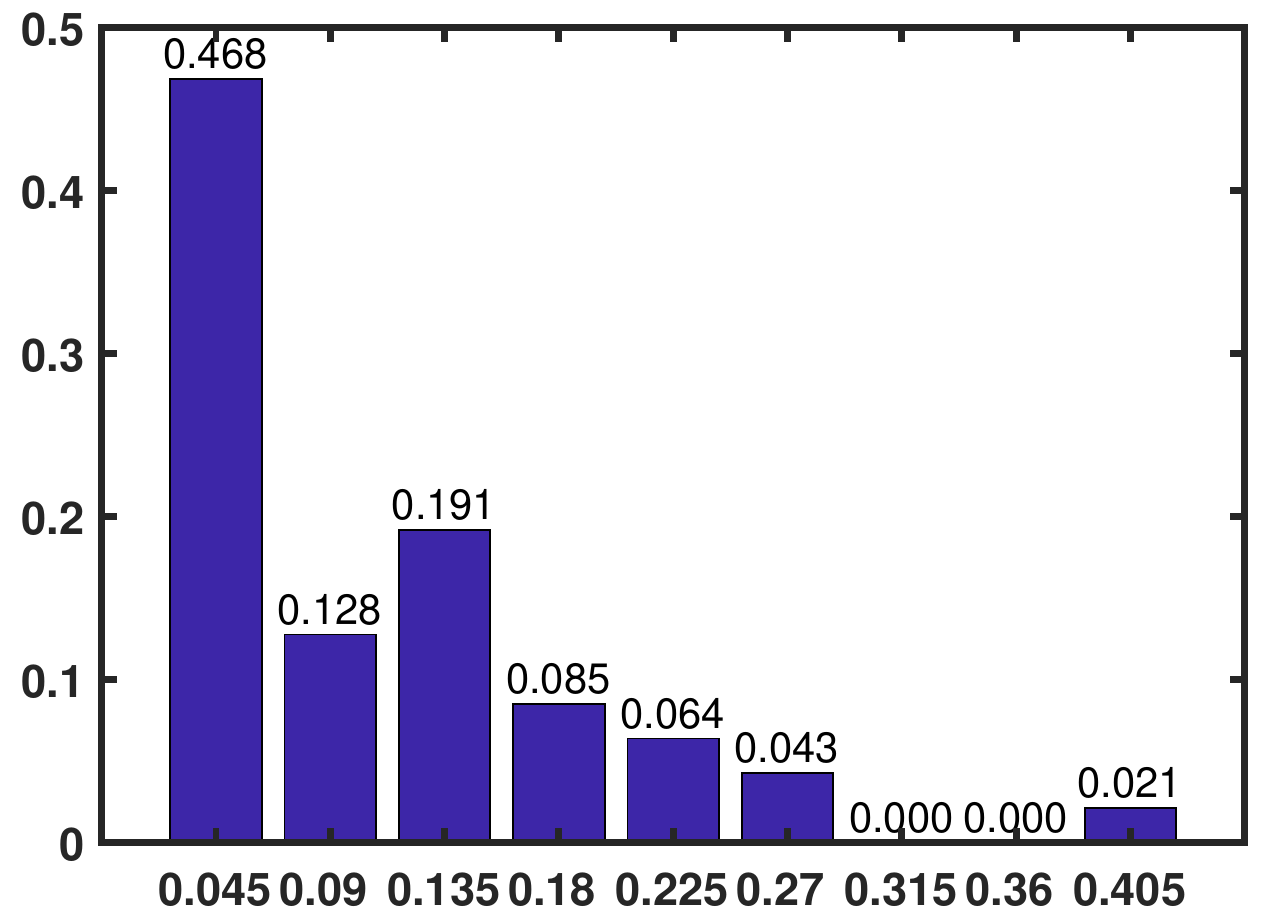}
      \caption{pose area}
      \label{fig:other_rtmppe_area}
    \end{subfigure} 
    \begin{subfigure}[b]{0.245\textwidth}
    	\includegraphics[height=1.0in]{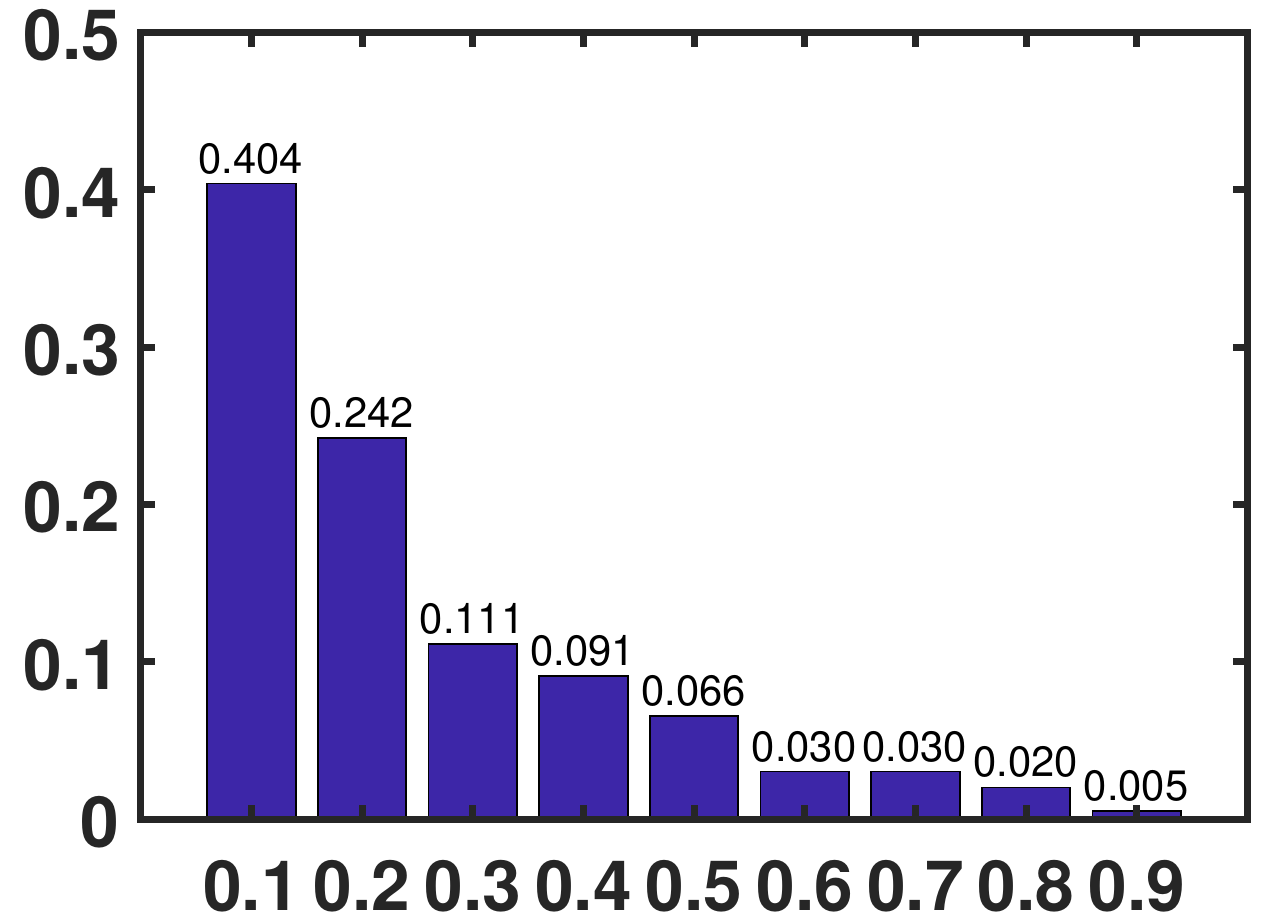}
      \caption{object probability}
      \label{fig:other_yolo_prob}
    \end{subfigure} 
    \begin{subfigure}[b]{0.245\textwidth}
    	\includegraphics[height=1.0in]{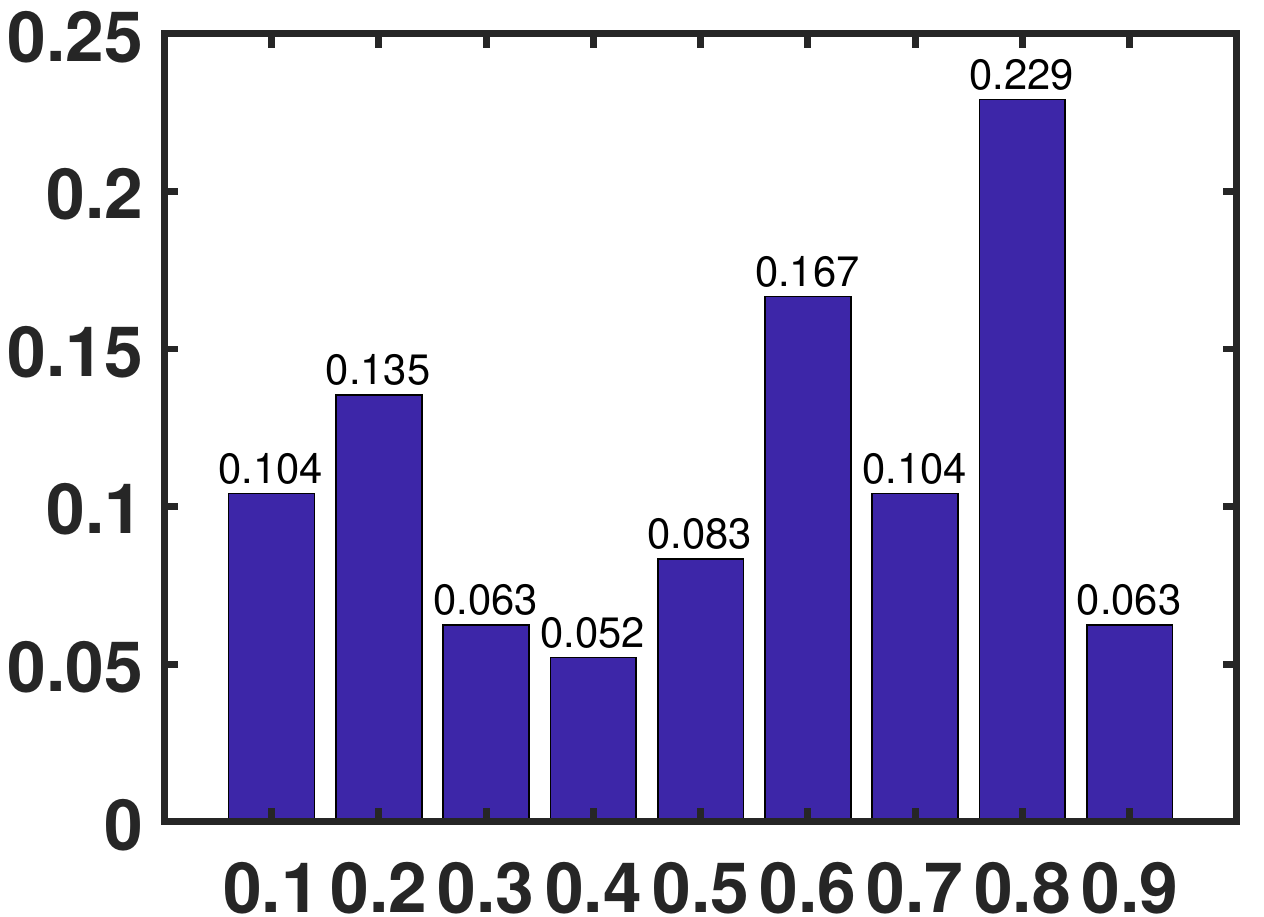}
      \caption{object area}
      \label{fig:other_yolo_area}
    \end{subfigure}
\caption{For the ground-truth images without any ``person'', the distributions of (a) the highest score (if any) obtained from the pose estimator, (b) the highest normalized area of the highest score object (if any) obtained from the pose estimator, (c) the detection probability for the dominant object (if any) obtained from the object detector, and (d) the normalized area of the dominant object (if any) obtained from the object detector.}
\label{fig:nonperson_analysis}
\vspace{-0.5\baselineskip}
\end{figure*}

Figure~\ref{fig:person_analysis} shows the distributions of the features obtained from our pose estimator and our object detector for ``person'' as a common object between the pose estimator and the object detector. In some images, no person is detected by the pose estimator and the object detector, because the pose estimator or the objector have detection error or actually there is no person in the image. We consider such detections as non-person object detection. Figure~\ref{fig:nonperson_analysis} shows the distributions of those features obtained from our pose estimator and our object detector, when there is no person in our ground-truth images. We have removed the frequency of the first component, {\it i.e.} score or area == 0, from all of the curved in Figure~\ref{fig:nonperson_analysis}, because probability of zero score/area is very high and we want to bold the probabilities of the other score/area values.

Figure~\ref{fig:portrait_rtmppe_score} shows pose estimator's score does not have enough sensitivity to detect a person, because the distribution is similar to a uniform probability mass function (PMF). Similarly, Figure~\ref{fig:portrait_yolo_area} shows object detector's normalized area does not have enough sensitivity to detect a person, because the distribution is pretty uniform. But, object detector's probability in Figure~\ref{fig:portrait_yolo_prob} and pose estimator's normalized area in Figure~\ref{fig:other_rtmppe_area} are not similar to uniform distribution, and we can infer some cut-off thresholds from them.

\begin{figure}[!tb]
\centering
\includegraphics[width=0.85\linewidth]{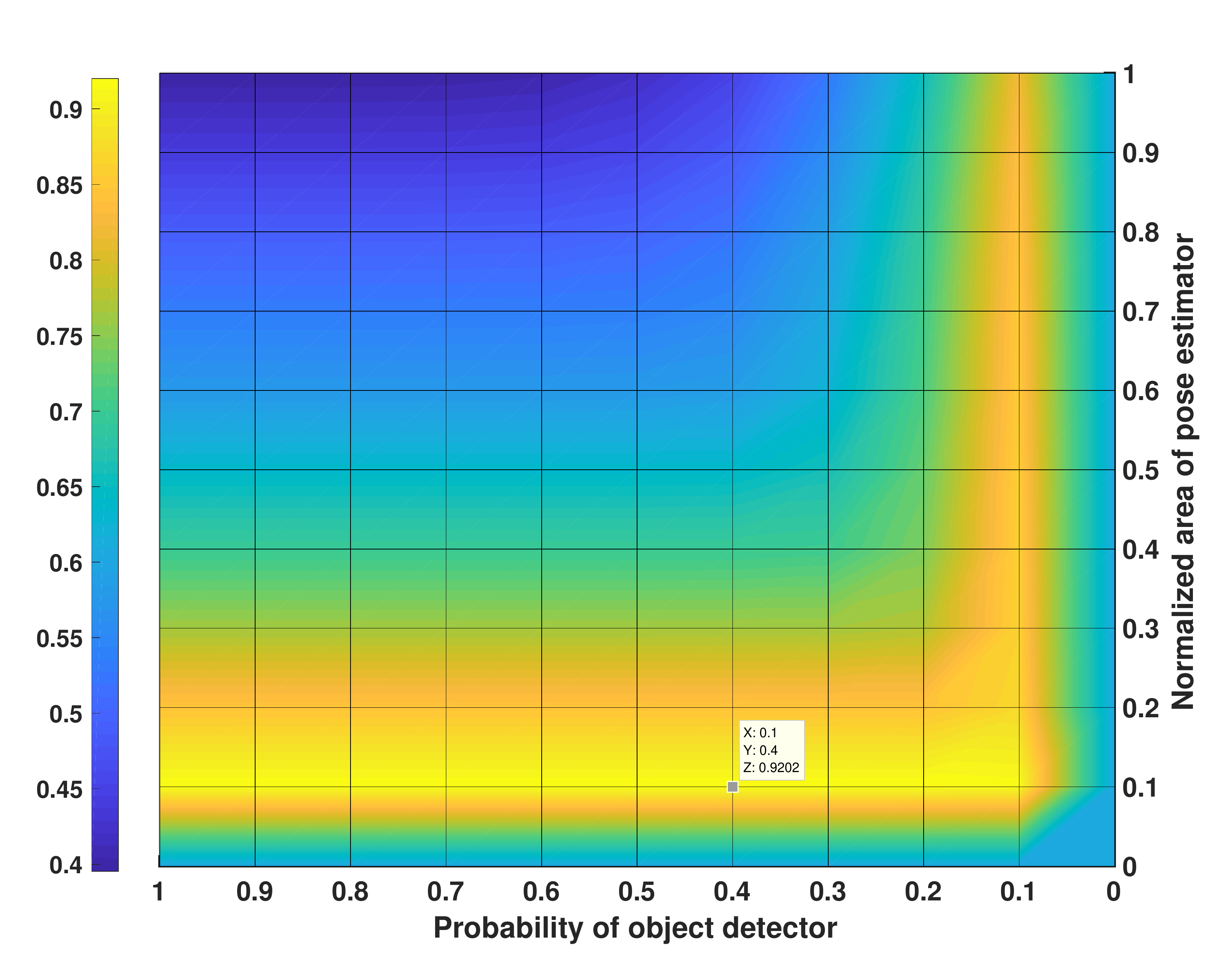}\\
\vspace{1.0mm}
\caption{The 2D ROC curve for the normalized area of the pose estimator and the detection probability of the object detector as a heat map.}
\label{fig:2droc}
\end{figure}

First, we derive the 2D probability density function (PDF) of the these mutual features including the normalized area by the pose estimator and the detection probability by the object detector. Second, we determine the 2D receiver operating characteristic (ROC) curve of these two parameters as a heat map. Finally, we search on the heat map and find the optimal point for these two mutual features. As shown in Figure~\ref{fig:2droc}, it can be inferred from 2D ROC of these two features that the optimal cut-off thresholds for ``person'' object detection in an image are object detector's probability 40\% and pose estimator's normalized area 10\% that leads to a detection accuracy ({\it i.e.} 92.2\%) higher than other detectors' accuracy solely.

\subsubsection{Indexing with Integrated Object Detection}
\label{sec:exp_iod}
\begin{figure}[!tb]
\centering
\includegraphics[width=0.80\linewidth]{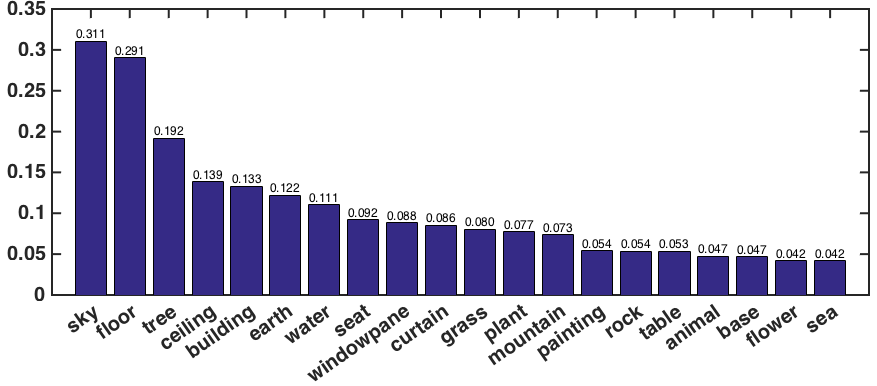}\\
\vspace{1.0mm}
\caption{The distribution of the highly-repetitive semantic classes detected in the dataset.}
\label{fig:obj_dist}
\vspace{-0.5\baselineskip}
\end{figure}

As shown in Figure~\ref{fig:method}, we do indexing of the 500px dataset by performing various detections. It is observed that once we integrate the object detectors in the dataset, all of the potentially detectable objects appear in the output. 
In addition, by collecting and distributing the results, the distribution (i.e., the frequency of the total) of the semantic classes detected in our dataset except for the highly-repetitive ones ("person" and "wall") is illustrated in Figure~\ref{fig:obj_dist}. 

\begin{figure}[!tb]
\centering
\includegraphics[width=0.80\linewidth]{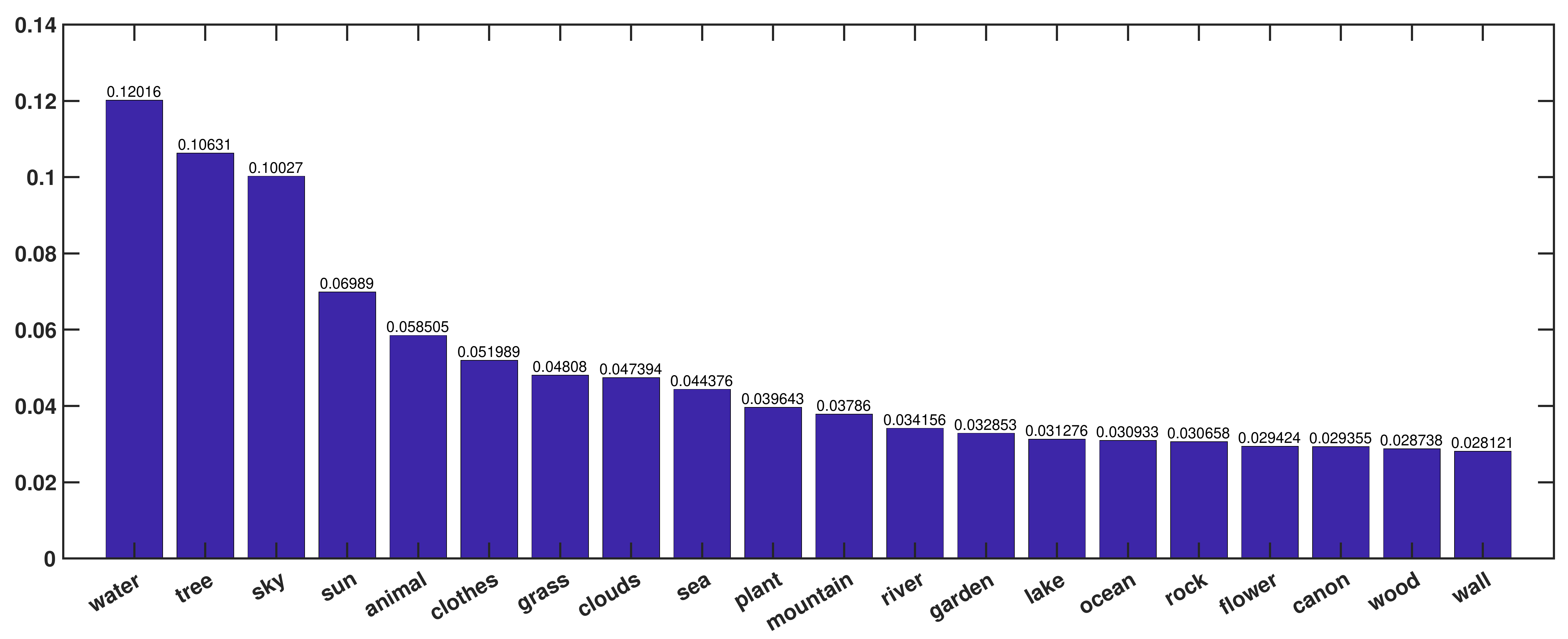}\\
\vspace{1.0mm}
\caption{The distribution of the mostly-available objects in the photos of the dataset which are manually extracted.}
\label{fig:manual_obj_dist}
\vspace{-0.5\baselineskip}
\end{figure}

The detectable objects by the IOD are not complete list of all available objects in any photo, but they can cover mostly-used objects in the photos. To investigate the case, we manually extract the available objects in 1600 random images from the dataset. Figure~\ref{fig:manual_obj_dist} shows the distribution of the objects highly available in the photos of the dataset.

\subsubsection{Portrait Category Detection}
\label{sec:exp_cade}
As mentioned in Section~\ref{sec:cade}, we start with top-down hierarchical clustering to specify the genre of the input image, and then we do multi-class categorization for portrait images. We train our model having 40 suggested features on a set of 6407 annotated portraits from the 500px dataset, and we test the model on another set of 1508 annotated portraits. The mean average accuracy of the model is listed in Table~\ref{tab:decomp_cade} categorized by various styles.

Also, we just consider the first 16 features for object detectors including general max and number of detected people in the image as mentioned in Section~\ref{sec:cade}, and train a model using the same ground truth as before. The current model is our baseline model, because it can be used for any other object detector, as the features can be defined in other object detector domains as well. To compare rationally with this baseline, we test the same set of images from our ground truth. The second line in Table~\ref{tab:decomp_cade} listed the baseline results. Because we remove limb features, the baseline has no ability to detect sub-genres such as hand-only, leg-only, no-face, and sideview.

\begin{table*}[!tb]
\begin{center}
\setlength{\tabcolsep}{1em}
{\renewcommand{\arraystretch}{1.3}
 \begin{tabular}{c|ccccccccc} 
  \hline
   & facial & full-body & group & hand & leg & no-face & sideview & two & upper-body \\
  \hline
  Our CaDe & 94.35 & 92.40 & 85.90 & 44.0 & 74.57 & 79.35 & 67.50 & 74.74 & 90.81\\
  \hline
  16-feat Baseline & 61.81 & 77.50 & 62.81 & N/A & N/A & N/A & N/A & 51.92 & 47.95\\
  \hline
 \end{tabular}
}
\caption{The accuracy results of our category detector (CaDe) for ground-truth images from the 500px dataset compared to a baseline.}
\label{tab:decomp_cade}
\end{center}
\vspace{-2.0\baselineskip}
\end{table*}

\begin{figure}[!tb]
\centering
\includegraphics[width=0.80\linewidth]{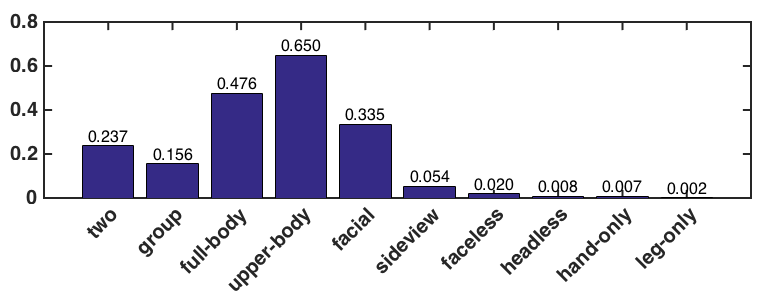}\\
\caption{The distribution of the portrait categories with respect to the number of corresponding images.}
\label{fig:catdist}
\vspace{-0.5\baselineskip}
\end{figure}

After portrait category indexing of the 500px dataset, the distribution of the portrait categories with respect to the number of corresponding images in each category divided by the total number of images is shown in Figure~\ref{fig:catdist} stating the number of photos in full-body, upper-body, facial, two, group, and side-view categories are high, but there are not enough samples for face-less, head-less, leg-only, and hand-only categories which is not a big deal, because these categories are not very popular.

\subsubsection{Artistic Pose Clustering}
\label{sec:exp_arpose}
\begin{figure}[!tb]
\centering
\includegraphics[width=0.80\linewidth]{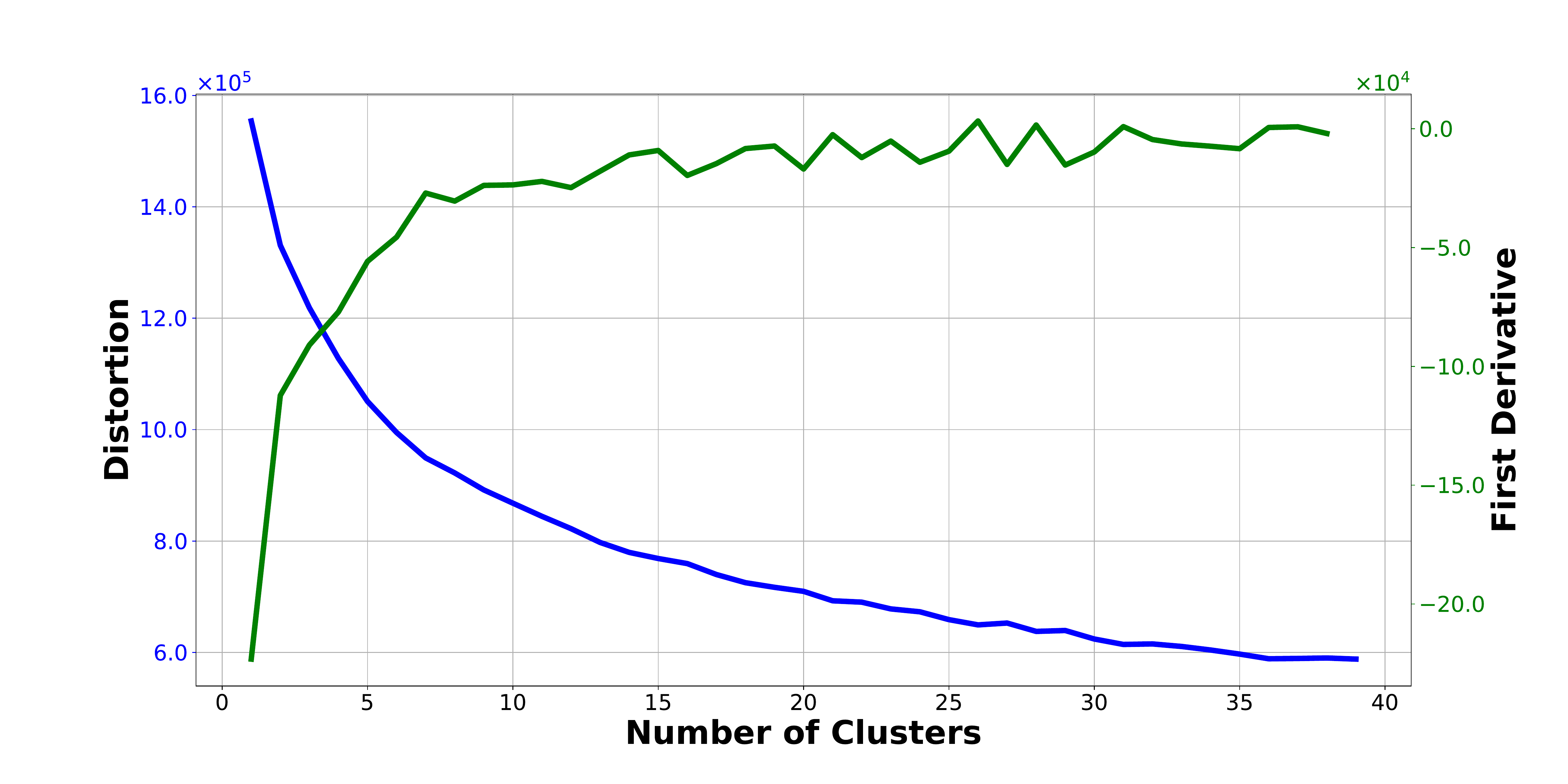}\\
\vspace{1.0mm}
\caption{The result of elbow method on the dataset. We could spot the elbow around 13-17 clusters. We are also showing the first derivative of the distortion, to show where it is going to flatten out.}
\label{fig:elbow}
\vspace{-1.0\baselineskip}
\end{figure}

Regarding artistic pose clustering, we conduct an experiment to cluster similar professional poses using our features explained in Section~\ref{sec:arpose}. We do the clustering with a various number of cluster heads, and we find the optimal number of cluster heads for the 500px dataset using elbow method. That being said, we use the elbow method and do the clustering 40 times with the different number of clusters ranged from 1 to 40. This method calculates the sum of squared errors (the distance of each point to the center of its cluster) and it is expected to see an elbow pattern in the plot of this error when the number of clusters is increasing. The result of this method on the 500px dataset is depicted in Figure~\ref{fig:elbow} in Section~\ref{sec:exp_arpose}, which indicates that the best choice for the number of clusters in this dataset is between 13 and 17.

\subsection{User Study for Composition}
\label{sec:exp_comp}
Many computer vision papers compare with their peers to show the effectiveness of their approach, but currently there exists no other similar or comparable system in the literature to compare with our proposed framework. To evaluate the functionality and the performance of our method, and measure how much the recommended photos are relevant to the query and helpful to the photographer, we conduct a quantitative user study based on the human subject results to compare our method with two other reasonable baselines. The first baseline is a semantic and scene retrieval method based on state-of-the-art CNN model~\citep{sharif2014cnn} and the other baseline is a non-CNN retrieval method based on the color, shape, and texture features~\citep{mitro2016content}. To create the baselines, all 4096 generic descriptors via public CNN model \citep{chatfield2014return} trained on ImageNet \citep{deng2009imagenet} are extracted for the 500px dataset images as well as the features of non-CNN method \citep{mitro2016content}.

\begin{figure}[!tb]
\centering
\hspace{-0.12in}
\resizebox{0.63\textwidth}{!}{\begin{tabular}{m{.55in}|m{3.5in}}

\includegraphics[height=0.73in]{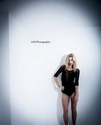}&
\includegraphics[height=0.73in]{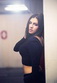} 
\hspace{0.04in}
\includegraphics[height=0.73in]{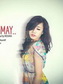}
\hspace{0.04in}
\includegraphics[height=0.73in]{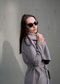}
\hspace{0.04in}
\includegraphics[height=0.73in]{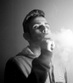}\\

\includegraphics[height=0.73in]{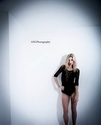}&
\includegraphics[height=0.67in]{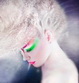}
\hspace{-0.1in}
\includegraphics[height=0.67in]{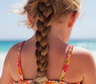}
\hspace{-0.1in}
\includegraphics[height=0.67in]{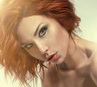}
\hspace{-0.1in}
\includegraphics[height=0.67in]{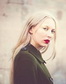}\\

\includegraphics[height=0.73in]{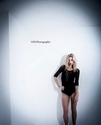}&
\includegraphics[height=0.73in]{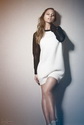}
\hspace{0.02in}
\includegraphics[height=0.73in]{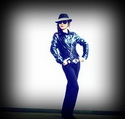}
\hspace{0.02in}
\includegraphics[height=0.73in]{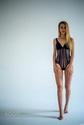}
\hspace{0.02in}
\includegraphics[height=0.73in]{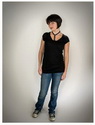}\\

\end{tabular}
}
\caption{The results of CNN (1st row), non-CNN (2nd row), and our method (3rd row) for a sample shot at the left side.}
\label{fig:retrieve}
\vspace{-1.0\baselineskip}
\end{figure}

We select a variety of image queries (38 queries) based on many types of categories such as background scene and semantics, single versus group, full-body, upper-body, facial, standing versus sitting, and male versus female. To be fair, we do not use customized queries as shown in Figure~\ref{fig:compose}, and we just focus on a single highlighted feature in each image query with the same question throughout the study. Using a PHP-based website with a usage guidance, the hypothesis tests are asked, and the outputs of the methods are randomly shown in each row to be chosen by 87 participants. Our framework received 74.20\% of the 1st ranks among the tests compared to 20.76\% CNN as 2nd rank and 5.04\% non-CNN as 3rd rank. Figure~\ref{fig:retrieve} illustrates the results of all methods (CNN: 1st, non-CNN: 2nd, ours: 3rd row) with respect to a similar shot at the left side. As it is realized from Figure~\ref{fig:retrieve} and our study, because the other methods cannot capture genre categories, scene structure, and corresponding poses of the query shot, it is common as shown in the figure that mixed categories are suggested by other semantic/category/pose-agnostic methods. As we hierarchically index the 500px dataset and recognize the right genre, category, pose and semantic classes, our semantic/category/pose-aware framework accessing to the indexed dataset can retrieve better related photography ideas.

\begin{figure}[!tb]
\centering
\includegraphics[width=0.80\linewidth]{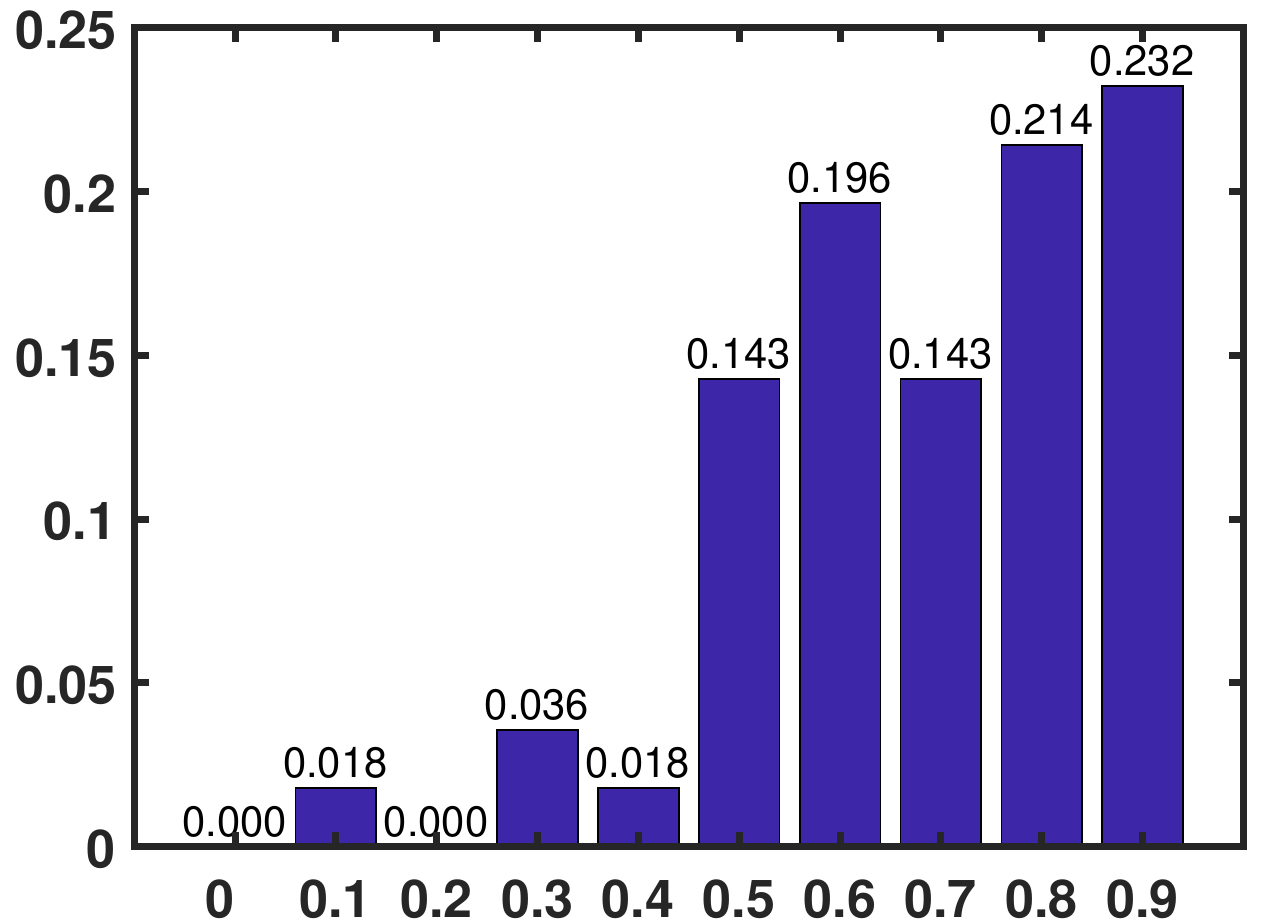}
\caption{The histogram of the acceptability rate for our recommended photos versus the total number of recommendations including other baselines.}
\label{fig:recomeval}
\vspace{-1.0\baselineskip}
\end{figure}

As mentioned, the expected value of the accepted recommended photos by the participants with respect to the total number of recommendations including the baselines is 74.20\%. More accurately, the histogram of the acceptability rate for the queries of the user study is shown in Figure~\ref{fig:recomeval}. The x-axis shows the acceptability rate ranged from 0 to 1 with 0.1-width bins, {\it i.e.}, what percentage of the participants has accepted our recommended photos for some queries. The y-axis shows the frequency of our accepted recommendations by the total number of the examined corresponding queries ({\it i.e.} probabilities) which fall into each bin. The histogram has indicated that 23.2\% of our recommended photos were accepted by over 90\% of the participants, 44.6\% of them with over 80\%, 58.9\% of them with over 70\%, and 92.8\% of them with over 50\%. Consequently, the majority of the recommended photos are accepted with a mean of 74.20\%. 

\vspace{-2.0\baselineskip}
\subsection{Runtime Analysis}
For training purposes, we mostly used an NVIDIA Tesla K40 GPU which took couple of days for the intensive computations on the dataset. To analyze the runtime performance of our method, we collect images with various styles and categories. Offline processing (indexing) of the 500px dataset is not included in the runtime results. We next perform IOD (including object detection, pose estimator, and scene parser), CaDe, ArPose as three parallel processes on those images to measure the average duration of the decomposition step. We then perform PAIR in composition step to get a ranked list of retrieved results for each image. Also, we randomly select the preferred style set, and perform the matching algorithm for each image as a single available shot. We sum up the average time for these three sequential steps to get the average runtime. The runtime is roughly proportional to the number of decomposed images because the IOD step which uses deep-learned models for detection is time-consuming. Among the detectors in IOD, pose estimator takes longer than the others and pretty independent from the number of people in the image with an average duration around 94.7 ms. Using a parallel structure shown in Figure~\ref{fig:method}, the end-to-end runtime is around 103.5 ms.

\section{Conclusions and Discussions}
\label{sec:conc_disc}

We have collected a large dataset for portrait and landscape photography ideas and introduced a new framework for composition assistance which guides amateur photographers to capture better shots. As the number of photography ideas increases, retrieving and matching the camera photo directly with an image in the dataset becomes more challenging. Furthermore, the retrieving system not only finds similar images but also searches for images with similar semantic constellations with better composition through decomposition and composition steps. After providing feedback for the photographer, the camera tries to match the final pose with one of the retrieved feedbacks, and make an astonishing shot. The performance of our framework has been evaluated by various experiments. Another merit of this work is the integration of the deep-based detectors which can make the whole process automatic. 

\subsection{Genre Extendibility}
The general idea behind this work can be extended to other photography genres such as candid, fashion, close-up, and architectural photography using other appropriate detectors. The criteria for one genre are generally different from those for another. For instance, the pose is crucial in portrait photography, while leading lines and vanishing points can be important in architectural photography. 

\subsection{Enhancement in User Interaction}
The user-specified preferences (USP) should be quantified by the individual, but it may be difficult for them to accurately adjust the detectors' importance for their personal preference. They may want to do hierarchical preference, as some results are eliminated in each branch when going down the user-specified decision tree. Qualitatively they check the results and come up with a better decision, but it may be time-consuming for them. One can optimize the decision weights for a specific user after learning his/her behavior, and then they can just request for the ordering (not the weight values). We believe that there is still room to improve the interactions with the individual.

\subsection{Clustering of Photography Ideas}
Some future directions include working on an unsupervised learning approach that can cluster all the images based on various photography ideas. Recognizing the ideas is not easy for amateurs, and one shot can have multiple ideas. After clustering, we might detect new ideas. Therefore, it would be interesting to explore a metric to estimate the potential novelty of the current shot based on computing similarity to other shots.
\vspace{-1\baselineskip}

\subsection{Innovative Shot}
Another promising idea is to design a system where the camera automatically detects an innovative situation and takes a shot. Conventional methods in machine learning just use the history of the field to help amateurs take professional photos, and of course, these approaches can not go beyond it. After recognizing new photography ideas, the system can think of it as a compact space, not a finite discrete space, and it attempts to find a solution in this compact space. Fortunately, the complexity of the problem can change from an integer programming to a linear programming, but the way we define these compact spaces is hard based on the complexity of finding new photography ideas.

\begin{acknowledgements}
This research has been supported in part by Penn State University. We gratefully acknowledge the support of NVIDIA Corporation with the donation of the Tesla K40 GPUs used for this research. The work would not have been possible without the large quantity of creative photography ideas of the thousands of photographers who share their visual creations on the Internet. 
The authors would like to thank the participants of the user study.
\end{acknowledgements}

\bibliographystyle{spbasic}
\bibliography{references}

\end{document}